%% file: aaai25.tex
\documentclass[letterpaper]{article} 
\usepackage{aaai25}  
\usepackage{times}  
\usepackage{helvet}  
\usepackage{courier}  
\usepackage[hyphens]{url}  
\usepackage{graphicx} 
\usepackage{natbib}  
\usepackage{caption} 
\usepackage{algorithm}
\usepackage{algorithmic}
\usepackage{xspace}
\usepackage{xcolor}
\usepackage{booktabs}
\usepackage{multirow}
\usepackage{multicol}
\usepackage{amsmath}
\usepackage{amsfonts}       
\usepackage{amsthm}
\usepackage{cleveref}
\usepackage{subcaption}
\usepackage{mathtools}
\usepackage{listings}

\usepackage{newfloat}
\usepackage{listings}
\DeclareCaptionStyle{ruled}{labelfont=normalfont,labelsep=colon,strut=off} 
\floatstyle{ruled}
\newfloat{listing}{tb}{lst}{}
\floatname{listing}{Listing}
\theoremstyle{plain}
\newtheorem{theorem}{Theorem}[section]

\newtheorem{lemma}{Lemma}[section]
\newtheorem{corollary}{Corollary}[theorem]
\theoremstyle{definition}
\newtheorem{definition}[theorem]{Definition}

\theoremstyle{remark}

\newcommand{\optim}{SMMF\xspace}
\newcommand{\SCALAR}[1]{#1}
\newcommand{\MATRIX}[1]{\boldsymbol{\uppercase{#1}}}
\newcommand{\VECTOR}[1]{\boldsymbol{\lowercase{#1}}}
\newcommand{\WEIGHT}{\MATRIX{W}}
\newcommand{\weight}{\VECTOR{w}}
\newcommand{\BSET}{\{0, 1\}}
\newcommand{\RSET}{\mathbb{R}}
\newcommand{\ZSET}{\mathbb{Z}}
\newcommand{\NSET}{\mathbb{N}}

\newcommand{\BIGO}[1]{\mathcal{O}(#1)}
\newcommand{\BIGOU}[2]{\mathcal{O}_{#1}(#2)}
\newcommand{\tenfont}{\fontsize{10pt}{10pt}\selectfont }
\newcommand{\ninefont}{\fontsize{9pt}{9pt}\selectfont }

\title{\optim: Square-Matricized Momentum Factorization for Memory-Efficient Optimization}
\author {
    Kwangryeol Park, Seulki Lee\\
}
\affiliations{
    Artificial Intelligence Graduate School$^1$ and Department of Computer Science and Engineering$^2$, UNIST, South Korea\\
    \{pkr7098$^1$, seulki.lee$^2$\}@unist.ac.kr
}

\usepackage{bibentry}

\begin{document}

\maketitle
\begin{abstract}
\label{sec:abstract}
We propose \optim (Square-Matricized Momentum Factorization), a memory-efficient optimizer that reduces the memory requirement of the widely used adaptive learning rate optimizers, such as Adam, by up to 96\%. \optim enables flexible and efficient factorization of an arbitrary rank (shape) of the first and second momentum tensors during optimization, based on the proposed square-matricization and one-time single matrix factorization. From this, it becomes effectively applicable to any rank (shape) of momentum tensors, i.e., bias, matrix, and any rank-$d$ tensors, prevalent in various deep model architectures, such as CNNs (high rank) and Transformers (low rank), in contrast to existing memory-efficient optimizers that applies only to a particular (rank-2) momentum tensor, e.g., linear layers. We conduct a regret bound analysis of \optim, which shows that it converges similarly to non-memory-efficient adaptive learning rate optimizers, such as AdamNC, providing a theoretical basis for its competitive optimization capability. In our experiment, \optim takes up to 96\% less memory compared to state-of-the-art memory-efficient optimizers, e.g., Adafactor, CAME, and SM3, while achieving comparable model performance on various CNN and Transformer tasks.
\end{abstract}
\begin{links}
    \link{Code}{https://github.com/eai-lab/SMMF}
    \link{Extended version}{https://arxiv.org/abs/2412.08894}
\end{links}
\section{Introduction}
To identify the optimal weight parameters of deep neural networks, various optimization methods~\cite{abdulkadirov2023survey, martens2016second, amari2010information, liu1989limited} have been studied. One of the most popular approaches is SGD (Stochastic Gradient Descent)~\cite{ruder2016overview} which takes the weight update direction towards the current gradient with a learning rate uniformly applied to all weight parameters. To further improve SGD’s optimization performance, many adaptive learning rate optimizers, such as Adam~\cite{adam} and RMSProp~\cite{rmsprop}, have been proposed to leverage 1) history of the gradients to compute the momentum direction~\cite{ruder2016overview} and 2) the squared gradients to compute the adaptive learning rate for each weight parameter. Despite their lack of theoretical convergence guarantee in non-convex settings of many deep learning tasks, those adaptive learning rate optimizers have been empirically found to outperform SGD in practice.

However, since the momentum value of each weight parameter, which linearly increases over the size of a deep learning model, should be maintained in memory during the whole training process, the adaptive learning rate optimizers can easily limit the size of models that can be trained on memory-constrained platforms, e.g., embedded systems. Even when training small models like Transformer-base~\cite{transformer}, 1.4 GiB of memory is required. This means it would be unusable in environments with extremely limited memory devices, such as Raspberry Pi (1 GiB). To tackle the memory challenge of the adaptive learning rate optimization, several memory-efficient optimizers have been proposed. Adafactor~\cite{adafactor} and CAME~\cite{came} factorize the $2^{nd}$ momentum in the form of a matrix into a set of vectors to decrease the memory space required to store momentums, achieving comparable performance to Adam. SM3~\cite{sm3} reduces memory usage by approximating the similar elements of the $2^{nd}$ momentum into a smaller set of variables. Although they effectively reduce the memory space of adaptive learning rate optimizers by projecting a gradient tensor onto several rank-one vectors, 1) they apply only to a specific rank (shape) and pattern of momentum tensors, 2) their memory space is still huge (1.1 GiB) making them unsuitable for memory constrained devices, and 3) their optimization performance has not been theoretically analyzed and compared to that of Adam family~\cite{adam}.

In this paper, we propose \optim (Square-Matricized Momentum Factorization), a memory-efficient optimizer amicable to an arbitrary rank (shape) and pattern of both the $1^{st}$ and $2^{nd}$ momentum tensors, i.e., a vector, matrix, and rank-$d$ tensor, which reduces the amount of memory required in model optimization by up to 96\% compared to existing memory-efficient optimizers, e.g., Adafactor, CAME, and SM3. Unlike such existing memory-efficient optimizers, either confined to a particular 1) momentum rank (shape) (i.e., a rank-2 matrix) and/or 2) momentum pattern (i.e., a set of similar elements in a matrix)~\cite{sm3}, the proposed \optim performs competitive optimization without being restricted by the rank (shape) and pattern of momentums allowing the models to be trained on extremely memory constrained embedded systems from $\sim$0.001 to $\sim$1 GiB.

Given a rank-$\SCALAR{d}$ momentum tensor as $\RSET^{n_1\times \ldots \times n_d}$, \optim first finds $\hat{n},\hat{m}=\arg\min_{n, m}\lvert n - m \rvert$ such that $n m = \prod_{r=1}^{d}n_r$. Next, it converts the momentum $\RSET^{n_1\times \ldots \times n_d}$ into a matrix closest to square matrix $\RSET^{\hat{n} \times \hat{m}}$ with $\hat{n}$ and $\hat{m}$, which we call square-matricization. Then, the matrix $\RSET^{\hat{n} \times \hat{m}}$ is factorized into two vectors, $\RSET^{\hat{n}\times 1}$ and $\RSET^{1\times \hat{m}}$ at one go, by using NNMF (Non-Negative Matrix Factorization)~\cite{NNMF}. Since \optim only stores the resulting two vectors $\RSET^{\hat{n}\times 1}$ and $\RSET^{1\times \hat{m}}$ in memory, factorized from both the $1^{st}$ and $2^{nd}$ momentum $\RSET^{\hat{n} \times \hat{m}}$ that has been squared-matricized from the original rank-$d$ momentum $\RSET^{n_1\times \ldots \times n_d}$, it can decrease more memory when given high-rank momentums, e.g., the rank-4 weight tensors in CNNs. It is different from existing memory-efficient optimizers, e.g., Adafactor and CAME, that store $\prod_{r=1}^{d-2} n_r$ pairs of vectors factorized from a rank-$d$ momentum $\RSET^{n_1\times \ldots \times n_d}$ in memory.

We analyze the regret bound of the proposed \optim, proving that its optimization performance in a convex setup is similar to one of the Adam-based optimizers, i.e., AdamNC~\cite{amsgrad} that applies the beta schedule to Adam. To the best of our knowledge, \optim is the first factorization-based memory-efficient optimizer that conducts a regret bound analysis; none of the existing memory-efficient optimizers, e.g., Adafactor and CAME, provides such a theoretical study. The experiments on various CNN and Transformer models (\Cref{sec:experiment}) show the competitive results substantiating our analysis.
\section{Related Work} 
\label{sec:related_work}
\textbf{Adafactor}~\cite{adafactor} factorizes the $2^{nd}$ momentum matrix via Non-Negative Matrix Factorization (NNMF)~\cite{NNMF} that decomposes a non-negative matrix into two vectors by differentiating the I-divergence~\cite{Idivergence}. Theoretically, it reduces the memory complexity of the $2^{nd}$ momentum in the form of a non-negative matrix, i.e., $\MATRIX{V} \in \RSET^{n \times m}$, from $\BIGO{nm}$ to $\BIGO{n+m}$ with the two factorized vectors. Empirically, it shows comparable optimization to Adam on Transformers.

\noindent\textbf{CAME}~\cite{came}, a variant of Adafactor, is proposed as a memory-efficient optimizer for large batch optimization. To alleviate the unstable behavior of Adafactor, it introduces the factorized confidence term that guides the optimization direction, empirically achieving faster convergence on language models~\cite{t5, gpt2} at the cost of using more memory than Adafactor. Since CAME also requires a momentum to be a non-negative matrix to be factorized with NNMF, it slices a high-rank weight tensor, appearing in CNN models such as MobileNet~\cite{mobilenet}, into multiple matrices and factorize them separately. Hence, given a rank-$d$ $2^{nd}$ momentum $\MATRIX{V}$$\in$$\RSET^{n_1\times \ldots\times n_d}$, the memory complexity of CAME becomes $\BIGO{(n_{d-1} + n_d)\prod_{r=1}^{d-2}n_r}$, which is similar to Adafactor.

\noindent\textbf{SM3}~\cite{sm3}, unlike Adafactor and its variants such as CAME, applies the min-max scheme to approximate the similar elements of the $2^{nd}$ momentum to a smaller set of variables. It shows competitive optimization performance to Adam and Adafactor on Transformers that exhibit a grid pattern of similar elements in their weight matrices. Given a rank-$d$ momentum tensor $\RSET^{n_1 \times \ldots \times n_d}$, the memory complexity of SM3 becomes $\BIGO{\sum_{r=1}^d n_r}$ if similar elements appear on each axis of the weight tensor, which can be found in some Transformer weight matrices~\cite{sm3}.

Although those existing memory-efficient optimizers effectively reduce the memory requirement and perform competitive optimization primarily on the Transformer architectures~\cite{transformer, bert, gpt2, t5} by projecting the gradient onto rank-1 vectors, each optimizer has limitations. First, since Adafactor and CAME rely on matrix factorization~\cite{NNMF}, a momentum tensor should be first sliced into multiple matrices before being factorized, degrading the memory reduction effect given a high-rank momentum tensor. Next, SM3 needs sets of similar elements in a momentum tensor to perform effective optimization, neither easy nor guaranteed to find in the huge weight parameter space of many deep neural networks. However, unlike Adafactor and CAME, the proposed \optim applies one-time single matrix factorization to any rank (shape) of momentum tensors based on the proposed square-matricization without the memory increase caused by tensor-to-matrices slice. Also, since the proposed \optim utilizes NNMF, it does not require strong patterns on the weight parameter space, readily applying to an arbitrary pattern of weight tensors, in contrast to SM3 that assumes the existence of specific element patterns on each axis in a weight tensor. 
\section{\optim \\ (Square-Matricized Momentum Factorization)}
\label{sec:smmf}
\input{algorithms/brief_algorithm}

\Cref{alg:brief_algorithm} shows the overall procedure of the proposed \optim (Square-Matricized Momentum Factorization), applied to the weight tensor (momentum) at each layer of the model. In short, given the $1^{st}$ and $2^{nd}$ momentum tensors as $\MATRIX{M}, \MATRIX{V} \in \RSET^{n_1 \times \ldots \times n_d}$, \optim reduces the memory complexity required for optimization into $\BIGOU{M}{\hat{n} + \hat{m}}$ and $\BIGOU{V}{\hat{n} + \hat{m}}$ for $\MATRIX{M}$ and $\MATRIX{V}$, respectively, with $\hat{n},\hat{m}=\arg\min_{n, m}\lvert n - m \rvert$ such that $n m = \prod_{r=1}^{d}n_r$ where $n_r, n, m, \hat{n}$, and $\hat{m}$ are in $\NSET$. It first transforms $\MATRIX{M}, \MATRIX{V} \in \RSET^{n_1 \times \ldots \times n_d}$ into a matrix closest to the square (square-matricization), i.e., $\MATRIX{M}, \MATRIX{V} \in \RSET^{\hat{n} \times \hat{m}}$ where $\hat{n} \simeq \hat{m}$, and then applies NNMF (\Cref{alg:nnmf}) to $\MATRIX{M}, \MATRIX{V} \in \RSET^{\hat{n} \times \hat{m}}$ as one-time single matrix factorization (compression). Since the $1^{st}$ momentum $\MATRIX{M}$ can be negative, unlike the $2^{nd}$ momentum $\MATRIX{V}$ that is non-negative, we apply NNMF to the absolute values of $\MATRIX{M}$ and store the sign of each element of $\MATRIX{M}$ as a separate set of binary values (1-bit). Although it incurs extra memory overhead $\BIGOU{M}{\hat{n} \hat{m}}$ on top of $\BIGOU{M}{\hat{n} + \hat{m}}$ and $\BIGOU{V}{\hat{n} + \hat{m}}$, its memory footprint is 32 times smaller than storing the original $\MATRIX{M}$ with the 32-bit floating-point format. The following subsections describe each step of \optim.
\input{algorithms/square_matrixlization}
\subsection{Square-Matricization}
\label{sec:square-matricization}
In \Cref{alg:brief_algorithm}, \optim first obtains a rank-$d$ gradient $\MATRIX{G}_t \in \RSET^{n_1\times \ldots \times n_d}$ for the weight and bias tensor at each layer of the model and converts it into a matrix closest to a square matrix $\bar{\MATRIX{G}}_t \in \RSET^{\hat{n} \times \hat{m}}$ where $\hat{n} \simeq \hat{m}$, for factorization, naturally leading to the square-matricization of the $1^{st}$ and $2^{nd}$ momentum, $\MATRIX{M}$ and $\MATRIX{V}$. To this end, we propose a square-matricization method that reshapes $\MATRIX{G}_{t} \in \RSET^{n_1 \times \ldots \times n_d}$ into a matrix closest to a square matrix $\bar{\MATRIX{G}}_{t} \in \RSET^{\hat{n} \times \hat{m}}$ such that $n m = \prod_{r=1}^{d} n_{r}$ and $(\hat{n}, \hat{m}) = \arg\min_{n, m}(n + m) = \arg\min_{n, m}\lvert n - m \rvert$, where $\hat{n}, \hat{m}, n, m \in \NSET$. Following theorems show that the square-matricization of $\bar{\MATRIX{G}}_t \in \RSET^{\hat{n} \times \hat{m}}$, i.e., having $\hat{n} \simeq \hat{m}$, also minimizes $\hat{n} + \hat{m}$.

\begin{theorem}    \label{thm:the_theorem_of_square_matrixlization_discrete2}    
    Given $n_r \in \NSET$, $r \in [1, d]$, and a constant $\SCALAR{N} = \prod_{r=1}^d n_r$, then $\prod_{r=1}^{d-2} n_r(n_{d-1} + n_d)$ decreases if both $n_{d-1}$ and $n_d$ increase (Proof provided in \Cref{proof:the_theorem_of_square_matrixlization_discrete2}).
\end{theorem}
\begin{corollary}
    \label{cor:the_theorem_of_square_matrixlization_discrete2}
    Given $\SCALAR{N} = \prod_{r=1}^d n_r$, there exist $N = \hat{n} \hat{m}$ such that $\hat{n} + \hat{m} = \min \prod_{r=1}^{d-2} n_r(n_{d-1} + n_d)$, $(\hat{n}, \hat{m}) \in \NSET$. 
\end{corollary}
\begin{theorem}
    \label{thm:the_theorem_of_square_matrixlization_discrete}
    Given $n_r, n, m \in \NSET$, $r \in [1, d]$, $\SCALAR{N} = \prod_{r=1}^d n_r = nm$, and $n \leq m$, then $\hat{n}, \hat{m} = \arg\min_{n, m}(n + m) = \arg\min_{n, m}\lvert n - m \rvert$ (Proof provided in \Cref{proof:the_proof_of_square_matrixlization_discrete}).
\end{theorem}
From \Cref{cor:the_theorem_of_square_matrixlization_discrete2}, square-matricizing $\MATRIX{G}_t \in \RSET^{n_1 \times \ldots \times n_d}$ into $\bar{\MATRIX{G}}_t \in \RSET^{\hat{n} \times \hat{m}}$ reduces the memory complexity since $\prod_{r=1}^{d-2} n_r(n_{d-1} + n_d) \leq \prod_{r=1}^d n_r$. Also, based on \Cref{thm:the_theorem_of_square_matrixlization_discrete}, minimizing $\lvert n - m \rvert$ is equivalent to minimizing $n + m$. From this, we derive the square-matricization algorithm (\Cref{alg:square_matrixlization}) that finds $\hat{n}$ and $\hat{m}$, which minimizes $\hat{n}+\hat{m}$ by solving $(\hat{n}, \hat{m}) = \arg\min_{n, m} \lvert n - m \rvert$. By reshaping a rank-$d$ gradient into a matrix closest to a square matrix through square-matricization, it becomes able to perform one-time single matrix factorization, which minimizes the memory complexity of $\BIGOU{M, V}{\hat{n} + \hat{m}}$ for the $1^{st}$ and $2^{nd}$ momentum $\MATRIX{M}$ and $\MATRIX{V}$. Thus, the memory usage of \optim becomes smaller than those of existing memory-efficient optimizers~\cite{adafactor, came} that should slice a high-rank tensor into a bunch of matrices for multiple factorizations, i.e., $\BIGO{\prod_{r = 1}^{d - 2} n_r (n_{d - 1} + n_d)}$ given $\MATRIX{V} \in \RSET^{n_1 \times \ldots \times n_d}$. That is, the memory complexity of CNNs having high-rank gradient tensors grows over the rank of gradients~\cite{adafactor, came}, whereas that of \optim does not.

\subsection{Decompression and Compression}
\label{sec:decompression_and_compression}
\textbf{Decompression $\rightarrow$ Compression.} After square-matricizing the gradient, \optim decompresses the $1^{st}$ and $2^{nd}$ momentum from two vectors factorized at the previous step $t{-}1$ to update the momentums, as in \Cref{alg:brief_algorithm}. Then, it compresses the momentums obtained at step $t$ into vectors and updates the weight $\WEIGHT$ using the decompressed momentums. We call this process the $decompression \rightarrow compression$ scheme, in which the gradient $\bar{\MATRIX{G}}_t$ at the current step $t$ is reflected to the $1^{st}$ and $2^{nd}$ momentum before it is factorized, enabling the precise update of the weight.

The significance of information in the current gradient, e.g., tensor patterns, has been emphasized in some previous works~\cite{sm3}. Since reshaping and factorization of the gradient may ruin some potentially important patterns of the momentums contained in the previous gradient $\MATRIX{G}_{\tau < t}$, reflecting the current gradient $\MATRIX{G}_t$ (and its pattern) is crucial in model performance. Thus, by performing decompression with $\MATRIX{G}_t$ prior to updating and compressing (factorizing) $\MATRIX{M}_t$ and $\MATRIX{V}_t$, it is expected to improve the optimization performance. On the contrary, existing optimizers, such as Adafactor, first compress  $\MATRIX{G}_t$ and then updates momentums through decompression, which we call the $compression \rightarrow decompression$ scheme. In this scheme, some useful information of $\MATRIX{G}_t$ would be lost by compression (factorization), implying that an optimizer can hardly utilize the intact state of $\MATRIX{G}_t$, likely to degrade the model performance.
\input{algorithms/matrix_decompression}

\noindent\textbf{Decompression.} First, in the decompression phase (\Cref{alg:matrix_decompression}), the $1^{st}$ and $2^{nd}$ momentum, \{$\hat{\lvert \MATRIX{M}}_{t-1} \rvert$, $\hat{\MATRIX{V}}_{t-1}$\} $\in \RSET^{\hat{n} \times \hat{m}}$, are defactorized from the two vectors for each, i.e., \{$\VECTOR{r}_{\MATRIX{M}_{t-1}}, \VECTOR{c}_{\MATRIX{M}_{t-1}}$\} for $\lvert \hat{\MATRIX{M}}_{t-1} \rvert$ and $\{\VECTOR{r}_{\MATRIX{V}_{t-1}}$, $\VECTOR{c}_{\MATRIX{V}_{t-1}}\}$ for $\hat{\MATRIX{V}}_{t-1}$, which have been factorized from the square-matricized momentums at the previous step $t$$-$$1$, by performing outer product between them. To apply NNMF to the $1^{st}$ momentum $\boldsymbol{\hat{M}}_{t}$, its sign values are stored as a binary matrix $\MATRIX{S}_{\MATRIX{M}_{t-1}}$$\in$$ \BSET^{\hat{n} \times \hat{m}}$ in the compression phase and restored back to the defactorized $1^{st}$ momentum $\boldsymbol{\hat{M}}_{t}$ in an element-wise manner. Then, $\boldsymbol{M}_{t}$ and $\boldsymbol{V}_{t}$ are updated by using two coefficients $\beta_{1, t}$ and $\beta_{2, t}$.

\input{algorithms/matrix_compression}
\noindent\textbf{Compression.} Next, in the compression phase (\Cref{alg:matrix_compression}), the sign values of $\MATRIX{M}_t$ is stored as a binary sign matrix $\MATRIX{S}_{\MATRIX{M}_t}$ for the next step of optimization, and both $\lvert \MATRIX{M}_t \rvert$ and $\MATRIX{V}_t$ are factorized into two vectors for each, i.e., $\{ \boldsymbol{r}_{M_{t}},\boldsymbol{c}_{M_{t}} \}$ and $\{ \boldsymbol{r}_{V_{t}},\boldsymbol{c}_{V_{t}} \}$, respectively. To reduce the computation required for compression, it determines whether to normalize $\VECTOR{r}$ or $\VECTOR{c}$ based on the shape of the matrix.

\noindent\textbf{Weight Update.} Lastly, the weight update term $\boldsymbol{U}\in \mathbb{R}^{\hat{n}\times\hat{m}}$ is computed as  $\boldsymbol{M}_{t}/\sqrt{\boldsymbol{V}_{t}+\epsilon}$ and reshaped back into the original dimension of the gradient $\boldsymbol{G}_{t} \in \mathbb{R}^{n_1\times \ldots \times n_d}$ to update the weight $\boldsymbol{W}_{t}$.

\subsection{Time (Computation) Complexity of \optim} 
\label{sec:comput_complexity}
The time complexity of SMMF consists of two parts, i.e., square-matricization and decompression/compression. First, computing $\hat{n}$ and $\hat{m}$ for square-matricization (\Cref{alg:square_matrixlization}) is $\BIGO{\sqrt{N}}$, where $N$ is the number of elements in the momentum tensor. However, this computational overhead is negligible since $\hat{n}$ and $\hat{m}$ are calculated only once before starting model training (optimization). Next, the time complexity of decompression (\Cref{alg:matrix_decompression}) and compression (\Cref{alg:matrix_compression}) are both $\BIGO{N}$, which is asymptotically equivalent to existing memory-efficient optimizers, i.e., Adafactor and CAME. While taking a similar computational complexity to existing memory-efficient optimizers~\cite{adafactor,came}, \optim is able to save up to 96\% of memory, as shown in \Cref{sec:experiment}.
\section{Regret Bound Analysis}
\label{sec:regret}
We analyze the convergence of \optim by deriving the upper bound of the regret that indicates an optimizer's convergence~\cite{adam}. The regret $R(T)$ is defined as the sum of differences between two convex functions $f_t(\weight_t)$ and $f_t(\weight^*)$ for all $t \in [1, T]$, where $\VECTOR{w}^*$ is an optimal point.
\ninefont
\begin{equation}
    \label{eq:regret}
    R(T) = \sum_{t = 1}^T(f_t(\VECTOR{w}_t) - f_t(\VECTOR{w}^*))
\end{equation}
\tenfont
Since \optim factorizes (compresses) the momentum tensors, unlike Adam, we introduce some compression error terms in our analysis as follows. First, $\hat{\VECTOR{m}}_t$ and $\hat{\VECTOR{v}}_t$ are the decompressed and vectorized vectors of the $1^{st}$ and $2^{nd}$ momentum $\MATRIX{M}_t$ and $\MATRIX{V}_t$, containing $\VECTOR{e}_{m, t}$ and $\VECTOR{e}_{v, t}$, which denote compression errors of $\MATRIX{M}_t$ and $\MATRIX{V}_t$, respectively, i.e., $\VECTOR{e}_{m, t} = \hat{\VECTOR{m}}_t - \VECTOR{m}_t$ and $\VECTOR{e}_{v, t} = \hat{\VECTOR{v}}_t - \VECTOR{v}_t$. Similarly, we also define $\tilde{\VECTOR{e}}_{m, t}$, $\tilde{\VECTOR{e}}_{v, t}$, $\tilde{\VECTOR{g}}_{m, t}$, $\tilde{\VECTOR{g}}_{v, t}$, and $\hat{\VECTOR{g}}_{m, t}$. The detailed definitions are given in \Cref{lemma:compact_form_of_m_and_v_hat,lemma:m_t_with_g_tilde} in \Cref{proof:regret_bounding}.
\begin{theorem}
    \label{thm:regret_bounding}
    Let $\weight_t$ and $\VECTOR{v}_t$ be the vectorized $\WEIGHT_t$ and $\MATRIX{V}_t$, respectively, in \Cref{alg:brief_algorithm}, and $\eta_t = \eta / \sqrt{t}$, $\beta_1 = \beta_{1, 1}$, $\beta_2 = \beta_{2, 1}$, $\beta_{1, t} \leq \beta_1$ for all $t \in [1, T]$, and $\zeta_1 > 0$. We assume that $\lvert\lvert \weight_t - \weight^*\rvert\rvert_2 \leq D$, $\lvert\lvert \weight_k - \weight_l\rvert\rvert_{\infty} \leq D_{\infty}$, and all $\weight$ is in $\mathcal{F}$ where $D$ is the diameter of the feasible space $\mathcal{F}$. Furthermore, let $\beta_{1, t}$ and $\beta_{2, t}$ follow the following conditions, where the conditions (a) and (b) are from \cite{amsgrad}.
    \ninefont
    \begin{align*}
        \label{cond:01}
        & (a) ~\eta_{t - 1}\sqrt{\VECTOR{v}_{t, i}} \geq \eta_{t} \sqrt{\VECTOR{v}_{t - 1, i}} \\
        & (b) ~\dfrac{1}{\eta_t} \sqrt{\sum_{j = 1}^t \prod_{k = 1}^{t - j} \beta_{2, t - k + 1}(1 - \beta_{2, j} \tilde{\VECTOR{g}}_{v, j, i})} \geq \dfrac{1}{\zeta_2} \sqrt{\sum_{j = 1}^t \tilde{\VECTOR{g}}_{v, j, i}} \\ 
        & \quad\,for\,some\,\zeta_2 > 0\,and\,all\,t \in [:T]
    \end{align*}
    \tenfont
    Given the above conditions, the upper bound of the regret $R(T)$ on a convex function becomes:
    \ninefont
    \begin{align}
        R(T)
            &\leq \sum_{i = 1}^d \dfrac{D^2 \sqrt{\VECTOR{v}_{T, i}}}{2\eta_T (1 - \beta_1)}  + \sum_{t = 1}^T \sum_{i = 1}^d\dfrac{\beta_{1, t} D_{\infty}^2 \sqrt{\VECTOR{v}_{t, i}}}{2 \eta_{t} (1 - \beta_{1, t})} \\
            \nonumber
            & \quad +\dfrac{\zeta_1 \zeta_2 (1 + \beta_1) \sqrt{d}}{(1 - \beta_1)^3} \sqrt{\sum_{i = 1}^d \lvert \lvert \VECTOR{g}_{1:T, i} \rvert \rvert_2} ~ .
    \end{align}
    \tenfont
\end{theorem}
The full proof of \Cref{thm:regret_bounding} is provided in \Cref{proof:regret_bounding}. It shows that \optim has a similar regret bound ratio to AdamNC~\cite{amsgrad} (a variant of Adam), i.e., $\BIGO{\sqrt{T}}$ where $T$ is the total optimization steps, under the conditions (a) and (b). It allows \optim to perform consistent optimization comparable to Adam-based and existing memory-efficient optimizers, as shown in the experiment (\Cref{sec:experiment}). The conditions (a) and (b) can be satisfied by properly scheduling $\beta_{2, t}$~\cite{amsgrad}.
\section{Experiment}
\label{sec:experiment}
We implement the proposed \optim using PyTorch~\cite{pytorch}, which is available both on an GitHub and in \Cref{sec:code}. For evaluation, we compare the memory usage and optimization performance of \optim against four (memory-efficient) optimizers, i.e., Adam, Adafactor, SM3, and CAME, with two types of deep learning tasks: 1) CNN models for image tasks and 2) Transformer models for NLP tasks. The detailed experimental setups and training configurations are provided in \Cref{config:training_configurations}.

\noindent\textbf{CNN-based Models and Tasks.} We apply the five optimizers, including \optim, to two representative image tasks, i.e., image classification and object detection, and evaluate them by 1) training ResNet-50~\cite{resnet} and MobileNetV2~\cite{mobilenet} on CIFAR100~\cite{cifar} and ImageNet~\cite{imagenet}, and 2) training YOLOv5s and YOLOv5m~\cite{ultralytics2021yolov5} on COCO~\cite{lin2015microsoft}.

\noindent\textbf{Transformer-based Models and Tasks.} For NLP tasks, we train several Transformer-based models  from small scale to large scale with three training methods, i.e., full-training, pre-training, and fine-tuning. We 1) full-train Transformer-base and big models~\cite{transformer} on WMT32k~\cite{bojar2014findings_wmt32k}, 2) pre-train BERT~\cite{bert}, GPT-2~\cite{gpt2}, and T5~\cite{t5} on BookCorpus~\cite{bookcorpus} \& Wikipedia, and 3) fine-tune BERT, GPT-2, T5-small, and LLaMA-7b~\cite{llama1} on QNLI, MNLI, QQP, STSB, and MRPC~\cite{glue} datasets. To fine-tune LLaMA-7b, we use LoRA~\cite{lora}. Additionally, we train various Transformer models on other NLP tasks such as question-answering~\cite{squad,squadv2}, translation~\cite{wmt16}, and summarization~\cite{cnn_dailymail,xsum,ilpost}.  Due to the page limit, we provide the detailed experiment results in \Cref{sec:additional_fine_tuning_performance}.

\noindent\textbf{Optimization Time Measurement.} We measure the optimization time of five optimizers with the CNN and Transformer models, i.e., 1) MobileNetV2 and ResNet-50 on ImageNet, 2) Transformer-base and big on WMT32K.
\input{tables/memory_footprint_and_result_of_cnn_models}
\subsection{CNN-based Models and Tasks}
\label{subsec:cnn_models_training}

\textbf{Image Classification.} The first and second tables of \Cref{tab:memory_footprint_and_result_of_cnn_models} summarize the optimizer memory usage, end-to-end training (one-batch) memory usage, and top-1 classification accuracy of MobileNetV2 and ResNet-50 on CIFAR100 and ImageNet, respectively, showing that \optim achieves comparable accuracy among the five optimizers with the lowest memory footprint. For instance, for ResNet-50 on ImageNet, \optim substantially reduces the optimizer memory usage from 220 to 3.7 MiB (59x smaller) compared to Adafactor while achieving a higher accuracy (73.7\%). Its memory usage is also smaller than the other two memory-efficient optimizers, i.e., SM3 (99 vs. 3.7 MiB) and CAME (346 vs. 3.7 MiB). On the other hand, both Adafactor and CAME take more memory than not only \optim but also Adam due to the overhead of slicing the momentum tensor into multiple matrices for factorization based on the last two rank of the tensor corresponding to the size of a CNN kernel (i.e., $C^i_k\times C^o_k \times H_k \times W_k$). Since $H_k$ and $W_k$ are usually small, e.g., $H_k=W_k=1$, $3$ or $5$, whereas $C^i_k$ and $C^o_k$ are large, e.g., $512$, the memory reduction effect of both Adafactor and CAME becomes marginal in CNNs. \Cref{fig:cnn} (left) plots the top-1 validation accuracy of MobileNetV2 on ImageNet over training steps. 

\input{images/cnn_model_performance}
\noindent\textbf{Object Detection.} The third table of \Cref{tab:memory_footprint_and_result_of_cnn_models} summarizes the optimizer memory usage, end-to-end training (one-batch) memory usage, and the model performance metric (i.e., mAP50) of YOLOv5s and YOLOv5m on the COCO object detection task. \Cref{fig:cnn} (right) shows the mAP50 of YOLOv5s on COCO over training steps. Similar to image classification tasks, \optim achieves comparable mAP50, e.g., 59.6 in YOLOv5m and 54.1 in YOLOv5s, over all the four optimizers with the lowest memory up to 78x reduction, e.g., 267 vs. 3.4 MiB. This result implies that YOLOv5s (65 MiB) can be trained on an off-the-shelf memory-constrained device, e.g., Coral Dev Micro (64 MiB)~\cite{coral_devboard_micro}, along with other memory-efficient training methods such as gradient checkpointing~\cite{gradient_checkpointing}.

These experiment results show that \optim performs consistent and reliable optimization for both image classification and object detection tasks with different CNN models, taking the smallest memory compared to existing memory-efficient optimizers, i.e., Adafactor, SM3, and CAME. 
\input{tables/memory_footprint_and_result_of_full_training}

\subsection{Transformer-based Models and Tasks}
\label{subsec:transformer_models_training}
\textbf{Full-training.} As shown in \Cref{tab:memory_footprint_and_result_of_full_training}, \optim achieves comparable perplexity with up to 70x smaller optimizer memory when full-training (i.e., training models from scratch) both the Transformer-base and big models. Since \optim square-matricizes both the $1^{st}$ and $2^{nd}$ momentums and factorizes them, its memory usage is at least half lower than the other memory-efficient optimizers, i.e., Adafactor, SM3, and CAME. Given that most Transformer architectures consist of two-dimensional matrices, e.g., attention and linear layers, the memory reduction effect of SM3 that is good at compressing a high-rank tensor, becomes insignificant, making its memory usage similar to Adafactor and CAME. On the other hand, \optim can effectively reduce memory required to factorize a two-dimensional matrix with square-matricization, e.g., saving 69\% of memory for the embedding weight matrix of BERT, as the weight matrix in $\RSET^{30522\times768}$ becomes $\RSET^{5087\times4608}$. Although the square-matricization may spoil the gradient pattern of the Transformer's weight parameters~\cite{sm3}, \optim performs comparable optimization (e.g., 6.7 perplexity for WMT32k) using much less memory by fully utilizing the intact latest gradient before it is compressed. It is possible by taking the proposed $decompression$$\rightarrow$$compression$ scheme that first reflects the intact gradient pattern, if any, to the momentum and then performs factorization, unlike existing optimizers, e.g., Adafactor, SM3, and CAME, which apply the $compression$$\rightarrow$$decompression$ scheme. \Cref{fig:transformer_base_ppl} (left) shows the test perplexity of the Transformer-base model full-trained on WMT32k from scratch.

\input{tables/memory_footprint_and_result_of_pre_training}

\noindent\textbf{Pre-training.} \Cref{tab:memory_footprint_and_result_of_pre_training} shows the optimizer memory usage including $\MATRIX{S}_{\MATRIX{M}}$, end-to-end training (one-batch) memory usage, and the perplexity of BERT, GPT-2, and T5 pre-trained for the BookCorpus \& Wikipedia dataset. For GPT-2 pre-trained with Adafactor, we failed to obtain its perplexity since it diverged (NaN) even with multiple trials of various settings, e.g., different machines, hyper parameters, seeds, etc. On the other hand, \optim performs competitive optimization for all pre-training of BERT, GPT-2, and T5 using up to 60x lower optimizer memory. \Cref{fig:transformer_base_ppl} (right) shows that \optim (yellow) curtails the optimizer memory usage in the pre-training of BERT about 1.26 GiB (from 1.3 to 0.04 GiB) while exhibiting the similar test perplexity trajectory to CAME (purple) taking much more memory to maintain the similar optimization performance. Overall, Adafactor, CAME, and \optim show similar perplexity trajectories, confirming that \optim retains the optimization performance in pre-training of Transformers with the lowest memory, enabled by 1) square-matricization of any rank (shape) of momentums (e.g., a vector, matrix, tensor, etc.) and 2) the $decompression$$\rightarrow$$compression$ scheme that uses the uncompressed gradients at the current update step. They jointly allow it to pre-train Transformers aptly during a massive number of pre-training steps by persistently conducting effective and efficient optimization over a long time horizon.

That being said, \optim also exhibits some occasional loss spike~\cite{loss-spike-01, loss-spike-02} at the early steps of optimization (training), which stabilizes as the training proceeds. It is a well-known phenomenon that commonly occurs in the pre-training of many large language models~\cite{t5, gpt2} optimized with existing optimizers, such as Adam, Adafactor, SM3, and CAME. We discuss it in more detail in \Cref{sec:limitation}.

\noindent\textbf{Fine-tuning.} \Cref{tab:memory_footprint_and_result_of_fine_tuning} summarizes the optimization memory usage including $\MATRIX{S}_{\MATRIX{M}}$, end-to-end training(one-batch) memory usage, and the model performance of GPT-2, T5-small, and LLaMA-7b, which are fine-tuned for the QNLI, MNLI, QQP, STSB, and MRPC datasets from pre-trained models. As shown in the table, \optim achieves comparable model performance in the six datasets compared to the other four optimizers with the lowest memory usage. For instance, \optim provides similar accuracy (90.6\%) for GPT-2 on QQP compared to CAME, using much smaller optimizer and end-to-end training memory, respectively (16 vs. 489 MiB and 0.96 vs. 1.43 GiB). It demonstrates that \optim is also apt at fine-tuning Transformer models, which entails delicate and intricate updates of weight parameters, otherwise likely to degrade the learning performance for downstream tasks. In practice, it suggests that some Transformer models, such as T5, would be fine-tuned on a low-end device, e.g., Raspberry Pi Zero (0.5 GiB), with similar model performance to CAME (i.e., 83.0\% vs. 82.8\%), as \optim curtails the end-to-end training memory requirement of fine-tuning down to 0.47 GiB. On the other way around, \optim can scale up training of Transformers by enabling memory-efficient optimization of enormous Transformer models that require a gigantic amount of memory, e.g., hundreds of GiB. 

Similar to the CNN-based models and tasks, the experimental results of Transformers demonstrate that \optim steadily performs competitive optimization for a wide range of Transformer models, tasks, and training methods (i.e., full-, pre-, and fine-tuning) with the smallest memory usage.

\input{tables/memory_footprint_and_result_of_fine_tuning}
\input{tables/time}
\input{images/transformer_base_ppl}
\subsection{Optimization Time Measurement}
\label{subsec:optimization_time}
\Cref{tab:time} shows the optimization time measured for a single training step of the five optimizers when training MobileNetV2 and ResNet-50 on ImageNet, and Transformer-base and big on WMT32k. Except for Adam, the four optimizers take similar optimization times, where \optim takes a little more time than the other three. That is because it trades off the memory space and optimization time, i.e., the time required for square-matricization and the sign matrix operations for the $1^{st}$ momentum (\Cref{alg:matrix_decompression,alg:matrix_compression}). However, \optim offers huge memory reduction compared to the amount of the increased optimization time, e.g., 7.9x memory reduction vs. 1.2x time increase for Adam and \optim with the 8-bit format $\boldsymbol{S_{M}}$ in \Cref{alg:matrix_decompression,alg:matrix_compression} applied to Transformer-big on WMT32k, and 7.8x vs. 1.6x for Adam and \optim with  $\boldsymbol{S_{M}}$ applied to ResNet-50 on ImageNet.
\section{Limitations and Discussions}
\label{sec:limitation}

\textbf{Overhead of Binary Signs.} 
It is not easy to effectively factorize the $1^{st}$ momentum $\MATRIX{M}$ having negative elements. While \optim circumvents this by storing the binary sign matrix (1-bit) that is much smaller than the original matrix (32-bit), there are some methods that can further reduce the memory required to store the sign matrix. For instance, Binary Matrix Factorization \cite{bmf2,bmf3} can be employed to factorize the binary matrix into two lower-rank matrices, reducing the memory space for storing the binary matrix with a high restoration rate.

\noindent\textbf{Optimization Time.} \Cref{sec:comput_complexity} finds that the time complexity of \optim is similar to existing memory-efficient optimizers, e.g., Adafactor, showing that \optim effectively saves a significant amount of memory (up to 96\%) compared to them with slightly increased optimization time.

\noindent\textbf{Loss Spike at Initial Training Steps.} We observe some loss spike at the initial training steps, especially in Transformer models, which is a well-known issue commonly observed in other works~\cite{loss-spike-01,loss-spike-02} and many optimizers, i.e., Adam (with the bias correction term), Adafactor, SM3, and CAME. With appropriate hyper-parameter tuning, e.g., learning rate and weight-decay, it can be stabilized as the training proceeds, like other optimizers.

\noindent\textbf{Extremely Large Models and Other Tasks.} 
Due to the limited computing resources, we have not been able to experiment \optim with extremely large models, e.g, GPT-4~\cite{gpt4}, LLaMA-2 70B~\cite{llama2}, and diffusion models~\cite{ldm}. From our theoretical study and empirical result, we expect \optim to perform competitive optimization with them. We hope to have a chance to test \optim with them.

\section{Conclusion}
We introduce \optim, a memory-efficient optimizer that decreases the memory requirement of adaptive learning rate optimization methods, e.g., Adam and Adafactor. Given arbitrary-rank momentum tensors, \optim reduces the optimization memory usage through the proposed square-matricization and matrix compression of the first and second momentums. The empirical evaluation shows that \optim reduces up to 96\% of optimization memory when compared to existing memory-efficient optimizers, i.e., Adafactor, SM3, and CAME, with competitive optimization performance on various models (i.e., CNNs and Transformers) and tasks (i.e., image classification, object detection, and NLP tasks over full-training, pre-training, and fine-tuning). Our analysis of regret bound proves that \optim converges in a convex function with a similar bound of AdamNC.
\section*{Acknowledgement}
This work was partly supported by Institute of Information \& communications Technology Planning \& Evaluation(IITP) grant funded by the Korea goverment(MSIT) (No.RS-2024-00508465) and Institute of Information \& communications Technology Planning \& Evaluation(IITP) grant funded by the Korea government(MSIT) (No.RS-2020-II201336, Artificial Intelligence Graduate School Program(UNIST)).


\onecolumn
\appendix
\input{sections/appendix}
\bibliography{aaai25}

\end{document}

%% file: algorithms/brief_algorithm.tex
\begin{algorithm} [!tb]
    \small
    \caption{Overall \optim applied to each layer. The elements of $\VECTOR{r}$, $\VECTOR{c}$, $\MATRIX{M}$, $\MATRIX{V}$, and $\MATRIX{S}$ are initially set to zeros.}
    \label{alg:brief_algorithm}
    \begin{algorithmic}
        \STATE {\bfseries Input:} Step $t$, total step $T$, model $f(\cdot)$ with rank-$d$ weight tensor $\WEIGHT_t \in \RSET^{n_1 \times \ldots \times n_d}$, learning-rate $\eta_t$, regularization constant $\epsilon$, $1^{st}$ and $2^{nd}$ momentum hyper-parameters $\beta_{1,t}$ and $\beta_{2,t}$.
        \FOR{$t = 1$ {\bfseries to} $T$}
            \STATE $\MATRIX{G}_t = \nabla f(\WEIGHT_{t-1})$
            \STATE $\bar{\MATRIX{G}}_t = $ Square-Matricization($\MATRIX{G}_t$, $n_1 \ldots n_d$) ~~\quad\qquad [Algo~\ref{alg:square_matrixlization}]
            \STATE $\hat{\MATRIX{M}}_{t-1} = $ Decompression($\VECTOR{r}_{\MATRIX{M}_{t-1}}$, $\VECTOR{c}_{\MATRIX{M}_{t-1}}$, $\MATRIX{S}_{\MATRIX{M}_{t-1}}$)  [Algo~\ref{alg:matrix_decompression}]
            \STATE $\hat{\MATRIX{V}}_{t-1} = $ Decompression($\VECTOR{r}_{\MATRIX{V}_{t-1}}$, $\VECTOR{c}_{\MATRIX{V}_{t-1}}$, $\MATRIX{1}$) ~\quad\qquad [Algo~\ref{alg:matrix_decompression}]
            \STATE $\MATRIX{M}_{t} = \beta_{1, t} \hat{\MATRIX{M}}_{t-1} + (1 - \beta_{1, t})\bar{\MATRIX{G}}_t$
            \STATE $\MATRIX{V}_{t} = \beta_{2, t} \hat{\MATRIX{V}}_{t-1} + (1 - \beta_{2, t})\bar{\MATRIX{G}}_t^2$
            \STATE ($\VECTOR{r}_{\MATRIX{M}_t}$, $\VECTOR{c}_{\MATRIX{M}_t}$, $\MATRIX{S}_{\MATRIX{M}_t}$) = Compression($\MATRIX{M}_{t}$) ~~~\qquad\qquad [Algo~\ref{alg:matrix_compression}]
            \STATE ($\VECTOR{r}_{\MATRIX{V}_t}$, $\VECTOR{c}_{\MATRIX{V}_t}$, .) = Compression($\MATRIX{V}_{t}$) ~~\quad\qquad\qquad\qquad [Algo~\ref{alg:matrix_compression}]
            \STATE $\MATRIX{U}$ = Reshape($\MATRIX{M}_t/\sqrt{\MATRIX{V}_t + \epsilon}$, $n_1 \ldots n_d$)
            \STATE $\WEIGHT_t$ = $\WEIGHT_{t-1} - \eta_t\MATRIX{U}$
        \ENDFOR     
    \end{algorithmic}
\end{algorithm}

%% file: algorithms/square_matrixlization.tex
\begin{algorithm}[!tb]
\small
    \caption{Square-Matricization. It needs to be calculated only once before starting model training (optimization).}
    \label{alg:square_matrixlization}
    \begin{algorithmic}
        \STATE {\bfseries Input:} Rank-$d$ tensor $\MATRIX{G} \in \RSET^{n_1\times \ldots \times n_d}$ and the length of each axis $n_1 \ldots n_d$
        \STATE {\bfseries Output:} Reshaped matrix $\bar{\MATRIX{G}} \in \RSET^{\hat{n} \times \hat{m}}$
        \STATE $\SCALAR{N} = \prod_{i=1}^d n_i$; $s = \lfloor \sqrt{\SCALAR{N}} \rfloor$

        \FOR{$i=s$ {\bfseries to} $1$}
            \IF{($\SCALAR{N}$ mod $i$) == 0}
                \STATE $\bar{\MATRIX{G}} =$ Reshape($\MATRIX{G}$, $\hat{n}$, $\hat{m}$) where $\hat{n} = \SCALAR{N} / i $ and $\hat{m} = i$; break
            \ENDIF
        \ENDFOR
    \end{algorithmic}
\end{algorithm}

%% file: algorithms/matrix_decompression.tex
\begin{algorithm}[!tb]
    \small
    \caption{Decompression.}
    \label{alg:matrix_decompression}
    \begin{algorithmic}
        \STATE {\bfseries Input:} Factorized vectors $\VECTOR{r} \in \RSET^{\hat{n} \times 1}$ and $\VECTOR{c} \in \RSET^{1 \times \hat{m}}$, and a binary sign matrix $\MATRIX{S} \in \BSET^{\hat{n} \times \hat{m}}$. [$\SCALAR{M}_{i, j}$ is $i^{th}$ row and $j^{th}$ column element in a matrix $\MATRIX{M} \in \RSET^{\hat{n} \times \hat{m}}$]
        \STATE {\bfseries Output:} Decompressed matrix $\MATRIX{M} \in \RSET^{\hat{n} \times \hat{m}}$
        \STATE $\MATRIX{M} = \VECTOR{r} \otimes \VECTOR{c}$ \quad\quad\quad [$\otimes$ is the outer product operator]
        \STATE $\SCALAR{M}_{i, j} = 
            \begin{cases}
                \SCALAR{M}_{i, j}, & $if $\SCALAR{S}_{i, j} = 1 \\
                -\SCALAR{M}_{i, j}, & $otherwise$
            \end{cases}
        $
    \end{algorithmic}
\end{algorithm}

%% file: algorithms/matrix_compression.tex
\begin{algorithm}[!tb]
\small
    \caption{Compression. $\VECTOR{1}_d$ is a vector that all elements are one and its length is $d$.}
    \label{alg:matrix_compression}
    \begin{algorithmic}
        \STATE {\bfseries Input:} Matrix $\MATRIX{M} \in \RSET^{\hat{n} \times \hat{m}}$ to be factorized
        \STATE {\bfseries Output:} Factorized vectors $\VECTOR{r} \in \RSET^{\hat{n} \times 1}$ and $\VECTOR{c} \in \RSET^{1 \times \hat{m}}$, and binary sign matrix $\MATRIX{S} \in \BSET^{\hat{n} \times \hat{m}}$
        \STATE $\SCALAR{S}_{i, j}=
            \begin{cases}
                1,& if \SCALAR{M}_{i,j} \geq 0\\
                0,& otherwise
            \end{cases}
            ,\quad(\VECTOR{r},\VECTOR{c}) = (|\MATRIX{M}|\VECTOR{1}_{\hat{m}}, \VECTOR{1}_{\hat{n}}^{\top} |\MATRIX{M}|)$
        \STATE $\VECTOR{r} =        
            \begin{cases}
            \VECTOR{r}/(\VECTOR{1}_{\hat{n}}^{\top}\VECTOR{r}),& $if $\hat{n} \le \hat{m} \\
                \VECTOR{r}, & $otherwise$
            \end{cases}
            ,\quad\VECTOR{c} =
            \begin{cases}
                \VECTOR{c}/(\VECTOR{c}\VECTOR{1}_{\hat{m}}), &$if $ \hat{n} > \hat{m} \\
                \VECTOR{c}, &$otherwise$
            \end{cases}$     
    \end{algorithmic}
\end{algorithm}

%% file: tables/memory_footprint_and_result_of_cnn_models.tex
\begin{table}[!tb]
    \setlength\tabcolsep{1.7pt}
    \small
    \centering
        \begin{tabular}{c|c@{(}r@{, }r@{)}|c@{\:(}r@{, }r@{)\:}|c@{(}r@{, }r@{)}|c@{(}r@{, }r@{)}|c@{\:(}r@{,}r@{)\:}}
        \toprule
        \multicolumn{16}{c}{\ninefont \textbf{CNN Models and Tasks}}    \\
        \multicolumn{16}{c}{\ninefont (Optimizer and End-to-End Memory [MiB]), Model Performance}   \\
        \midrule
        \ninefont \textbf{Dataset}  & \multicolumn{15}{|c}{(1) CIFAR100 (\textbf{Image Classification})}   \\
        \hline
        \ninefont \textbf{Model}    & \multicolumn{3}{c|}{\ninefont \textbf{Adam}}  & \multicolumn{3}{c|}{\ninefont \textbf{Adafactor}} & \multicolumn{3}{c|}{\ninefont \textbf{SM3}}   & \multicolumn{3}{c|}{\ninefont \textbf{CAME}}  & \multicolumn{3}{c}{\ninefont \textbf{\optim}} \\
        \hline
        \ninefont MobileNet && \ninefont 18 & \ninefont 36  && \ninefont 26 & \ninefont 43  && \ninefont 9  & \ninefont 27  && \ninefont 43 & \ninefont 60  && \ninefont \textbf{0.7}   & \ninefont \textbf{19}  \\
        \ninefont (V2)      & \multicolumn{3}{c}{\ninefont 73.6}   & \multicolumn{3}{|c}{\ninefont 69.4}    & \multicolumn{3}{|c}{\ninefont 70.0}    & \multicolumn{3}{|c}{\ninefont 66.2}    & \multicolumn{3}{|c}{\ninefont 74.1}    \\
        \hline
        \ninefont ResNet && \ninefont 184    & \ninefont 366 && \ninefont 215    & \ninefont 397 && \ninefont 93 & \ninefont 227 && \ninefont 340    & \ninefont 526 && \ninefont \textbf{3.5}   & \ninefont \textbf{185}    \\ 
        \ninefont (50)  & \multicolumn{3}{c}{\ninefont 72.4}   & \multicolumn{3}{|c}{\ninefont 72.9}    & \multicolumn{3}{|c}{\ninefont 73.8}    & \multicolumn{3}{|c}{\ninefont 67.2}    & \multicolumn{3}{|c}{\ninefont 74.5}    \\
        \hline
        \hline
        \ninefont \textbf{Dataset}  & \multicolumn{15}{|c}{(2) ImageNet (\textbf{Image Classification})}   \\
        \hline
        \ninefont MobileNet && \ninefont 27 & \ninefont 54  && \ninefont 30 & \ninefont 58  && \ninefont 14 & \ninefont 41  && \ninefont 47 & \ninefont 75  && \ninefont \textbf{0.8}   & \ninefont \textbf{28}  \\
        \ninefont (V2)      & \multicolumn{3}{c}{\ninefont 68.6}   & \multicolumn{3}{|c}{\ninefont 69.3}    & \multicolumn{3}{|c}{\ninefont 67.6}    & \multicolumn{3}{|c}{\ninefont 69.5}    & \multicolumn{3}{|c}{\ninefont 69.5}    \\
        \hline
        \ninefont ResNet && \ninefont 195    & \ninefont 394 && \ninefont 220    & \ninefont 419 && \ninefont 99 & \ninefont 298 && \ninefont 346    & \ninefont 546 && \ninefont \textbf{3.7}   & \ninefont \textbf{197}    \\ 
        \ninefont (50)  & \multicolumn{3}{c}{\ninefont 73.7}   & \multicolumn{3}{|c}{\ninefont 69.5}    & \multicolumn{3}{|c}{\ninefont 75.8}    & \multicolumn{3}{|c}{\ninefont 72.3}    & \multicolumn{3}{|c}{\ninefont 73.7}    \\
        \hline
        \hline
        \ninefont \textbf{Dataset}  & \multicolumn{15}{|c}{(3) COCO (\textbf{Object Detection})}   \\
        \hline
        \ninefont YOLO && \ninefont 57 & \ninefont 121  && \ninefont 61 & \ninefont 92  && \ninefont 28 & \ninefont 92  && \ninefont 94 & \ninefont 159  && \ninefont \textbf{1.4}   & \ninefont \textbf{65}  \\
        \ninefont (v5s)      & \multicolumn{3}{c}{\ninefont 52.7}   & \multicolumn{3}{|c}{\ninefont 53.6}    & \multicolumn{3}{|c}{\ninefont 50.3}    & \multicolumn{3}{|c}{\ninefont 53.3}    & \multicolumn{3}{|c}{\ninefont 54.1}    \\
        \hline
        \ninefont YOLO && \ninefont 168    & \ninefont 340 && \ninefont 174    & \ninefont 258 && \ninefont 84 & \ninefont 260 && \ninefont 267    & \ninefont 438 && \ninefont \textbf{3.4}   & \ninefont \textbf{176}    \\ 
        \ninefont (v5m)  & \multicolumn{3}{c}{\ninefont 58.0}   & \multicolumn{3}{|c}{\ninefont 58.8}    & \multicolumn{3}{|c}{\ninefont 57.0}    & \multicolumn{3}{|c}{\ninefont 59.3}    & \multicolumn{3}{|c}{\ninefont 59.6}    \\
        \hline
        \end{tabular}
    \caption{\textbf{(First and second tables) Image classification}: the optimizer memory usage (MiB) including the binary sign matrix $\MATRIX{S}_{\MATRIX{M}}$, end-to-end training (one-batch) memory usage (MiB) at 100 iterations, and top-1 validation accuracy of MobileNetV2 and ResNet-50 on CIFAR100 and ImageNet. \textbf{(Third table) Object detection}: the optimizer memory usage (MiB) including $\MATRIX{S}_{\MATRIX{M}}$, end-to-end training (one-batch) memory usage (MiB) at 100 iterations, and validation mAP50 of YOLO (v5s and v5m) on COCO.}    \label{tab:memory_footprint_and_result_of_cnn_models}
\end{table}

%% file: images/cnn_model_performance.tex
\begin{figure}[!tb]
    \centering
     \begin{subfigure}[b]{0.49\columnwidth}
        \centering
        \centerline{\includegraphics[width=\columnwidth]{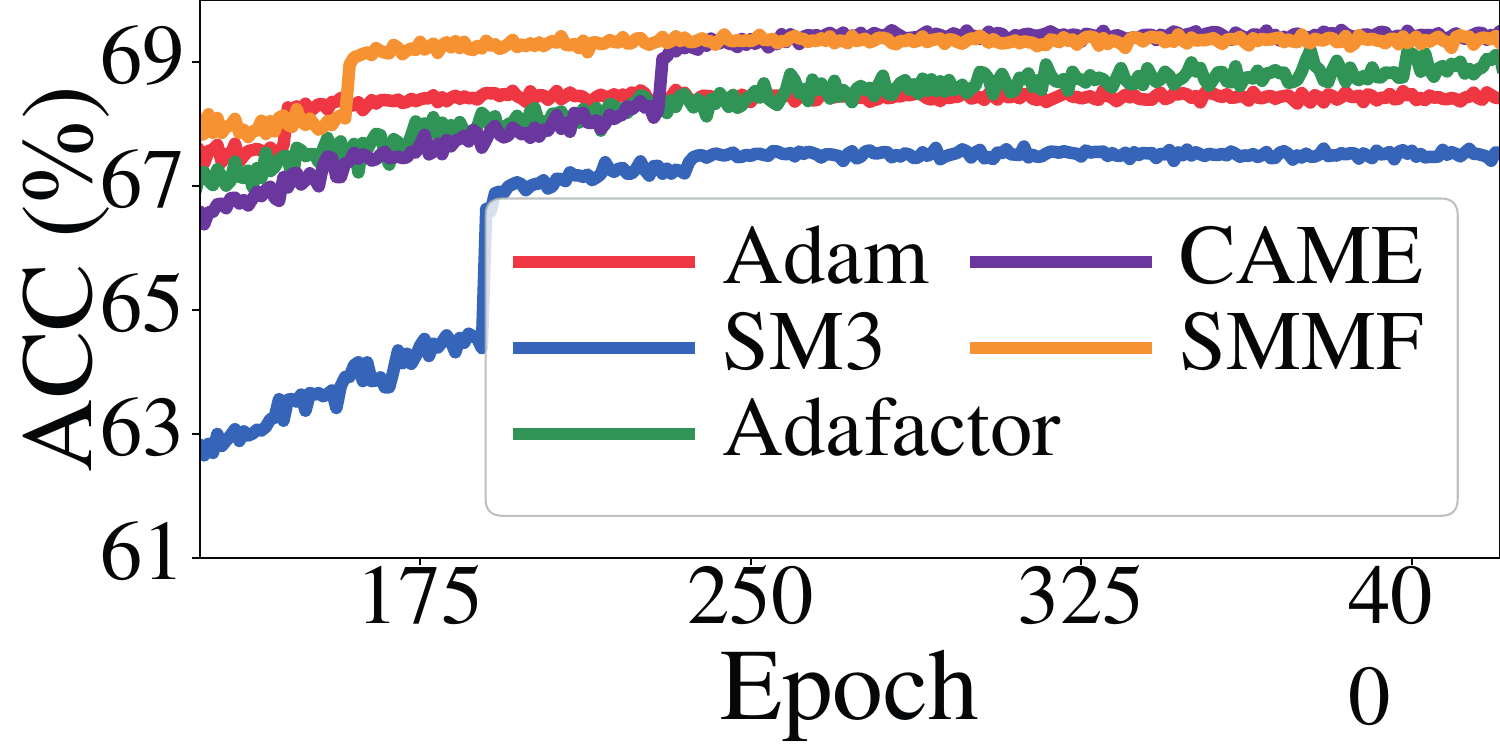}}
        \label{fig:cnn_left}
    \end{subfigure}
    \begin{subfigure}[b]{0.49\columnwidth}
        \centering
        \centerline{\includegraphics[width=\columnwidth]{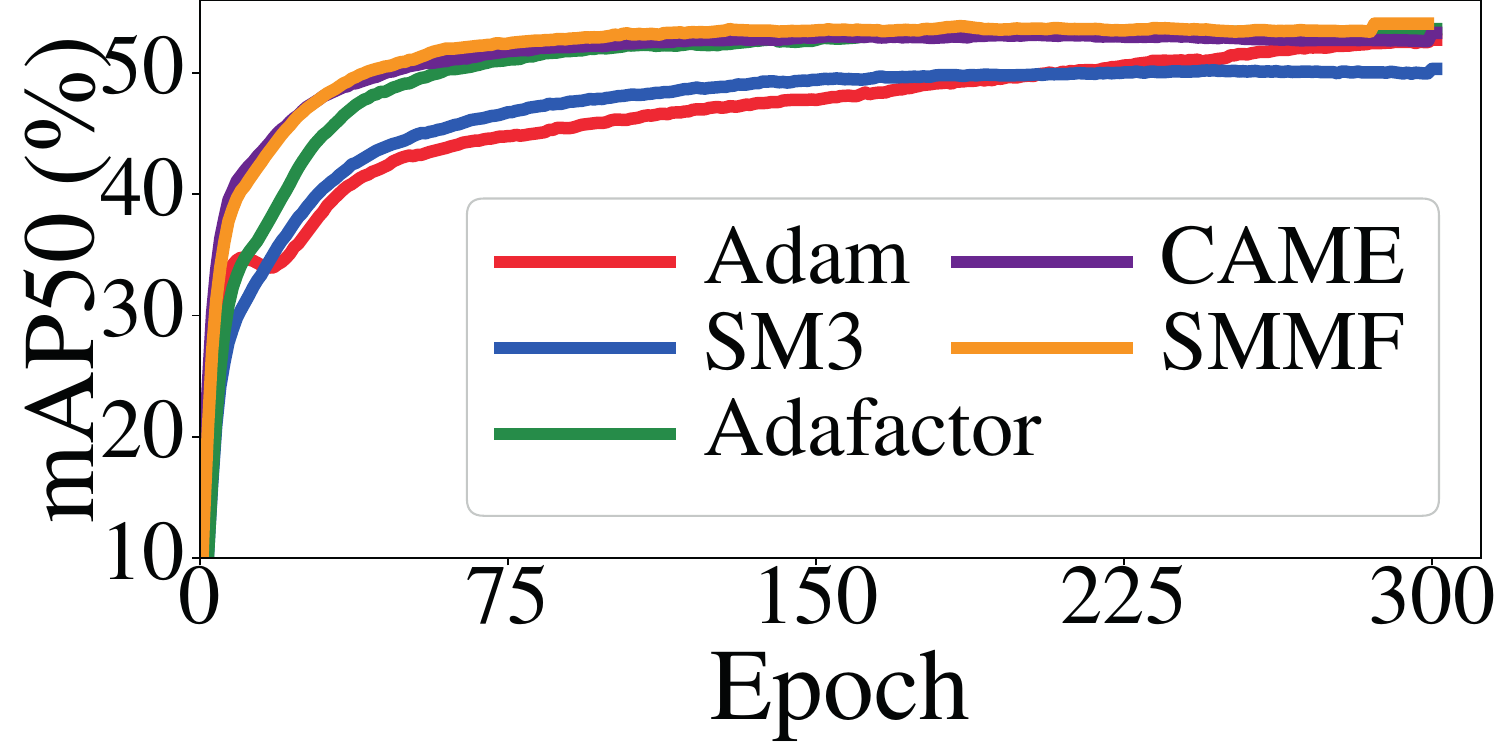}}
        \label{fig:cnn_right}
    \end{subfigure}
    \caption{\textbf{(Left)} The validation top-1 accuracy of MobileNetV2 on ImageNet. \textbf{(Right)} The validation mAP50 of YOLOv5s on COCO of the five optimizers.}
    \label{fig:cnn}
\end{figure}

%% file: tables/memory_footprint_and_result_of_full_training.tex
\begin{table}[!tb]
    \setlength\tabcolsep{1.0pt}
    \centering
    \begin{tabular}{c|c@{\ninefont (}r@{\ninefont , }r@{\ninefont)}|c@{\ninefont (}r@{\ninefont , }r@{\ninefont)\;\;}|c@{\ninefont (}r@{\ninefont , }r@{\ninefont)}|c@{(\ninefont }r@{\ninefont , }r@{\ninefont)}|c@{\ninefont (}r@{\ninefont , }r@{\ninefont)\:}}
         \toprule
         \multicolumn{16}{c}{\ninefont \textbf{Transformer Models and Tasks (Full-Trining})} \\
         \multicolumn{16}{c}{\ninefont (Optimizer and End-to-End Memory [GiB]), Model Performance} \\    
        \midrule
        \ninefont \textbf{Dataset} & \multicolumn{15}{|c}{\ninefont WMT32k} \\
        \hline
        \ninefont \textbf{Model}    &   \multicolumn{3}{c|}{\ninefont \textbf{Adam}}   &   \multicolumn{3}{c|}{\ninefont \textbf{Adafactor}}  &   \multicolumn{3}{c|}{\ninefont \textbf{SM3}}    &   \multicolumn{3}{c|}    {\ninefont \textbf{CAME}}   &   \multicolumn{3}{c}{\ninefont \textbf{\optim}}  \\
    
        \bottomrule
        \ninefont Transformer   && \ninefont 0.7    & \ninefont 1.4 &\;\;& \ninefont 0.4    & \ninefont  1.1       && \ninefont 0.4          & \ninefont 1.1       &\:&\ninefont  0.4          & \ninefont 1.1       &\:& \ninefont \textbf{\:.01}          & \textbf{\ninefont 0.8}        \\
        \ninefont (base)        & \multicolumn{3}{c|}{\ninefont 6.6}  & \multicolumn{3}{c|}{\ninefont 6.6}  & \multicolumn{3}{c|}{\ninefont 7.8}  & \multicolumn{3}{c|}{\ninefont 6.6}  & \multicolumn{3}{c}{\ninefont 6.7}  \\
        \hline
        \ninefont Transformer   && \ninefont 2.1    & \ninefont 4.2 &\;\;& \ninefont 1.1    & \ninefont 3.2       && 1.1         & \ninefont 3.2       && \ninefont 1.1         & \ninefont 3.2       &\:& \textbf{\ninefont \:.04}         & \textbf{\ninefont 2.1}       \\
        \ninefont (big)                 & \multicolumn{3}{c|}{\ninefont 6.9}  & \multicolumn{3}{c|}{\ninefont 6.6}  & \multicolumn{3}{c|}{\ninefont 7.9}  & \multicolumn{3}{c|}{\ninefont 7.5}  & \multicolumn{3}{c}{\ninefont 6.8}  \\
        \hline
    \end{tabular}
    \caption{\textbf{Full-training}: the optimizer memory usage (GiB), including the $\MATRIX{S}_{\MATRIX{M}}$, end-to-end training (one-batch) memory usage (GiB) at 100 iterations, and test perplexity of the Transformer-base and big models on WMT32k.
    }
    \label{tab:memory_footprint_and_result_of_full_training}
\end{table}

%% file: tables/memory_footprint_and_result_of_pre_training.tex
\begin{table}[!tb]
    \setlength\tabcolsep{1.0pt}
    \centering
    \begin{tabular}{c|c@{\ninefont (}r@{\ninefont , }r@{\ninefont)}|c@{\ninefont (}r@{\ninefont , }r@{\ninefont)\;\;}|c@{\ninefont (}r@{\ninefont , }r@{\ninefont)}|c@{(\ninefont }r@{\ninefont , }r@{\ninefont)}|c@{\ninefont (}r@{\ninefont , }r@{\ninefont)\:}}
         \toprule
         \multicolumn{16}{c}{\ninefont \textbf{Transformer Models and Tasks (Pre-Training)}} \\
         \multicolumn{16}{c}{\ninefont (Optimizer and End-to-End Memory [GiB]), Model Performance} \\
        
        \midrule
        \ninefont \quad\textbf{Dataset}\quad\quad & \multicolumn{15}{|c}{\ninefont Book Corpus \& Wikipedia} \\
        \hline
        \ninefont \textbf{Model}    &   \multicolumn{3}{c|}{\ninefont \textbf{Adam}}   &   \multicolumn{3}{c|}{\ninefont \textbf{Adafactor}}  &   \multicolumn{3}{c|}{\ninefont \textbf{SM3}}    &   \multicolumn{3}{c|}    {\ninefont \textbf{CAME}}   &   \multicolumn{3}{c}{\ninefont \textbf{\optim}}  \\
        \hline
        \multirow{2}{*}{\ninefont BERT} && \ninefont 2.5         & \ninefont 6.3       &\;\;& \ninefont 1.3         & \ninefont 5.0       && \ninefont 1.3         &\ninefont  5.0       && \ninefont 1.3         & \ninefont 5.0      &\:& \textbf{\ninefont \:.04}         & \textbf{\ninefont 3.8}       \\
        & \multicolumn{3}{c|}{\ninefont 16.1}  & \multicolumn{3}{c|}{\ninefont 30.6} & \multicolumn{3}{c|}{\ninefont 27.5} & \multicolumn{3}{c|}{\ninefont 20.1} & \multicolumn{3}{c}{\ninefont 20.4} \\
        \hline
        \multirow{2}{*}{\ninefont GPT-2} && \ninefont 2.6         & \ninefont 6.7       &\;\;& \ninefont 1.3         & \ninefont 5.3       && \ninefont 1.3         & \ninefont 5.3       && \ninefont 1.3         & \ninefont 5.3      &\:& \textbf{\ninefont \:.04}         & \textbf{\ninefont 4.0}       \\
        &   \multicolumn{3}{c|}{\ninefont 19.2} & \multicolumn{3}{c|}{\ninefont NaN} & \multicolumn{3}{c|}{\ninefont 19.4} & \multicolumn{3}{c|}{\ninefont 19.1} & \multicolumn{3}{c}{\ninefont 19.2} \\ 
        \hline 
        \multirow{2}{*}{\ninefont T5}   && \ninefont 1.7         & \ninefont 4.2       &\;\;& \ninefont 0.8         & \ninefont 3.4       && \ninefont 0.8         & \ninefont 3.4       && \ninefont 0.9         & \ninefont 3.4      &\:& \ninefont \textbf{\:.03}        & \ninefont \textbf{2.5}       \\
        & \multicolumn{3}{c|}{\ninefont 2.6} & \multicolumn{3}{c|}{\ninefont 2.6}  & \multicolumn{3}{c|}{\ninefont 2.8} & \multicolumn{3}{c|}{\ninefont 2.6} & \multicolumn{3}{c}{\ninefont 2.6} \\
        \hline                    
    \end{tabular}
    \caption{\textbf{Pre-training}: the optimizer memory usage (GiB), including the $\MATRIX{S}_{\MATRIX{M}}$, end-to-end training (one-batch) memory usage (GiB) at 100 iterations, and test perplexity of BERT, GPT-2, and T5. We use Adam without the bias correction term to obtain lower perplexity.}
    \label{tab:memory_footprint_and_result_of_pre_training}
\end{table}

%% file: tables/memory_footprint_and_result_of_fine_tuning.tex
\begin{table*}[!tb]
    \setlength\tabcolsep{5pt}
    \begin{center}
            \begin{tabular}{l||c|c@{\ninefont (}r@{\ninefont , }r@{\ninefont ) }r|c@{\ninefont (}r@{\ninefont , }r@{) }r|c@{\ninefont (}r@{\ninefont , }r@{\ninefont ) }r|c@{\ninefont (}r@{\ninefont , }r@{\ninefont ) }r|c@{\ninefont (}r@{\ninefont , }r@{\ninefont ) }r}
            \toprule
            \multicolumn{22}{c}{\textbf{Transformer Models and Tasks (Fine-Tuning)}} \\
            \multicolumn{22}{c}{(Optimizer Memory [MiB] and End-to-End Memory [GiB]), Model Performance} \\
            \toprule
            \textbf{\ninefont Optimizer} &  \textbf{\ninefont Model}       &   \multicolumn{4}{c|}{\textbf{\ninefont QNLI (ACC)}}    &   \multicolumn{4}{c|}{\textbf{\ninefont MNLI (ACC)}}    &   \multicolumn{4}{c|}{\textbf{\ninefont QQP (ACC)}}    &   \multicolumn{4}{c|}{\textbf{\ninefont STSB (Pearson)}
            }    &   \multicolumn{4}{c}{\textbf{\ninefont MRPC (ACC)}} \\
                        
            \bottomrule
            \ninefont Adam    &           && \ninefont    957 & \ninefont    1.89    & \ninefont    84.5    && \ninefont   952 & \ninefont   1.89    & \ninefont   72.4    && \ninefont   973 & \ninefont   1.89    & \ninefont   86.4    && \ninefont   962 & \ninefont   1.89    & \ninefont   83.2    && \ninefont   861 & \ninefont   1.89    & \ninefont   81.4 \\
            \ninefont Adafactor&          && \ninefont   478 & \ninefont   1.43    & \ninefont   74.7    && \ninefont   478 & \ninefont   1.43    & \ninefont   71.7    && \ninefont   488 & \ninefont   1.43    & \ninefont   80.1    && \ninefont   481 & \ninefont   1.43    & \ninefont   84.4    && \ninefont   481 & \ninefont   1.43    & \ninefont   82.6 \\
            \ninefont SM3     & \ninefont   GPT-2    && \ninefont   478 & \ninefont   1.43    & \ninefont   88.0    && \ninefont   478 & \ninefont   1.43    & \ninefont   81.1    && \ninefont   487 & \ninefont   1.43    & \ninefont   88.8    && \ninefont   481 & \ninefont   1.43    & \ninefont   84.1    && \ninefont   481 & \ninefont   1.43    & \ninefont   83.3\\
            \ninefont CAME    &           && \ninefont   468 & \ninefont   1.43    & \ninefont   88.6    && \ninefont   479 & \ninefont   1.43    & \ninefont   81.9    && \ninefont   489 & \ninefont   1.43    & \ninefont   90.6    && \ninefont   481 & \ninefont   1.43    & \ninefont   86.4    && \ninefont   478 & \ninefont   1.43    & \ninefont   83.3 \\
            \ninefont \optim  &           &&   \textbf{\ninefont 16} &   \textbf{\ninefont 0.96}    & \ninefont   88.9    &&   \textbf{\ninefont 16} &   \textbf{\ninefont 0.96}    & \ninefont   82.2    &&   \textbf{\ninefont 16} &   \textbf{\ninefont 0.96}   & \ninefont   90.6    &&   \textbf{\ninefont 16} &   \textbf{\ninefont 0.96}    & \ninefont   83.8    &&   \textbf{\ninefont 16} &   \textbf{\ninefont 0.96}   & \ninefont   81.6  \\
            \hline
            \ninefont Adam    &           && \ninefont   464 & \ninefont   0.92    & \ninefont   88.4    && \ninefont   464 & \ninefont   0.92    & \ninefont   77.5    && \ninefont   434 & \ninefont   0.92    & \ninefont   86.8    && \ninefont   456 & \ninefont   0.92     & \ninefont   84.7    && \ninefont   464 & \ninefont   0.92    & \ninefont   77.5 \\
            \ninefont Adafactor&          && \ninefont   233 & \ninefont   0.70    & \ninefont   90.1    && \ninefont   233 & \ninefont   0.70    & \ninefont   80.3    && \ninefont   233 & \ninefont   0.70    & \ninefont   88.7    && \ninefont   233 & \ninefont   0.70     & \ninefont   87.9    && \ninefont   233 & \ninefont   0.70    & \ninefont   80.3 \\
            \ninefont SM3     & \ninefont   T5      && \ninefont   233 & \ninefont   0.70    & \ninefont   88.8    && \ninefont   233 & \ninefont   0.70    & \ninefont   79.4    && \ninefont   233 & \ninefont   0.70    & \ninefont   88.1    && \ninefont   233 & \ninefont   0.70     & \ninefont   79.4    && \ninefont   233 & \ninefont   0.70    & \ninefont   79.4\\
            \ninefont CAME    & \ninefont  (small)  && \ninefont   233 & \ninefont   0.70    & \ninefont   90.7    && \ninefont   234 & \ninefont   0.70    & \ninefont   83.0    && \ninefont   233 & \ninefont   0.70    & \ninefont   90.4    && \ninefont   233 & \ninefont   0.70    & \ninefont   87.5    && \ninefont   234 & \ninefont   0.70    & \ninefont   83.0\\
            \ninefont \optim &           && \ninefont    \textbf{\ninefont 8} & \ninefont   \textbf{\ninefont 0.47}    & \ninefont   90.6    &&\textbf{\ninefont 8} & \ninefont   \textbf{\ninefont 0.47}    & \ninefont   82.8    && \ninefont    \textbf{\ninefont 8} & \ninefont   \textbf{\ninefont 0.47}    & \ninefont   90.2    && \ninefont    \textbf{\ninefont 8} & \ninefont   \textbf{\ninefont 0.47}    & \ninefont   84.7    && \ninefont    \textbf{\ninefont 8} & \ninefont   \textbf{\ninefont 0.47}    & \ninefont   82.8 \\
            \hline
            \ninefont Adam      &           && \ninefont 153    & \ninefont 24.9    & \ninefont 93.0    && \ninefont 153    & \ninefont 24.9    & \ninefont 87.5    && \ninefont 153    & \ninefont 24.9    & \ninefont 84.4    && \ninefont 153    & \ninefont 24.9    & \ninefont 96.6    && \ninefont 153    & \ninefont 24.9    & \ninefont 90.6 \\
            \ninefont Adafactor &           && \ninefont 86     & \ninefont 24.9    & \ninefont 93.8    && \ninefont 86 & \ninefont 24.9    & \ninefont 84.4    && \ninefont 86 & \ninefont 24.9    & \ninefont 93.0    && \ninefont 86 & \ninefont 24.9    & \ninefont 96.3    && \ninefont 86 & \ninefont 24.9    & \ninefont 85.9   \\
            \ninefont SM3       & LLaMA-7b  && \ninefont 86     & \ninefont 24.9    & \ninefont 65.6    && \ninefont 86 & \ninefont 24.9    & \ninefont 64.8    && \ninefont 86 & \ninefont 24.9    & \ninefont 71.9    && \ninefont 86 & \ninefont 24.9    & \ninefont 34.9    && \ninefont 86 & \ninefont 24.9    & \ninefont 70.3   \\
            \ninefont CAME      &           && \ninefont 86     & \ninefont 24.9    & \ninefont 69.5    && \ninefont 86 & \ninefont 24.9    & \ninefont 43.0    && \ninefont 86 & \ninefont 24.9    & \ninefont 75.8    && \ninefont 86 & \ninefont 24.9    & \ninefont 34.8    && \ninefont 86 & \ninefont 24.9    & \ninefont 70.3    \\
            \ninefont \optim    &           && \ninefont \textbf{3.9}   & \ninefont \textbf{24.8}   & \ninefont 91.4    && \ninefont \textbf{3.9} & \ninefont \textbf{24.8}    & \ninefont 87.5    && \ninefont \textbf{3.9} & \ninefont \textbf{24.8}    & \ninefont 90.6    && \ninefont \textbf{3.9} & \ninefont \textbf{24.8}    & \ninefont 96.5    && \ninefont \textbf{3.9}  & \ninefont \textbf{24.8    }& \ninefont 89.8  \\
            \hline
            \end{tabular}
    \end{center}
    \caption{\textbf{Fine-tuning}: the optimizer and end-to-end training (one-batch) memory usage [MiB, GiB] at 100 iterations including $\MATRIX{S}_{\MATRIX{M}}$, and the performance of GPT-2, T5-small, and LLaMA-7b fine-tuned on QNLI, MNLI, QQP, STSB, and MRPC.
    }
    \label{tab:memory_footprint_and_result_of_fine_tuning}
\end{table*}

%% file: tables/time.tex
\begin{table}[!tb]
    \setlength\tabcolsep{1.0pt}
    \centering
            \begin{tabular}{l||r@{\ninefont $\pm$}r|r@{\ninefont $\pm$}r|r@{\ninefont $\pm$}r|r@{\ninefont $\pm$}r|r@{\ninefont $\pm$}r}
            \toprule
            \multicolumn{11}{c}{\ninefont \textbf{Optimization Time (ms) for a Single Training Step (Iteration)}} \\
        
            \midrule
            \ninefont \textbf{Model}               &   \multicolumn{2}{c|}{\ninefont \textbf{Adam}}    &   \multicolumn{2}{c|}{\ninefont \textbf{Adafactor}}   &   \multicolumn{2}{c|}{\ninefont \textbf{SM3}} &   \multicolumn{2}{c|}{\ninefont \textbf{CAME}}    &   \multicolumn{2}{c}{\ninefont \textbf{\optim}}  \\
        
            \bottomrule
            \ninefont MobileNetV2         &   \ninefont 127 & \ninefont 16& \,\: \ninefont  168 & \ninefont 16\;\:&   \ninefont 140 & \ninefont 17&   \ninefont 160 & \ninefont 17&   \ninefont 205 & \ninefont 13    \\
            \ninefont ResNet-50           &   \ninefont 273 & \ninefont 16& \,\:  \ninefont 316 & \ninefont 14\;\:&   \ninefont 286 & \ninefont 14&   \ninefont 307 & \ninefont 20&   \ninefont 349 & \ninefont 15    \\
            \hline
            \ninefont Transformer-base    &   \ninefont 129 &  \ninefont 7& \,\:  \ninefont 160 &  \ninefont 7\;\:&   \ninefont 134 &  \ninefont 7&   \ninefont 156 & \ninefont 7 &   \ninefont 171 &  \ninefont 7     \\
            \ninefont Transformer-big     &   \ninefont 321 & \ninefont 18& \ninefont \,\:  372 & \ninefont 18\;\:&   \ninefont 325 & \ninefont 18&   \ninefont 369 & \ninefont 18&   \ninefont 389 & \ninefont 18    \\
            \hline
            \end{tabular}
    \caption{The single optimization time (milliseconds) per step of the four optimizers, and \optim (8-bit format $\boldsymbol{S_{M}}$ in \Cref{alg:matrix_decompression} and \ref{alg:matrix_compression}): 1) MobileNetV2 and ResNet-50 on ImageNet, and 2) Transformer-base and big on WMT32k.
    }
    \label{tab:time}
\end{table}

%% file: images/transformer_base_ppl.tex
\begin{figure}[!bt]
    \centering
    \begin{subfigure}[b]{0.49\columnwidth}
        \centering
        \centerline{\includegraphics[width=\columnwidth]{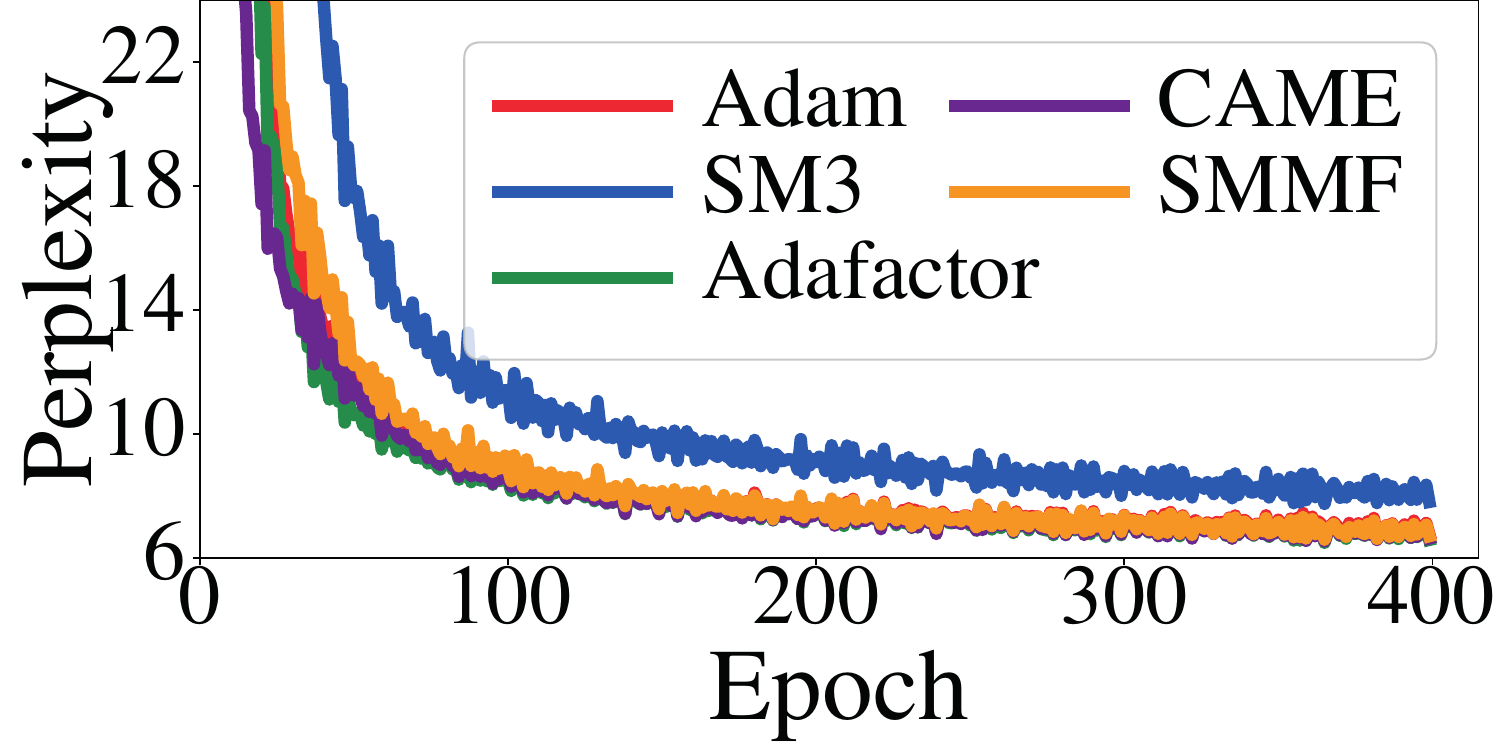}}
        \label{fig:transformer_base_ppl_left}
    \end{subfigure}
    \begin{subfigure}[b]{0.49\columnwidth}
        \centering
        \centerline{\includegraphics[width=\columnwidth]{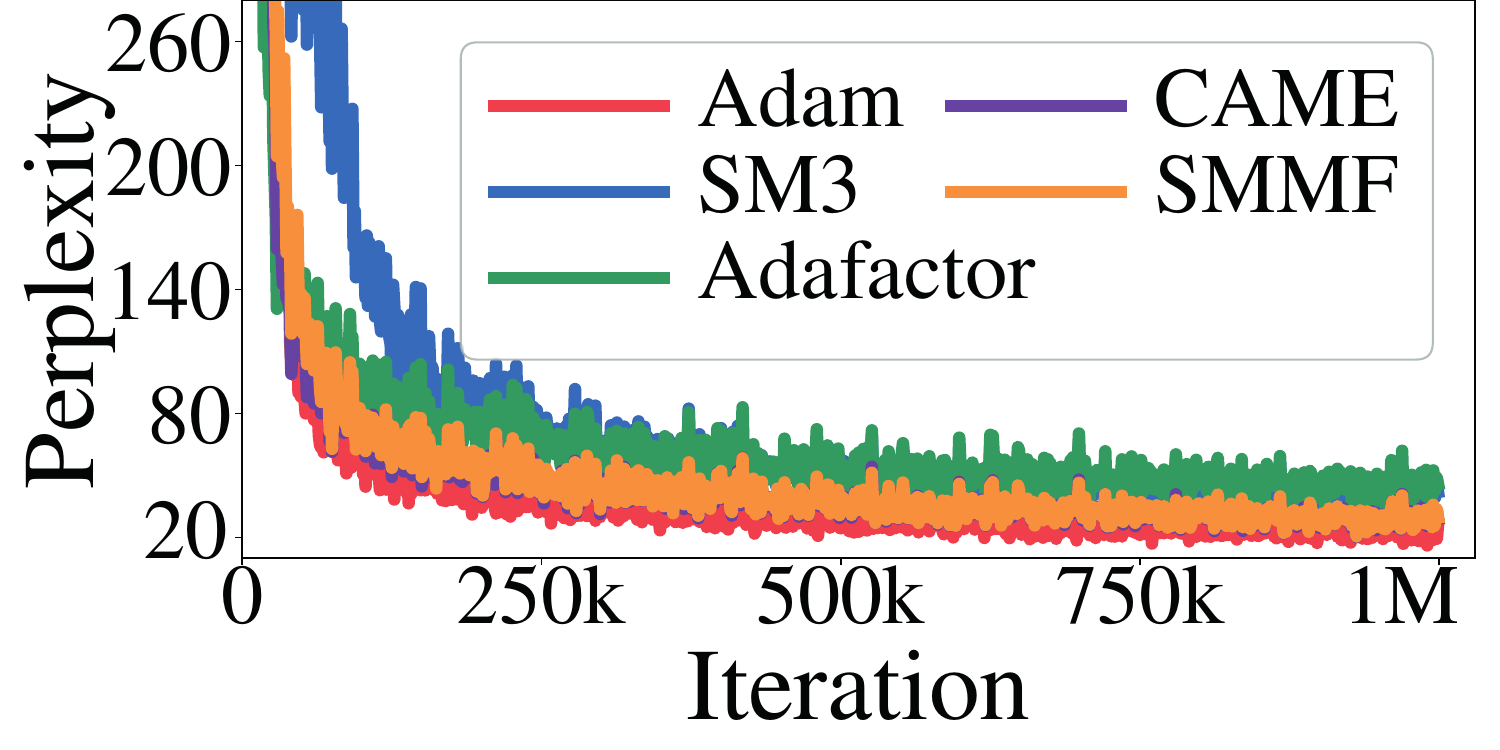}}
        \label{fig:transformer_base_ppl_right}
    \end{subfigure}

    \caption{The test perplexity of the Transformer-base model on WMT32k during full-training steps \textbf{(Left)} and BERT on BookCorpus \& Wikipedia during pre-training steps \textbf{(Right)}}
    \label{fig:transformer_base_ppl}
\end{figure}

%% file: sections/appendix.tex
\section{Relative Training Time comparison}
\label{sec:relative_training_time_comparison}
As described in \Cref{tab:time}, \optim trades off training time for memory efficiency to minimize the substantial memory consumption incurred by optimizer states, such as momentum tensors. This means that training time may increase slightly to achieve higher memory efficiency. While we have already compared the training time of the five optimizers, including Adam, Adafactor, SM3, CAME, and \optim, we present detailed training time figures for two tasks: image classification of MobileNetV2 on ImageNet and full-training a Transformer-base on WMT32k. These results demonstrate that \optim offers significant memory saving despite a slight increase in training time.

\begin{figure}[H]
    \centering
     \begin{subfigure}[b]{0.45\columnwidth}
        \centering
        \centerline{\includegraphics[width=\columnwidth]{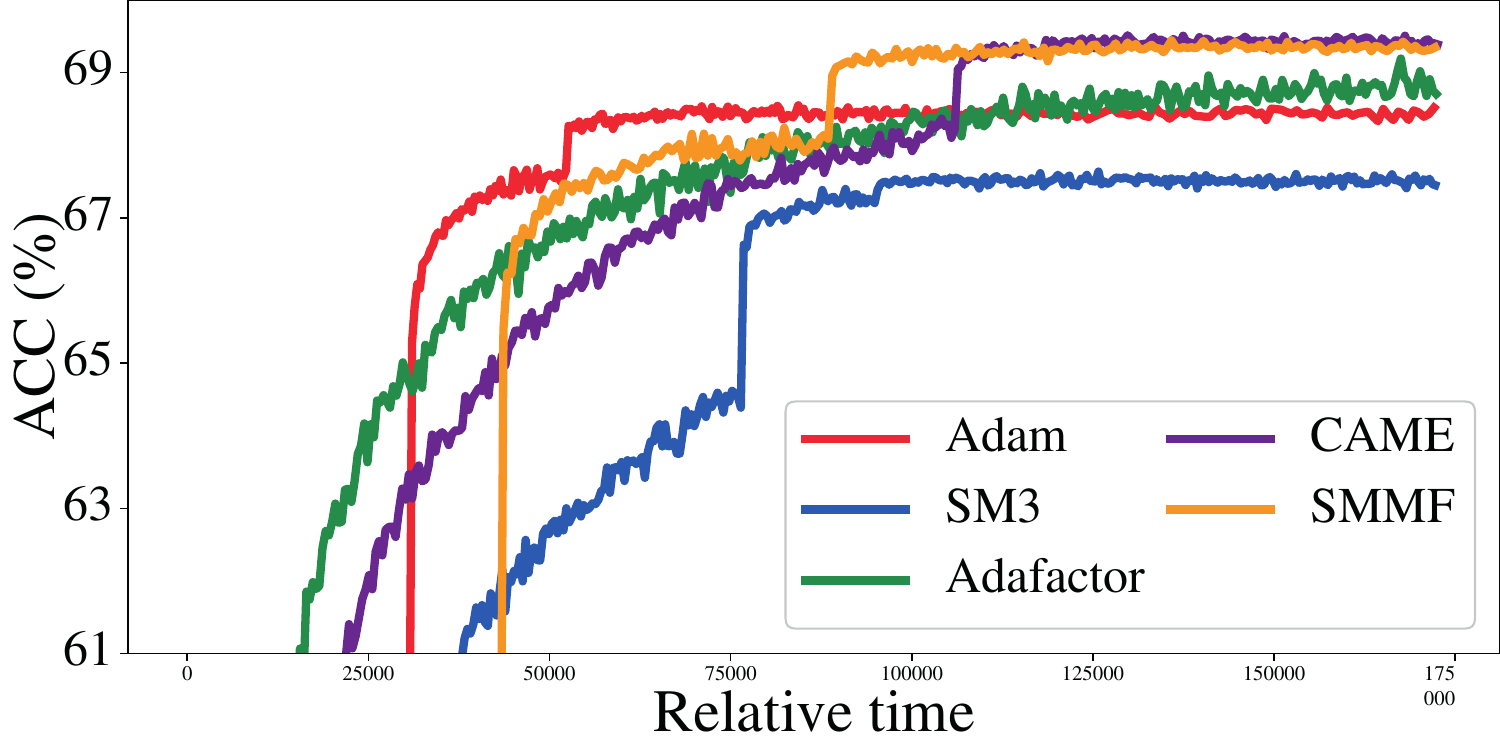}}
        \label{fig:cnn_left_relative_time}
    \end{subfigure}
    \begin{subfigure}[b]{0.45\columnwidth}
        \centering
        \centerline{\includegraphics[width=\columnwidth]{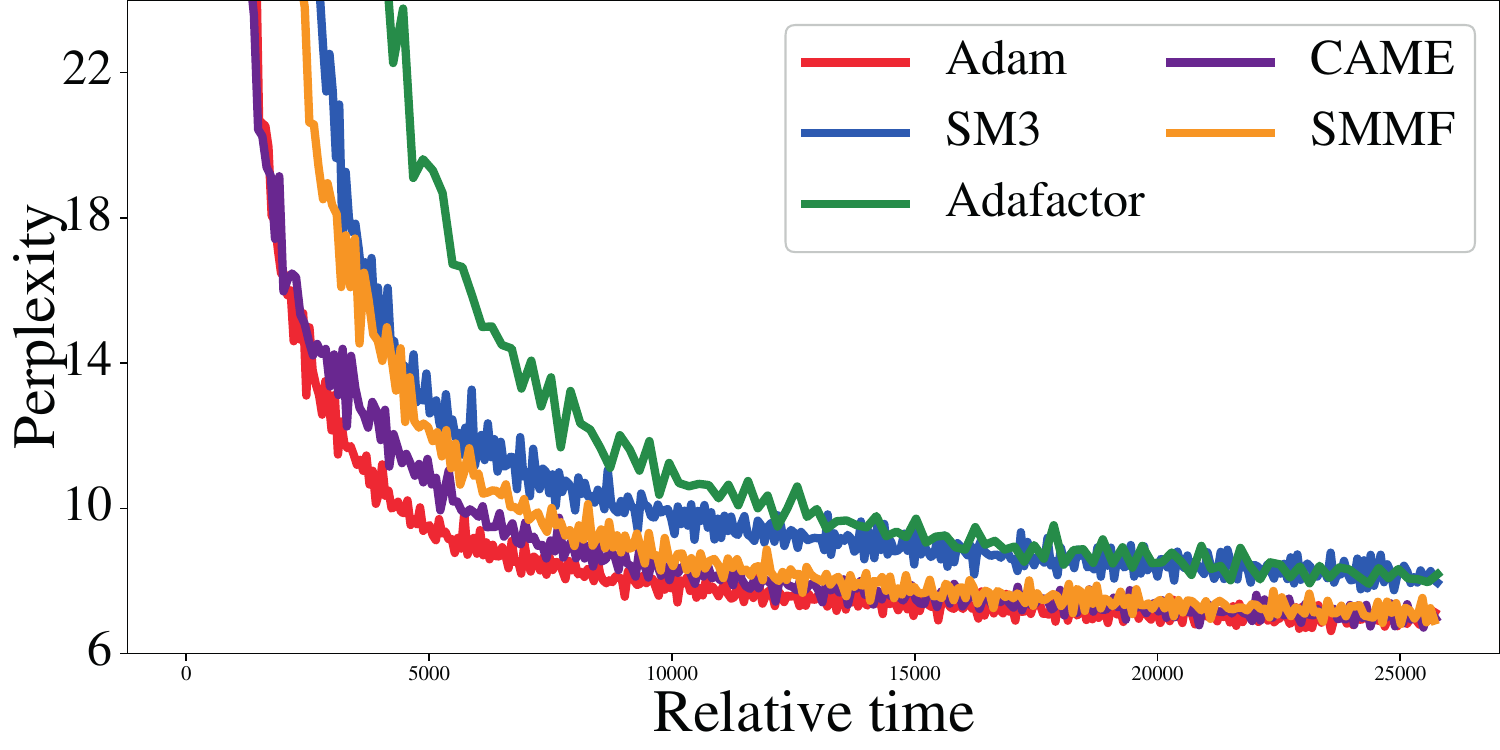}}
        \label{fig:cnn_right_relative_time}
    \end{subfigure}
    \caption{\textbf{(Left)} The validation top-1 accuracy of MobileNetV2 on ImageNet. \textbf{(Right)} The test perplexity of the Transformer-base model on WMT32k during full-training steps.}
    \label{fig:cnn_relative_time}
\end{figure}

The graphs, based on measurements from the Wandb~\footnote{https://wandb.ai/site/} platform, show the training time for two tasks, i.e., image classification and full-training . The left graph corresponds to training MobileNetV2 on ImageNet, and the right graph corresponds to training the Transformer-base model on WMT32k. As can be seen from both graphs, SMMF is slower than Adam (\Cref{fig:cnn_left,fig:transformer_base_ppl_left} for reference). This is because SMMF trades off computation and memory. Although SMMF is on average 1.4 times slower than Adam (\Cref{tab:time} for reference), it compresses memory by approximately 96\%, making it an algorithm that significantly reduces memory usage relative to its speed.

\section{NNMF (Non-Negative Matrix Factorization)}
\label{appendix:nnmf}
\input{algorithms/nnmf}
\section{Proof of \Cref{thm:the_theorem_of_square_matrixlization_discrete2}}
\begin{proof}
    \label{proof:the_theorem_of_square_matrixlization_discrete2}
    Let $N' = \prod_{r = 1} ^ {d - 2} n_r$. Then we have:
    \begin{equation}
        \prod_{r = 1} ^ {d - 2} n_r (n_{d - 1} + n_d) = N' (n_{d - 1} + n_d)
    \end{equation}
    Since $N = N' (n_{d - 1} \times n_d)$, we can rewrite the above equation:
    \begin{align}
        \prod_{r = 1} ^ {d - 2} n_r (n_{d - 1} + n_d)
            &= N \dfrac{(n_{d - 1} + n_d)}{n_{d - 1} \times n_d} \\
            &= N \left( \dfrac{1}{n_d} + \dfrac{1}{n_{d - 1}} \right)
    \end{align}
    From \Cref{lemma:3d_convex_proof}, we have the function, $f(n_d, n_{d - 1}) = \frac{1}{n_d} + \frac{1}{n_{d - 1}}$, which is symmetric with respect to $n_d = n_{d - 1}$ and a convex function where the Hessian matrix is positive definite and asymptotically converges to zero as $n_d$ and $n_{d-1}$ grow. It means the function decreases as $n_{d - 1} \simeq n_d$ (i.e., $\arg\min(n_{d - 1} - n_d))$ and both $n_d$ and $n_{d - 1}$ increase.
\end{proof}

\begin{lemma}
    \label{lemma:3d_convex_proof}
    Given $f(x, y) = \frac{1}{x} + \frac{1}{y}$ where $x, y > 0$, then the function $f(x, y)$ is a convex function.
    \begin{proof}
        In order to prove that a function $f(x, y)$ is a convex function, we have to show the Hessian matrix of $f(x, y)$ is positive semi-definite. The following is the Hessian of the function.
        \begin{equation}
            \MATRIX{H} = 
                \begin{bmatrix}
                    \dfrac{\partial ^ 2 f}{\partial x ^ 2}  &   \dfrac{\partial ^ 2 f}{\partial x \partial y} \\
                    \dfrac{\partial ^ 2 f}{\partial y \partial x}   &   \dfrac{\partial ^ 2 f}{\partial y ^ 2} \\
                \end{bmatrix}
                =
                \begin{bmatrix}
                    \dfrac{2}{x ^ 3}    &   0   \\
                    0   &   \dfrac{2}{y ^ 3}    \\
                \end{bmatrix}
        \end{equation}
        Since $x$ and $y$ are greater than 0, the eigenvalues of the Hessian matrix, $\MATRIX{H}$, are positive, which means the matrix is positive definite and the function $f(x, y)$ is a convex function. 
    \end{proof}
\end{lemma}

\section{Proof of \Cref{thm:the_theorem_of_square_matrixlization_discrete}}

\begin{proof}
    \label{proof:the_proof_of_square_matrixlization_discrete}
    Let the function, $f_-(n, m)$, be $m - n$ and the function, $f_+(n, m)$ be $m + n$.
    As rewritten in \Cref{thm:the_theorem_of_square_matrixlization_contiuous} \textit{proof}, $f_-(n, m)$ is $f_-(m) = m - \frac{\SCALAR{N}}{m}$ and $f_+(n, m)$ is $f_+(m) = m + \frac{\SCALAR{m}}{m}$.
    Since the given conditions are under the condition of \Cref{thm:the_theorem_of_square_matrixlization_contiuous}, \Cref{thm:the_theorem_of_square_matrixlization_discrete} is proved. Also, we can prove the theorem using \Cref{lemma:the_proof_of_monotonic_increasing01,lemma:the_proof_of_convex_function01}. The both lemmas show that the two functions, $f_{-}(n, m)$ and $f_{+}(n, m)$ are strictly monotonically increasing function under the given conditions, which means that the directions to the minima are equal (i.e., the sign values of each function's gradient are positive) and minimizing one function leads to minimizing another function.
\end{proof}

\begin{theorem}
    \label{thm:the_theorem_of_square_matrixlization_contiuous}
    Given $n_r, n$, and $m \in \RSET^+$, $r \in [1, d]$, $\SCALAR{N} = \prod_{r = 1} ^ d n_r = n m$, and $n \leq m$, then the following holds: 
    \begin{equation}
        \arg\min_{n, m} (m - n) = \arg\min_{n, m} (m + n)
    \end{equation}
    \begin{proof}
        \label{proof:the_proof_of_square_matrixlization}
        In order to prove \Cref{thm:the_theorem_of_square_matrixlization_contiuous}, $n$ and $m$ at the minima of the function, $f_-(n, m) = m - n$, and the function, $f_+(n, m) = m + n$, should be same under the given condition.
        Since $n$ depends on the $m$ and given constant $\SCALAR{N}$, the function, $f_-(n, m)$, becomes $f_-(m) = m - \frac{\SCALAR{N}}{m}$ and the function, $f_+(n, m)$, becomes $f_+(m) = m + \frac{\SCALAR{N}}{m}$.
        For the function, $f_-(m)$, $m$ at the minima of the function is always $\sqrt{\SCALAR{N}}$ since the function is strictly increasing function via \Cref{lemma:the_proof_of_monotonic_increasing01}, and the minima of the function is $0$ where $m = \sqrt{\SCALAR{N}} = n$ since the function is greater than or equals to $0$ under the given condition.
        Fot the function, $f_+(m)$, $m$ at the minima of the function is also always $\sqrt{\SCALAR{N}}$ since the function is a convex function and has only one $m$ which satisfies $\frac{d f_+(m)}{dm} = 0$ via \Cref{lemma:the_proof_of_convex_function01}.
    \end{proof}
\end{theorem}

\begin{lemma}
    \label{lemma:the_proof_of_monotonic_increasing01}
    Given $\SCALAR{N} \in \ZSET$ and $0 < x < \infty$, then the function, $f(x) : \RSET \rightarrow \RSET = x - \frac{\SCALAR{N}}{x}$, is a strictly increasing function.
    \begin{proof}
        In order to determine that the function, $f(x)$, is strictly increasing with respect to $x$, the following should be satisfied:
        \begin{equation}
            \label{eq:the_condition_of_stirctly_increasing}
            \dfrac{d f(x)}{d x} > 0
        \end{equation}
        The differentiation of the function, $f(x) = x - \frac{\SCALAR{N}}{x}$, with respect to $x$ is:
        \begin{equation}
            \label{eq:diff_of_the_condition_of_stirctly_increasing}
            \dfrac{d f(x)}{d x} = 1 + \dfrac{\SCALAR{N}}{x ^ 2}
        \end{equation}
        Under the given condition, $0 < x < \infty$, \Cref{eq:diff_of_the_condition_of_stirctly_increasing} is always positive.
    \end{proof}
\end{lemma}

\begin{lemma}
    \label{lemma:the_proof_of_convex_function01}
    Given $\SCALAR{N} \in \ZSET$ and $0 < x < \infty$, the function, $f(x): \RSET \rightarrow \RSET = x + \frac{\SCALAR{N}}{x}$, is a convex function where the number of $x$ which satisfies $\frac{d f(x)}{d x} = 0$ is one.
    \begin{proof}
        In order to show that the function, $f(x) = x + \frac{\SCALAR{N}}{x}$, is a convex function, the second derivative of the function should be greater than or equal to zero:
        \begin{equation}
            \label{eq:the_proof_of_convex_function01_eq_01}
            \dfrac{d ^ 2 f(x)}{d x ^ 2} = \dfrac{2 \SCALAR{N}}{x ^ 3}
        \end{equation}
        Under the given condition, $0 < x < \infty$, \Cref{eq:the_proof_of_convex_function01_eq_01} is always greater than zero.
        
        In order to prove that the number of $x$ which satisfies $\frac{d f(x)}{d x} = 0$ is one, we need the derivative of the function, $f(x)$:
        \begin{equation}
            \label{eq:the_proof_of_convex_function01_eq_02}
            \dfrac{d f(x)}{d x} = 1 - \dfrac{\SCALAR{N}}{x ^ 2} = 0
        \end{equation}
        The $x$ which satisfies \Cref{eq:the_proof_of_convex_function01_eq_02} are $\pm \sqrt{\SCALAR{N}}$, but the feasible $x$ is $x = \sqrt{\SCALAR{N}}$ under the given condition, $0 < x < \infty$, so that the number of possible $x$ is one.
    \end{proof}
\end{lemma}

\section{Proof of \Cref{thm:regret_bounding}}
\label{proof:regret_bounding}

\begin{proof}
    From \Cref{lemma:convex_property_02} we have:
    \begin{equation}
        \label{eq:regret_equation_01}
        R(T) = \sum_{t = 1}^T(f_t(\weight_t) - f_t(\weight^*)) \leq \sum_{t = 1}^T \sum_{i = 1}^d f(\weight_t)_i(\weight_t - \weight^*)_i
    \end{equation}

    From the update rule of \Cref{alg:brief_algorithm} without $\epsilon$ for simple proof, we have:
    \begin{align}
        \weight_{t + 1}
            &= \weight_t - \eta_t \dfrac{\VECTOR{m}_t}{\sqrt{\VECTOR{v}_t}} \\
            &= \weight_t - \eta_t \left\{ \dfrac{\beta_{1, t}}{\sqrt{\VECTOR{v}_t}} \VECTOR{m}_{t - 1} + \dfrac{(1 - \beta_{1, t})}{\sqrt{\VECTOR{v}_t}} \VECTOR{g}_t \right\}
    \end{align}

    In order to make the form of \Cref{eq:regret_equation_01}, we subtract $\weight^*$ and square both sides of the equation, and write the equation focusing on the $i^{th}$ element in the vector $\weight_t$.
    \begin{align}
        (\weight_{t + 1, i} - \weight_{, i}^*)^2
            &= (\weight_{t, i} - \weight_{, i}^*)^2 - 2\eta_t \left\{ \dfrac{\beta_{1, t}}{\sqrt{\VECTOR{v}_{t, i}}} \VECTOR{m}_{t - 1, i} + \dfrac{(1 - \beta_{1, t})}{\sqrt{\VECTOR{v}_{t, i}}} \VECTOR{g}_{t, i} \right\} (\weight_{t, i} - \weight_{, i}^*) \\
            &+ \eta_t^2 \left( \dfrac{\VECTOR{m}_{t, i}}{\sqrt{\VECTOR{v}_{t, i}}} \right)^2
    \end{align}

    By rearranging the equation, we have:
    \begin{align}
        \VECTOR{g}_{t, i}(\weight_{t, i} - \weight_{, i}^*) 
            &= \dfrac{\sqrt{\VECTOR{v}_{t, i}}}{2 \eta_t (1 - \beta_{1, t})}\left\{(\weight_{t, i} - \weight_{, i}^*)^2 - (\weight_{t + 1, i} - \weight_{, i}^*)^2\right\} \\
            &+ \dfrac{\eta_t \sqrt{\VECTOR{v}_{t, i}}}{2(1 - \beta_{1, t})} \left( \dfrac{\VECTOR{m}_{t, i}}{\sqrt{\VECTOR{v}_{t, i}}} \right)^2 - \dfrac{\beta_{1, t}}{(1 - \beta_{1, t})}\hat{\VECTOR{m}}_{t - 1, i}(\weight_{t, i} - \weight_{, i}^*)
    \end{align}

    By applying Young's inequality, we have:
    \begin{align}
        \VECTOR{g}_{t, i}(\weight_{t, i} - \weight_{, i}^*) 
            &\leq \dfrac{\sqrt{\VECTOR{v}_{t, i}}}{2 \eta_t (1 - \beta_{1, t})}\left\{(\weight_{t, i} - \weight_{, i}^*)^2 - (\weight_{t + 1, i} - \weight_{, i}^*)^2\right\} \\
            &+ \dfrac{\beta_{1, t}}{2 \eta_{t - 1} (1 - \beta_{1, t})}(\weight_{, i}^* - \weight_{t, i})^2 \sqrt{\VECTOR{v}_{t - 1, i}} + \dfrac{\eta_{t - 1} \beta_{1, t}}{2(1 - \beta_{1, t})} \dfrac{\hat{\VECTOR{m}}_{t - 1, i}^2}{\sqrt{\VECTOR{v}_{t - 1, i}}}\\
            &+ \dfrac{\eta_t \sqrt{\VECTOR{v}_{t, i}}}{2(1 - \beta_{1, t})} \left( \dfrac{\VECTOR{m}_{t, i}}{\sqrt{\VECTOR{v}_{t, i}}} \right)^2
    \end{align}
    
    Since we want to compute regret, we compute the below inequality by applying \Cref{lemma:sum_eta_m_v} and \Cref{lemma:sum_eta_hat_m_v}:
    \begin{align}
        \sum_{t = 1}^T \sum_{i = 1}^d\VECTOR{g}_{t, i}(\weight_{t, i} - \weight_{, i}^*) 
            &\leq \sum_{t = 1}^T \sum_{i = 1}^d\dfrac{\sqrt{\VECTOR{v}_{t, i}}}{2 \eta_t (1 - \beta_{1, t})}\left\{(\weight_{t, i} - \weight_{, i}^*)^2 - (\weight_{t + 1, i} - \weight_{, i}^*)^2\right\} \\
            &+ \sum_{t = 1}^T \sum_{i = 1}^d\dfrac{\beta_{1, t}}{2 \eta_{t - 1} (1 - \beta_{1, t})}(\weight_{, i}^* - \weight_{t, i})^2 \sqrt{\VECTOR{v}_{t - 1, i}} \\
            &+ \dfrac{\zeta_1 \zeta_2 (1 + \beta_1) \sqrt{d}}{(1 - \beta_1)^3} \sqrt{\sum_{i = 1}^d \sum_{t = 1}^T \VECTOR{g}_{t, i}^2}
    \end{align}
    under the given condition $\beta_1 = \beta_{1, 1}$ and $\beta_{1, t} \leq \beta_1$.

    By rearranging the inequality and using the condition $\eta_{t - 1}\sqrt{\VECTOR{v}_{t, i}} \geq \eta_{t} \sqrt{\VECTOR{v}_{t - 1, i}}$, we have:
    \begin{align}
        \sum_{t = 1}^T \sum_{i = 1}^d\VECTOR{g}_{t, i}(\weight_{t, i} - \weight_{, i}^*) 
            &\leq \sum_{i = 1}^d \dfrac{1}{2\eta_1 (1 - \beta_{1, 1})} (\weight_{1, i} - \weight_{, i}^*)^2\sqrt{\VECTOR{v}_{1, i}} \\
            &+ \sum_{t = 2}^T \sum_{i = 1}^d \dfrac{1}{2(1 - \beta_{1, t})}(\weight_{t, i} - \weight_{, i}^*)^2 \left( \dfrac{\sqrt{\VECTOR{v}_{t, i}}}{\eta_t} - \dfrac{\sqrt{\VECTOR{v}_{t - 1, i}}}{\eta_{t - 1}} \right) \\  
            &+ \sum_{t = 1}^T \sum_{i = 1}^d\dfrac{\beta_{1, t}}{2 \eta_{t} (1 - \beta_{1, t})}(\weight_{, i}^* - \weight_{t, i})^2 \sqrt{\VECTOR{v}_{t, i}} \\
            &+ \dfrac{\zeta_1 \zeta_2 (1 + \beta_1) \sqrt{d}}{(1 - \beta_1)^3} \sqrt{\sum_{i = 1}^d \sum_{t = 1}^T \VECTOR{g}_{t, i}^2}
    \end{align}

    Using the assumption $\lvert \lvert \weight_t - \weight^* \rvert \rvert_2 \leq D$ and $\lvert \lvert \weight_t - \weight^* \rvert \rvert_{\infty} \leq D_{\infty}$, we have:
    \begin{align}
       R(T) &\leq \dfrac{D^2}{2\eta_T (1 - \beta_1)} \sum_{i = 1}^d \sqrt{\VECTOR{v}_{T, i}}\\
            &+ \sum_{t = 1}^T \sum_{i = 1}^d\dfrac{\beta_{1, t}}{2 \eta_{t} (1 - \beta_{1, t})}D_{\infty}^2 \sqrt{\VECTOR{v}_{t, i}} \\
            &+ \dfrac{\zeta_1 \zeta_2 (1 + \beta_1) \sqrt{d}}{(1 - \beta_1)^3} \sqrt{\sum_{i = 1}^d \sum_{t = 1}^T \VECTOR{g}_{t, i}^2}
    \end{align}
\end{proof}

\begin{lemma}
    \label{lemma:convex_property_01}
    Let a function $f: \RSET^d \rightarrow \RSET$ be a convex function. Given $\forall \VECTOR{x} \in \RSET^d$, $\forall \VECTOR{y} \in \RSET^d$, and $\rho \in [0, 1]$, then the following holds:
    \begin{equation}
        \rho f(\VECTOR{x}) + (1 - \rho)f(\VECTOR{y}) \geq f(\rho \VECTOR{x} + (1 - \rho) \VECTOR{y})
    \end{equation}
\end{lemma}

\begin{lemma}
    \label{lemma:convex_property_02}
    Let a function $f: \RSET^d \rightarrow \RSET$ be a convex function. Given $\forall \VECTOR{x} \in \RSET^d$ and $\forall \VECTOR{y} \in \RSET^d$, then the following holds:
    \begin{align}
        f(\VECTOR{y}) &\geq f(\VECTOR{x}) + \nabla f(\VECTOR{x})^{\top} (\VECTOR{y} - \VECTOR{x}) \\
        &= f(\VECTOR{x}) + \sum_{i = 1}^d \nabla f(\VECTOR{x})_i(\VECTOR{y} - \VECTOR{x})_i
    \end{align}
    where $\nabla f(\VECTOR{x})$ is the $i^{th}$ element in the vector $\nabla f(\VECTOR{x})$ and $(\VECTOR{y} - \VECTOR{x})_i$ is the $i^{th}$ element in the vector.
\end{lemma}

\begin{lemma}
    \label{lemma:compact_form_of_m_and_v_hat}
    Let $\hat{\VECTOR{m}}_t$, $\hat{\VECTOR{v}}_t$, $\VECTOR{m}_t$, $\VECTOR{v}_t$, and $\VECTOR{g}_t$ are respectively vectorized $\hat{\MATRIX{m}}_t$, $\hat{\MATRIX{v}}_t$, $\MATRIX{m}_t$, $\MATRIX{v}_t$, and $\MATRIX{G}_t$, in \Cref{alg:brief_algorithm}. Given $\beta_{1, j}$ and $\beta_{2, j}$ at $j = t$ in the algorithm, the followings are satisfied:
    \begin{align}
        \hat{\VECTOR{m}}_t - \VECTOR{m}_t &= \VECTOR{e}_{m, t}  \\
        \hat{\VECTOR{v}}_t - \VECTOR{v}_t &= \VECTOR{e}_{v, t}  \\
        \hat{\VECTOR{g}}_{m, t} &= \left( \VECTOR{g}_t + \dfrac{\VECTOR{e}_{m, t}}{(1 - \beta_{1, t})} \right)  \\
        \hat{\VECTOR{g}}_{v, t} &= \left( \VECTOR{g}_t^2 + \dfrac{\VECTOR{e}_{v, t}}{(1 - \beta_{2, t})} \right)    \\
        \label{eq:compact_form_of_m_hat}
        \hat{\VECTOR{m}}_t &= \sum_{j = 1}^t (1 - \beta_{1, j}) \prod_{k = 1}^{t - j}\beta_{1, t - k + 1}\hat{\VECTOR{g}}_{m, j} = \VECTOR{m}_t + \VECTOR{e}_{m, t} \\
        \hat{\VECTOR{v}}_t &= \sum_{j = 1}^t (1 - \beta_{2, j}) \prod_{k = 1}^{t - j}\beta_{2, t - k + 1}\hat{\VECTOR{g}}_{m, j} = \VECTOR{v}_t + \VECTOR{e}_{v, t}
    \end{align}

    where $\VECTOR{e}_{m, t}$ and $\VECTOR{e}_{v, t}$ are $decompression \rightarrow compression$ errors at $t \in [1, T]$ such that $\hat{\VECTOR{m}}_t - \VECTOR{m}_t = \VECTOR{e}_{m, t}$ and $\hat{\VECTOR{v}}_t - \VECTOR{v}_t = \VECTOR{e}_{v, t}$.

    \begin{proof}
        Following \Cref{alg:brief_algorithm}, we have:
        \begin{align}
            \VECTOR{m}_1
                &= (1 - \beta_{1, 1})\VECTOR{g}_1 \\
            \hat{\VECTOR{m}}_1
                &= \VECTOR{m}_1 + \VECTOR{e}_{m, 1} = (1 - \beta_{m, 1}) \left( \VECTOR{g}_1 + \dfrac{\VECTOR{e}_{m, 1}}{(1 - \beta_{1, 1})} \right) \\
                &= (1 - \beta_{1, 1})\hat{\VECTOR{g}}_{m, 1} \\
            \VECTOR{m}_2
                &= \beta_{1, 2}\hat{\VECTOR{m}}_1 + (1 - \beta_{1, 2})\VECTOR{g}_2 \\
            \hat{\VECTOR{m}}_2
                &= \VECTOR{m}_2 + \VECTOR{e}_{m, 2} \\
                &= \beta_{1, 2} \hat{\VECTOR{m}}_1 + (1 - \beta_{m, 2})\hat{\VECTOR{g}}_{m, 2} \\
                \nonumber
                \vdots \\
            \hat{\VECTOR{m}}_t &= \sum_{j = 1}^t(1 - \beta_{1, j}) \prod_{k = 1}^{t - j}\beta_{1, t - k + 1} \hat{\VECTOR{g}}_{m, j}
        \end{align}

        We can apply the proof to $\hat{\VECTOR{v}}_t$ momentum using similar way.
    \end{proof}
\end{lemma}

\begin{lemma}
    \label{lemma:m_t_with_g_tilde}
    \label{lemma:v_t_with_g_tilde}
    Let $j$ be in $[1: \tau]$ (from initial step to current step) and let $\tilde{\VECTOR{e}}_{m, j}$ and $\tilde{\VECTOR{e}}_{v, j}$ follow:
    \begin{align}
        \label{eq:tilde_e_m}
        \tilde{\VECTOR{e}}_{m, j} &= 
            \begin{dcases*}
                \VECTOR{0},     & if $j = \tau$ \\
                \VECTOR{e}_{m, j},  & otherwise
            \end{dcases*} \\
        \label{eq:tilde_e_v}
        \tilde{\VECTOR{e}}_{v, j} &= 
            \begin{dcases*}
                \VECTOR{0},     & if $j = \tau$ \\
                \VECTOR{e}_{v, j},  & otherwise 
            \end{dcases*}
    \end{align}

    Then the following holds:
    \begin{align}
        \tilde{\VECTOR{g}}_{m,j} &= \VECTOR{g}_j + \frac{\tilde{\VECTOR{e}}_{m, j}}{(1 - \beta_{1, j})} \\
        \tilde{\VECTOR{g}}_{v,j} &= \VECTOR{g}_j^2 + \frac{\tilde{\VECTOR{e}}_{v, j}}{(1 - \beta_{2, j})} \\
        \label{eq:m_t_with_g_tilde}
        \VECTOR{m}_t &= \sum_{j = 1}^t(1 - \beta_{1, j}) \prod_{k = 1}^{t - j}\beta_{1, t - k + 1} \tilde{\VECTOR{g}}_{m, j}\\       
        \label{eq:v_t_with_g_tilde}
        \VECTOR{v}_t &= \sum_{j = 1}^t(1 - \beta_{2, j}) \prod_{k = 1}^{t - j}\beta_{2, t - k + 1} \tilde{\VECTOR{g}}_{m, j}
    \end{align}

    \begin{proof}
        From \Cref{eq:compact_form_of_m_hat}, we have:
        \begin{align}
             \VECTOR{m}_t = \sum_{j = 1}^t (1 - \beta_{1, j}) \prod_{k = 1}^{t - j}\beta_{1, t - k + 1}\hat{\VECTOR{g}}_{m, j} - \VECTOR{e}_{m, t}
        \end{align}

        By the definition of $\hat{\VECTOR{g}}_{m, j}$ at \Cref{lemma:compact_form_of_m_and_v_hat}, we have:
        \begin{align}
            \VECTOR{m}_t 
                &= \sum_{j = 1}^t (1 - \beta_{1, j}) \prod_{k = 1}^{t - j}\beta_{1, t - k + 1} \left( \VECTOR{g}_j + \dfrac{\VECTOR{e}_{m, j}}{(1 - \beta_{1, j})} \right) - \VECTOR{e}_{m, t} \\
                &= \sum_{j = 1}^t (1 - \beta_{1, j}) \prod_{k = 1}^{t - j}\beta_{1, t - k + 1} \VECTOR{g}_j + \sum_{j = 1}^t \prod_{k = 1}^{t - j}\beta_{1, t - k + 1} \VECTOR{e}_{m, j} - \VECTOR{e}_{m, t} \\
                \label{eq:m_t_with_e_tilde}
                &= \sum_{j = 1}^t (1 - \beta_{1, j}) \prod_{k = 1}^{t - j}\beta_{1, t - k + 1} \VECTOR{g}_j + \sum_{j = 1}^{t - 1} \prod_{k = 1}^{t - j}\beta_{1, t - k + 1} \VECTOR{e}_{m, j}
        \end{align}

        By introducing the definition of $\tilde{\VECTOR{e}}_{m, j}$, we have:
        \begin{align}
            \VECTOR{m}_t
                &= \sum_{j = 1}^t (1 - \beta_{1, j}) \prod_{k = 1}^{t - j}\beta_{1, t - k + 1} \VECTOR{g}_j + \sum_{j = 1}^{t} \prod_{k = 1}^{t - j}\beta_{1, t - k + 1} \tilde{\VECTOR{e}}_{m, j} \\
                &= \sum_{j = 1}^t(1 - \beta_{1, j}) \prod_{k = 1}^{t - j}\beta_{1, t - k + 1} \tilde{\VECTOR{g}}_{m, j}
        \end{align}

        We can apply the proof to $\VECTOR{v}_t$ using similar way.
    \end{proof}
\end{lemma}

\begin{lemma}
    \label{lemma:sum_eta_m_v}
    Given the conditions and notations in \Cref{thm:regret_bounding}, we have:
    \begin{align}
        \sum_{i = 1}^d\sum_{t = 1}^T \eta_t \dfrac{\VECTOR{m}_{t, i}^2}{\sqrt{\VECTOR{v}_{t, i}}}
            &\leq \dfrac{2\zeta_1 \zeta_2}{(1 - \beta_1)^2} \sqrt{\sum_{i = 1}^d \sum_{t = 1}^T \VECTOR{g}_{t, i}^2}
    \end{align}
    
    \begin{proof}
        \begin{align}
            \sum_{i = 1}^d\sum_{t = 1}^T \eta_t \dfrac{\VECTOR{m}_{t, i}^2}{\sqrt{\VECTOR{v}_{t, i}}}
                &= \sum_{i = 1}^d \sum_{t = 1}^{T - 1} \eta_t \dfrac{\VECTOR{m}_{t, i}^2}{\sqrt{\VECTOR{v}_{t, i}}} + \sum_{i = 1}^d \eta_T \dfrac{\VECTOR{m}_{T, i}^2}{\sqrt{\VECTOR{v}_{T, i}}} \\
                &= \sum_{i = 1}^d \sum_{t = 1}^{T - 1} \eta_t \dfrac{\VECTOR{m}_{t, i}^2}{\sqrt{\VECTOR{v}_{t, i}}} + \eta_T \sum_{i = 1}^d \dfrac{\left\{ \sum_{j = 1}^T(1 - \beta_{1, j}) \prod_{k = 1}^{T - j} \beta_{1, T - k + 1} \tilde{\VECTOR{g}}_{m, j, i} \right\}^2}{\sqrt{\sum_{j = 1}^T (1 - \beta_{2, j}) \prod_{k = 1}^{T - j} \beta_{2, T - k + 1} \tilde{\VECTOR{g}}_{v, j, i}}} \\
                &\leq \sum_{i = 1}^d \sum_{t = 1}^{T - 1} \eta_t \dfrac{\VECTOR{m}_{t, i}^2}{\sqrt{\VECTOR{v}_{t, i}}} \\
                &+ \eta_T \sum_{i = 1}^d \dfrac{\left\{ \sum_{j = 1}^T(1 - \beta_{1, j})^2 \prod_{k = 1}^{T - j} \beta_{1, T - k + 1}\right\} \left\{ \sum_{j = 1}^T \prod_{k = 1}^{T - j} \beta_{1, T - k + 1} \tilde{\VECTOR{g}}_{m, j, i}^2 \right\}}{\sqrt{\sum_{j = 1}^T (1 - \beta_{2, j}) \prod_{k = 1}^{T - j} \beta_{2, T - k + 1} \tilde{\VECTOR{g}}_{v, j, i}}} \\
                &\leq \sum_{i = 1}^d \sum_{t = 1}^{T - 1} \eta_t \dfrac{\VECTOR{m}_{t, i}^2}{\sqrt{\VECTOR{v}_{t, i}}} \\
                &+ \eta_T \sum_{i = 1}^d \dfrac{\left\{ \sum_{j = 1}^T \prod_{k = 1}^{T - j} \beta_{1, T - k + 1}\right\} \left\{ \sum_{j = 1}^T \prod_{k = 1}^{T - j} \beta_{1, T - k + 1} \tilde{\VECTOR{g}}_{m, j, i}^2 \right\}}{\sqrt{\sum_{j = 1}^T (1 - \beta_{2, j}) \prod_{k = 1}^{T - j} \beta_{2, T - k + 1} \tilde{\VECTOR{g}}_{v, j, i}}} \\
                &\leq \sum_{i = 1}^d \sum_{t = 1}^{T - 1} \eta_t \dfrac{\VECTOR{m}_{t, i}^2}{\sqrt{\VECTOR{v}_{t, i}}} + \eta_T \sum_{i = 1}^d \dfrac{\left( \sum_{j = 1}^T \beta_{1}^{T - j} \right) \left\{ \sum_{j = 1}^T \beta_{1}^{T - j} \tilde{\VECTOR{g}}_{m, j, i}^2 \right\}}{\sqrt{\sum_{j = 1}^T (1 - \beta_{2, j}) \prod_{k = 1}^{T - j} \beta_{2, T - k + 1} \tilde{\VECTOR{g}}_{v, j, i}}} \\  
                &\leq \sum_{i = 1}^d \sum_{t = 1}^{T - 1} \eta_t \dfrac{\VECTOR{m}_{t, i}^2}{\sqrt{\VECTOR{v}_{t, i}}} + \dfrac{\eta_T}{1 - \beta_1} \sum_{i = 1}^d \dfrac{\left\{ \sum_{j = 1}^T \beta_{1}^{T - j} \tilde{\VECTOR{g}}_{m, j, i}^2 \right\}}{\sqrt{\sum_{j = 1}^T (1 - \beta_{2, j}) \prod_{k = 1}^{T - j} \beta_{2, T - k + 1} \tilde{\VECTOR{g}}_{v, j, i}}}
        \end{align}

        The first inequality comes from Cauchy-Schwarz inequality, the thrid inequality comes from the condition, i.e., $\beta_{1, T - k + 1} \leq \beta_1$ and $\prod_{k = 1}^{T - j} \beta_1 = \beta_1^{T - j}$. The last inequality comes from $\sum_{j = 1}^T \beta_1^{T - j} \leq \frac{1}{1 - \beta_1}$. Subsequently, using the definition of $\zeta_1$ and $\zeta_2$ in the condition, for some $\zeta_1 > 0$ and $\zeta_2 > 0$, we have:
        \begin{align}
            \sum_{i = 1}^d\sum_{t = 1}^T \eta_t \dfrac{\VECTOR{m}_{t, i}^2}{\sqrt{\VECTOR{v}_{t, i}}}
                &\leq \sum_{i = 1}^d \sum_{t = 1}^{T - 1} \eta_t \dfrac{\VECTOR{m}_{t, i}^2}{\sqrt{\VECTOR{v}_{t, i}}} + \dfrac{\zeta_2}{1 - \beta_1} \sum_{i = 1}^d \dfrac{ \sum_{j = 1}^T \beta_{1}^{T - j} \tilde{\VECTOR{g}}_{m, j, i}^2}{\sqrt{\sum_{j = 1}^T \tilde{\VECTOR{g}}_{v, j, i}}} \\
                &\leq \sum_{i = 1}^d \sum_{t = 1}^{T - 1} \eta_t \dfrac{\VECTOR{m}_{t, i}^2}{\sqrt{\VECTOR{v}_{t, i}}} + \dfrac{\zeta_1 \zeta_2}{1 - \beta_1} \sum_{i = 1}^d \sum_{j = 1}^T \dfrac{\beta_{1}^{T - j} \tilde{\VECTOR{g}}_{v, j, i}}{\sqrt{\sum_{k = 1}^j \tilde{\VECTOR{g}}_{v, k, i}}} \\
                &\leq \dfrac{\zeta_1 \zeta_2}{1 - \beta_1} \sum_{i = 1}^d \sum_{j = 1}^T \dfrac{\sum_{l = 0}^{T - j}\beta_1^l{\tilde{\VECTOR{g}}_{v, j, i}}}{\sqrt{\sum_{k = 1}^j \tilde{\VECTOR{g}}_{v, k, i}}} \\
                &\leq \dfrac{\zeta_1 \zeta_2}{(1 - \beta_1)^2} \sum_{i = 1}^d \sum_{j = 1}^T \dfrac{\tilde{\VECTOR{g}}_{v, j, i}}{\sqrt{\sum_{k = 1}^j \tilde{\VECTOR{g}}_{v, k, i}}}
        \end{align}
        since $\tilde{\VECTOR{g}}_{m, j, i}^2$ and $\tilde{\VECTOR{g}}_{v, j, i}$ are positive. he last inequality comes from $\sum_{l = 0}^{T - j}\beta_1^l \leq \frac{1}{1 - \beta_1}$. Using \Cref{lemma:the_inquality_for_norm01}, the last inequality becomes:
        \begin{align}
            \sum_{i = 1}^d\sum_{t = 1}^T \eta_t \dfrac{\VECTOR{m}_{t, i}^2}{\sqrt{\VECTOR{v}_{t, i}}}
                &\leq \dfrac{2\zeta_1 \zeta_2}{(1 - \beta_1)^2} \sum_{i = 1}^d \sqrt{\sum_{t = 1}^T \tilde{\VECTOR{g}}_{v, t, i}} \\
                &\leq \dfrac{2\zeta_1 \zeta_2 \sqrt{d}}{(1 - \beta_1)^2} \sqrt{\sum_{i = 1}^d \sum_{t = 1}^T \tilde{\VECTOR{g}}_{v, t, i}} \\
                &= \dfrac{2\zeta_1 \zeta_2 \sqrt{d}}{(1 - \beta_1)^2} \sqrt{\sum_{i = 1}^d \sum_{t = 1}^T \VECTOR{g}_{t, i}^2}
        \end{align}
        where the second inequality comes from a concave property and the equation comes from \Cref{lemma:error_zero}.
    \end{proof}
\end{lemma}

\begin{lemma}
\label{lemma:sum_eta_hat_m_v} 
Given the conditions and notations in \Cref{thm:regret_bounding}, using similar way~\Cref{lemma:sum_eta_m_v} we have:
\begin{align}
    \sum_{i = 1}^d\sum_{t = 1}^T \eta_t \dfrac{\hat{\VECTOR{m}}_{t, i}^2}{\sqrt{\VECTOR{v}_{t, i}}}
        &\leq \dfrac{2\zeta_1 \zeta_2 \sqrt{d}}{(1 - \beta_1)^2} \sqrt{\sum_{i = 1}^d \sum_{t = 1}^T \VECTOR{g}_{t, i}^2}
\end{align}
\end{lemma}

\begin{lemma}
    Let $\MATRIX{E} \in \RSET^{n \times m}$ be a $decompression \rightarrow compression$ error matrix from NNMF algorithm~\cite{adafactor} and $\VECTOR{e}$ is a vectorized vector of $\MATRIX{E}$. Then, the followings hold:
    \label{lemma:error_zero}
    \begin{align}
        \sum_{i = 1}^n \sum_{j = 1}^m \SCALAR{E}_{i, j} = \sum_{i = 1}^{n \times m} \SCALAR{e}_t = 0 \text{, where }\VECTOR{e} = \bar{\text{vec}}(\MATRIX{E})
    \end{align}

    \begin{proof}
        Let $\MATRIX{U}$ be in $\RSET^{n \times m}$, $\SCALAR{U}_{i, j}$ be the element of $\MATRIX{U}$ at $(i, j)$, and $\SCALAR{U}_{i, j} \geq 0$. Let summation of all elements in the matrix be not zero. Then, the element of decompressed matrix, $\hat{\MATRIX{U}}$ satisfies:
        \begin{align}
            \hat{\SCALAR{U}}_{i, j} &= \dfrac{\sum_{l = 1}^m \SCALAR{U}_{i, l}\sum_{k = 1}^n \SCALAR{U}_{k, j}}{\sum_{l = 1}^m\sum_{k = 1}^n \SCALAR{U}_{k, l}}
            \label{eq:nnmf_adafactor} \\
            &= \SCALAR{U}_{i, j} + \SCALAR{E}_{i, j}
        \end{align}
        where the \Cref{eq:nnmf_adafactor} comes from the Adafactor NNMF~\cite{adafactor}.
        
        By rearranging the above equation, we have:
        \begin{align}
            \SCALAR{U}_{i, j} = \dfrac{\sum_{l = 1}^m \SCALAR{U}_{i, l}\sum_{k = 1}^n \SCALAR{U}_{k, j}}{\sum_{l = 1}^m\sum_{k = 1}^n \SCALAR{U}_{k, l}} - \SCALAR{E}_{i, j}
        \end{align}

        By adding all elements in $\MATRIX{U}$, we have:
        \begin{align}
            \sum_{i = 1}^n \sum_{j = 1}^m \SCALAR{U}_{i, j} = \sum_{i = 1}^n \sum_{j = 1}^m 
 \dfrac{\sum_{l = 1}^m \SCALAR{U}_{i, l}\sum_{k = 1}^n \SCALAR{U}_{k, j}}{\sum_{l = 1}^m\sum_{k = 1}^n \SCALAR{U}_{k, l}} - \sum_{i = 1}^n \sum_{j = 1}^m \SCALAR{E}_{i, j}
        \end{align}

        Since the denominator exactly equals to the left term, by multiplying the denominator we have:
        \begin{align}
            \left(\sum_{i = 1}^n \sum_{j = 1}^m \SCALAR{U}_{i, j}\right)^2 = \sum_{i = 1}^n \sum_{j = 1}^m \sum_{l = 1}^m \SCALAR{U}_{i, l}\sum_{k = 1}^n \SCALAR{U}_{k, j} - \sum_{l = 1}^m\sum_{k = 1}^n \SCALAR{U}_{k, l} \sum_{i = 1}^n \sum_{j = 1}^m \SCALAR{E}_{i, j}
        \end{align}

        By rearranging the above equation, we have:
        \begin{align}
            \left(\sum_{i = 1}^n \sum_{j = 1}^m \SCALAR{U}_{i, j}\right)^2 - \sum_{i = 1}^n  \sum_{l = 1}^m \SCALAR{U}_{i, l} \sum_{k = 1}^n \sum_{j = 1}^m \SCALAR{U}_{k, j} = - \sum_{l = 1}^m\sum_{k = 1}^n \SCALAR{U}_{k, l} \sum_{i = 1}^n \sum_{j = 1}^m \SCALAR{E}_{i, j}
        \end{align}

        Since the left term is zero, the right term becomes zero.
        \begin{align}
            0 = - \sum_{l = 1}^m\sum_{k = 1}^n \SCALAR{U}_{k, l} \sum_{i = 1}^n \sum_{j = 1}^m \SCALAR{E}_{i, j} 
        \end{align}
        Since the summation of all elements in $\MATRIX{U}$ is not zero, the only solution is:
        \begin{align}
            \sum_{i = 1}^n \sum_{j = 1}^m \SCALAR{E}_{i, j} = 0
        \end{align}
    \end{proof}
\end{lemma}

\begin{lemma}
    \cite{amsgrad} For all $y_t \in \RSET^+, \forall{t}\in[T]$, the following is satisfied \label{lemma:the_inquality_for_norm01}
    \begin{align}
        \sum_{i=1}^t\dfrac{y_i}{\sqrt{\sum_{j=1}^iy_i}}\le 2\sqrt{\sum_{i=1}^ty_i}
    \end{align}
\end{lemma}

\section{Hyper-parameters and Sensitivity}
\paragraph{Learning-rate.} Throughout extensive experiment, we found that the proper learning-rate of \optim follows the learning-rate of Adam~\cite{adam}, i.e.,  0.001. This is because the convergence proof of \optim follows the similar proof of Adam and the proof shows that the convergence of \optim is similar to the variant of Adam. That means, the other hyper-parameters such as weight-decay can be similar to that of \optim. In fact, the weight-decay, which is similar to Adam, is used for \optim.

\paragraph{Decay-rate.} We observe that \optim is sensitive to $\gamma$, decay-rate in \Cref{config:training_configurations}. From several experiments, we observe that as $\gamma$ approaches -0.5, the stability and performance tends to improve. However, the performance depends on the model architecture. The recommended range for $\gamma$ is between -0.5 and -0.8, with -0.8 being used in Adafactor~\cite{adafactor}. Throughout the experiments on CNN models and Transformers based models, the suitable decay-rate is -0.5 for CNN and -0.8 for Transformer based models.

\paragraph{Weight-decay.} The provided code implements \optim with weight-decay used in Adam~\cite{adam} and weight-decay used in AdamW~\cite{adamw}. From several experiments, we observe that the weight-decay is 1000 to 2000 times smaller than AdamW when using Adam's weight-decay method. If you use huge weight-decay, the optimizers, i.e., Adam, Adafactor, SM3, CAME, and \optim tend to show loss spike.

\section{Optimization Temporal Memory}
\label{sec:temporal_memory}
\begin{definition}
    Temporary Variable is necessary for the operation of an algorithm, but it can safely disappear entirely when a step of the algorithm is completed.
\end{definition}

The temporary variables themselves are considered overhead and are not taken into account when measuring optimizer memory complexity.

\begin{definition}
    Temporary Memory refers to the memory allocated to store temporary variables, which is released once those variables disappear.
\end{definition}
In neural network optimization algorithms, temporary variables are removed and memory allocation is freed after updating one step, i.e., one layer. In \Cref{alg:brief_algorithm}, $\bar{\MATRIX{G}}_t, \hat{\MATRIX{M}_{t-1}}, \hat{\MATRIX{V}}_{t-1}, \MATRIX{M}_t, \MATRIX{V}_t$, and $\MATRIX{U}$ represent temporary variables, which are cleared from memory after updating one layer. In the case of \optim, when computing the update term $\MATRIX{U}$, operations are carried out in an inplace manner, resulting in temporary memory usage equivalent to adaptive learning-rate optimizers, e.g., Adam~\cite{adam}, Adafactor~\cite{adafactor}, SM3~\cite{sm3}, and CAME~\cite{came} since the four optimizers have update term, $\MATRIX{U}$ having same shape of $\MATRIX{W}$,s to update the weight tensor, matrix, and vector, $\MATRIX{W}$. However, when it comes to non-temporary variables, i.e., optimizer memory (optimizer state), the memory footprint of the four optimizer differs. Memory-efficient optimizers, i.e., Adafactor, SM3, and CAME reduce the non-temporary variables (optimizer state) so that the memory complexity of them is less than Adam. In the same vein, \optim can be considered memory-efficient because it uses significantly less memory than Adam, Adafactor, SM3, and CAME, particularly in high-rank tensor situations, for non-temporary variables, i.e., $\VECTOR{r}_{\{M, V\}}, \VECTOR{c}_{\{M, V\}}$, and $\MATRIX{S}_{\MATRIX{M}}$, resulting reduced memory complexity.

\section{Low Rank Compression}
In this section, we use the meaning of "rank" as used in linear algebra, i.e., "rank of a matrix", rather than the meaning associated with dimensions used in the main text.

Deep neural networks such as CNN or Transformers have improved model performance by increasing the complexity of the model structure, the number of the learnable parameters, and the depth of layers. However, these deep neural network models are prone to be over-parameterized. Recent papers~\cite{sm3,lora,low_dim01,low_dim02} have pointed out the over-parameterization of deep neural network models, demonstrating that fully trained over-parameterized models in fact reside in low-rank spaces.

Adafactor, SM3, and CAME have shown that beyond low-rank compression, through compression algorithms, they can achieve optimization performance comparable to Adam using rank-1 compression. Since low rank compression where the rank is bigger than $1$ takes more memory compared to rank-1 compression, those optimizers which use rank-1 compression effectively reduce more memory footprint. Inspired by the rank-1 compression, \optim can achieve optimization performance comparable to Adam like rank-1 memory-efficient optimizers, i.e., Adafactor, SM3, and CAME, while reducing more memory usage by introducing proposed square-matricization algorithm and momentum compression algorithm. 

Even though the model parameter is not in low rank space, unlike the previous optimizers, i.e., Adafactor, SM3, and CAME, which use $compression \rightarrow decompression$ scheme, \optim can alleviate the degrade of the model performance by using $decompression \rightarrow compression$ scheme fully adding the current gradient (i.e., full rank matrix) to the weight matrix using much lower optimizer state memory with same temporal memory defined at \Cref{sec:temporal_memory}.

\section{Condition of Non-Negative Matrix Factorization}
Since Non-Negative Matrix Factorization (NNMF) doesn't need strong condition such as grid pattern in weight matrix being the base condition of SM3~\cite{sm3}. The only condition of NNMF regarding the pattern of weight parameter is specified in \Cref{thm:nnmf-pattern-condition}. From the theorem, given a non-negative matrix $\MATRIX{U} \in \RSET^{n \times m}$, the only condition under which NNMF fails is $\MATRIX{U}_{i, j} = 0$ or almost zero (underflow) for all $(i, j)$ elements, which is practically an impossible condition during training steps since the the probability is extremely low (e.g., $2^{-32 \times n \times m}$) except for the initial step of \Cref{alg:brief_algorithm}.

\begin{theorem}
\label{thm:nnmf-pattern-condition}
Given non-negative matrix $\MATRIX{U} \in \RSET^{n \times m}$, the only condition that the compressed matrix $\hat{\MATRIX{U}}$ becomes $\MATRIX{0}$ is $\MATRIX{U}_{i, j} = 0$ for all $(i, j)$ elements.
    \begin{proof}
        From \Cref{alg:nnmf} (NNMF) we can write $\hat{\MATRIX{U}}_{i, j}$ as:
        \begin{align}
            \hat{\SCALAR{U}}_{i, j} &= \dfrac{\sum_{l = 1}^m \SCALAR{U}_{i, l}\sum_{k = 1}^n \SCALAR{U}_{k, j}}{\sum_{l = 1}^m\sum_{k = 1}^n \SCALAR{U}_{k, l}}
        \end{align}
        If and only if the $\hat{\MATRIX{U}}$ is $\MATRIX{0}$, the summation of all elements in the matrix should be $0$ since the matrix is non-negative matrix.
        \begin{align}
            \sum_{i = 1}^n \sum_{j = 1}^m \hat{\SCALAR{U}}_{i, j} = \sum_{i = 1}^n \sum_{j = 1}^m  \dfrac{\sum_{l = 1}^m \SCALAR{U}_{i, l}\sum_{k = 1}^n \SCALAR{U}_{k, j}}{\sum_{l = 1}^m\sum_{k = 1}^n \SCALAR{U}_{k, l}} = 0
        \end{align}
        Focusing on the denominator, we have
        \begin{align}
            \sum_{i = 1}^n \sum_{j = 1}^m \sum_{l = 1}^m \SCALAR{U}_{i, l} \sum_{k = 1}^n \SCALAR{U}_{k, j} &= 0 \\
            \sum_{i = 1}^n \sum_{l = 1}^m \SCALAR{U}_{i, l} \sum_{j = 1}^m \sum_{k = 1}^n \SCALAR{U}_{k, j} &= 0 \\
            \left( \sum_{i = 1}^n \sum_{j = 1}^m \SCALAR{U}_{k, j} \right )^2&= 0
        \end{align}
        Since $\MATRIX{U}$ is non-negative matrix, the only condition is $U_{i, j} = 0$ for all $(i, j)$ elements in $\MATRIX{U}$. Even the above equation violates the condition of denominator, it also means that the only condition under which NNMF fails is the above equation.
    \end{proof}
\end{theorem}

\section{Dataset}
In the experiment, we use representative datasets, i.e., ImageNet-1k~\cite{imagenet} and CIFAR100~\cite{cifar} for image classification task. ImageNet-1k which is one of the largest image dataset consists of over 14 million images, and the images are classified into 1k classes with different image size. CIFAR100 consists of around 60,000 images, and the images are classified into 100 classes with same image size, 32x32. For the object detection task, we use COCO~\cite{lin2015microsoft} consisting of over 118,000 images with annotation. The three datasets are representative datasets in image modality.
WMT32k~\cite{bojar2014findings_wmt32k} which is representative De-En translation dataset and used for full-training is consists of multiple translation datasets, e.g., News Commentary v13\footnote{http://data.statmt.org/wmt18/translation-task/training-parallel-nc-v13.tgz}, Europarl v7\footnote{https://www.statmt.org/wmt13/training-parallel-europarl-v7.tgz}, Common Crawl corpus\footnote{https://www.statmt.org/wmt13/training-parallel-commoncrawl.tgz} originated from tensor2tensor github\footnote{https://github.com/tensorflow/tensor2tensor}. BookCorpus~\cite{bookcorpus} \& Wikipedia dataset used for pre-training is a representative text dataset that consists of 11,038 book dataset wikipedia English text data. QNLI, MNLI, QQP, STSB, and MRPC~\cite{glue} dataset are sub-datasets of GLUE~\cite{glue} used for fine-tuning (text-classification task). SQuAD~\cite{squad} and SQuADv2~\cite{squadv2} are datasets used for fine-tuning (question \& answering task). WMT16 Ro-En~\cite{wmt16} which is a representative translation dataset and used for fine-tuning T5-small~\cite{t5} and Marian which is a variant of BART~\cite{bart} where the layer normalization at embedding layer is removed. Il-Post~\cite{ilpost}~\cite{ilpost} used for summarization task contains news articles taken from IlPost and consists of multiple languages. Fanpage~\cite{fanpage} which is a kind of multilingual news article dataset from Fanpage used for multilingual summarization task is used for fine-tuning mBART~\cite{mbart}. Alpaca~\cite{alpaca} is a dataset consisting of over 50k human instruction pairs. It commonly used for tuning the Large-Language Model for human instructions such as "What is the capital of French?". The reasons for all the datasets used in the experiments are 1) publicly accessible, 2) sufficient amount of samples to check the generalization performance of the five optimizers in the main context, and representative datasets used in each task, i.e., image classification, object detection, translation, text pre-training, text classification, question-answering, and summarization.

\section{Finetuning performance}
\label{sec:additional_fine_tuning_performance}
In this section, we show the additional experiment on various datasets, tasks, and transformer based models. In most of cases, \optim shows comparable performance with lowest optimization memory and lowest end-to-end memory including 1-bit $\MATRIX{S}_{\MATRIX{M}}$. We fine-tune LLaMA-7b~\cite{llama1} on COLA and RTE dataset~\cite{glue} (text-classification task), and RoBERTa~\cite{roberta}, ALBERT base v2~\cite{albert}, BERT~\cite{bert}, and GPT-2~\cite{gpt2} on SQuAD~\cite{squad} (question-answering task), and T5-small~\cite{t5} on SQuADv2~\cite{squadv2} (question-answering task), and T5-small, and MarianMT which is a variant of BART~\cite{bart} where the layer normalization at embedding layer is removed on WMT16 En-to-Ro (translation task), and T5-small, and  BART-base on CNN/Daily Mail~\cite{cnn_dailymail} and XSUM~\cite{xsum}, and mBART~\cite{mbart} on Il-Post~\cite{ilpost} and Fanpage~\cite{fanpage} (multilingual summarization task) dataset. 

Additionally, we show the performance of fine-tuned BERT on QNLI, MNLI, QQP, STSB, and MRPC as an extension of \Cref{tab:memory_footprint_and_result_of_fine_tuning} (See \Cref{tab:memory_footprint_and_result_of_fine_tuning2})

\begin{table}[H]
    \setlength\tabcolsep{1.5pt}
    \begin{center}
            \begin{tabular}{l||c|c@{\ninefont (}r@{\ninefont , }r@{\ninefont ) }r|c@{\ninefont (}r@{\ninefont , }r@{) }r|c@{\ninefont (}r@{\ninefont , }r@{\ninefont ) }r|c@{\ninefont (}r@{\ninefont , }r@{\ninefont ) }r|c@{\ninefont (}r@{\ninefont , }r@{\ninefont ) }r}
            \toprule
            \multicolumn{22}{c}{\textbf{Transformer Models and Tasks (Fine-Tuning)}} \\
            \multicolumn{22}{c}{(Optimizer Memory [MiB] and End-to-End Memory [GiB]), Model Performance} \\
            \toprule
            \textbf{\ninefont Optimizer} &  \textbf{\ninefont Model}       &   \multicolumn{4}{c|}{\textbf{\ninefont QNLI (ACC)}}    &   \multicolumn{4}{c|}{\textbf{\ninefont MNLI (ACC)}}    &   \multicolumn{4}{c|}{\textbf{\ninefont QQP (ACC)}}    &   \multicolumn{4}{c|}{\textbf{\ninefont STSB (Pearson)}
            }    &   \multicolumn{4}{c}{\textbf{\ninefont MRPC (ACC)}} \\
                        
            \bottomrule
            \ninefont Adam    &           && \ninefont    849 & \ninefont     1.65    & \ninefont    91.0    && \ninefont    837 & \ninefont     1.65    & \ninefont    82.5    && \ninefont    848 & \ninefont     1.65    & \ninefont    89.8    && \ninefont    846 & \ninefont     1.65    & \ninefont    89.7    && \ninefont    854 & \ninefont     1.65    & \ninefont    86.5 \\
            \ninefont Adafactor&          && \ninefont    425 & \ninefont    1.24    & \ninefont    90.4    && \ninefont    420 & \ninefont    1.24    & \ninefont    80.4    && \ninefont    424 & \ninefont    1.24    & \ninefont    88.8    && \ninefont    424 & \ninefont    1.24    & \ninefont    88.5    && \ninefont    428 & \ninefont    1.24    & \ninefont    84.3 \\
            \ninefont SM3     & \ninefont    BERT    && \ninefont    425 & \ninefont    1.24    & \ninefont    90.8    && \ninefont    420 & \ninefont    1.24    & \ninefont    83.7    && \ninefont    423 & \ninefont    1.24    & \ninefont    90.9    && \ninefont    424 & \ninefont    1.24    & \ninefont    88.3    && \ninefont    428 & \ninefont    1.24    & \ninefont    84.6 \\
            \ninefont CAME    &           && \ninefont    426 & \ninefont    1.24    & \ninefont    81.9    && \ninefont    421 & \ninefont    1.24    & \ninefont    84.4    && \ninefont    425 & \ninefont    1.24    & \ninefont    91.3    && \ninefont    425 & \ninefont    1.24    & \ninefont    89.0    && \ninefont    429 & \ninefont    1.24    & \ninefont    82.8 \\
            \ninefont \optim  &           &&   \textbf{\ninefont 15} &   \textbf{\ninefont 0.83}    & \ninefont    91.8    &&   \textbf{\ninefont 15} &    \textbf{\ninefont 0.83}    & \ninefont    84.1    &&   \textbf{\ninefont 15} &   \textbf{\ninefont 0.83}    & \ninefont    91.8    &&   \textbf{\ninefont 15} &\textbf{\ninefont 0.83}    & \ninefont    88.8    &&   \textbf{\ninefont 15} &   \textbf{\ninefont 0.83}    & \ninefont    85.0  
            \end{tabular}
    \end{center}
    \caption{\textbf{Fine-tuning}: the optimizer and end-to-end training (one-batch) memory usage [MiB]/[GiB] at 100 iterations including $\MATRIX{S}_{\MATRIX{M}}$, and the performance of BERT fine-tuned on QNLI, MNLI, QQP, STSB, and MRPC.
    }
    \label{tab:memory_footprint_and_result_of_fine_tuning2}
\end{table}

\begin{table}[H]
    \centering
    \begin{tabular}{c|c@{(}r@{, }r@{) }c|c@{(}r@{, }r@{) }c|c@{(}r@{, }r@{) }c|c@{(}r@{, }r@{) }c|c@{(}r@{, }r@{) }c}
        \toprule
        \multicolumn{21}{c}{Training LLaMA-7b on COLA and RTE using LoRA}  \\
        \multicolumn{21}{c}{(Optimizer Memory [MiB] and End-to-End Memory [GiB]), Model Performance}  \\

        \midrule
        Dataset   & \multicolumn{4}{c|}{Adam}   & \multicolumn{4}{c|}{Adafactor}  & \multicolumn{4}{c|}{SM3}  & \multicolumn{4}{c|}{CAME} & \multicolumn{4}{c}{\optim}   \\
        
        \bottomrule
        COLA (Matthew correlation)  && 153  & 24.9 & 66.1  && 86    & 24.9  & 60.4  && 86   & 24.9  & x  && 96  & 24.9 & x  && \textbf{3.9}  & \textbf{24.8} &   67.0   \\
        RTE (ACC)  && 153   & 24.9  & 85.2  && 86   & 24.9  & 88.3  && 86   & 24.9  & 56.3  && 96   & 24.9  & 52.3  && \textbf{3.9}  & \textbf{24.8}  &   85.9   \\
    \end{tabular}
    \caption{The optimizer memory (MiB) including $\MATRIX{S}_{\MATRIX{{M}}}$, end-to-end memory training (one-batch) memory usage at 100 iterations, and the test performance of LLaMA-7b~\cite{llama1} fine-tuned on the two datasets COLA and RTE~\cite{glue} using LoRA~\cite{lora}. The x means that the model performance is lower than 20\%. We train the model using learning-rate 5E-5 and 1E-4, which are normal LoRA learning-rate during the five epochs, and choose the best performance.}
    \label{tab:additional_exp_llama_glue}
\end{table}

\begin{table}[H]
    \centering
    \begin{tabular}{c|c@{(}r@{, }r@{) }r|c@{(}r@{, }r@{) }r|c@{(}r@{, }r@{) }r|c@{(}r@{, }r@{) }r|c@{(}r@{, }r@{) }r}
        \toprule
        \multicolumn{21}{c}{Training RoBERTa, ALBERT, BERT, and GPT-2 on SQuAD}  \\
        \multicolumn{21}{c}{(Optimizer and End-to-End Memory [MiB]), Model Performance (F1)}  \\

        \midrule
        Model   & \multicolumn{4}{c|}{Adam}   & \multicolumn{4}{c|}{Adafactor}  & \multicolumn{4}{c|}{SM3}  & \multicolumn{4}{c|}{CAME} & \multicolumn{4}{c}{\optim}   \\
        
        \bottomrule
        RoBERTa && 972  & 1440  & 91.7  && 488  & 967   & 91.7  && 488  & 968   & 89.8  && 488  & 967   & 91.7  && \textbf{16.3} & \textbf{507}   & 91.8  \\
        ALBERT base v2  && 85   & 146   & 90.7  && 43   & 102   & 90.9  && 43   & 102   & 86.5  && 43   & 103   & 89.6  && \textbf{1.5}  & \textbf{61}  & 90.8  \\
        BERT  && 856   & 1270  & 86.5   && 430   & 850  & 86.4  && 430  & 850   & 84.9  && 430  & 850   & 87.0  && \textbf{14}  & \textbf{450}  & 86.6  \\
        GPT-2 && 957   & 1454  & 78.1   && 483   & 983  & 78.4  && 483  & 983   & 72.9  && 484  & 983   & 78.0 && \textbf{16}  & \textbf{522}  & 79.1 
    \end{tabular}
    \caption{The optimizer memory (MiB) including $\MATRIX{S}_{\MATRIX{{M}}}$, end-to-end memory training (one-batch) memory usage at 100 iterations, and the test performance of RoBERTa~\cite{roberta}, ALBERT~\cite{albert}, BERT~\cite{bert}, and GPT-2~\cite{gpt2} fine-tuned on SQuAD~\cite{squad}. We use pre-trained RoBERTa from (\url{https://huggingface.co/FacebookAI/roberta-base}), and ALBERT from (\url{https://huggingface.co/albert/albert-base-v2})}
    \label{tab:additional_exp_BERT_family_glue}
\end{table}

\begin{table}[H]
    \centering
    \begin{tabular}{c|c@{(}r@{, }r@{) }r|c@{(}r@{, }r@{) }r|c@{(}r@{, }r@{) }r|c@{(}r@{, }r@{) }r|c@{(}r@{, }r@{) }r}
        \toprule
        \multicolumn{21}{c}{Training T5-small on SQuADv2}  \\
        \multicolumn{21}{c}{(Optimizer and End-to-End Memory [MiB]), Model Performance (Perplexity)}  \\

        \midrule
        Model   & \multicolumn{4}{c|}{Adam}   & \multicolumn{4}{c|}{Adafactor}  & \multicolumn{4}{c|}{SM3}  & \multicolumn{4}{c|}{CAME} & \multicolumn{4}{c}{\optim}   \\
        
        \bottomrule
        T5-small && 463  & 706   & 2.013 && 233  & 481   & 2.013  && 234  & 481   & 2.261  && 234  & 481   & 2.172  && \textbf{8} & \textbf{256}   & 2.013  \\
    \end{tabular}
    \caption{The optimizer memory (MiB) including $\MATRIX{S}_{\MATRIX{{M}}}$, end-to-end memory training (one-batch) memory usage at 100 iterations, and the test performance of T5-small~\cite{t5} fine-tuned on SQuADv2~\cite{squadv2}.}
    \label{tab:additional_exp_t5_small_glue}
\end{table}

\begin{table}[H]
    \centering
    \begin{tabular}{@{\quad\quad\quad}c@{\quad\quad\quad}|@{\quad\quad\quad}c@{(}r@{, }r@{) }r@{\quad\quad\quad}|c@{\quad\quad(}r@{, }r@{) }r@{\quad\quad}}
        \toprule
        \multicolumn{9}{c}{Training T5-small and MarianMT on WMT16 En-to-Ro Translation Task}  \\
        \multicolumn{9}{c}{(Optimizer and End-to-End Memory [MiB]), Model Performance (BLEU score)}  \\

        \midrule
        Model   & \multicolumn{4}{c@{\quad\quad\quad}|}{Adam}   & \multicolumn{4}{c}{\optim}   \\
        
        \bottomrule
        T5-small && 462  & 719   & 26.7  && \textbf{8.3} & \textbf{265}   & 26.7  \\
        MarianMT && 569  & 875   & 27.0  && \textbf{10.2} & \textbf{316}  & 26.9  \\ 
    \end{tabular}
    \caption{The optimizer memory (MiB) including $\MATRIX{S}_{\MATRIX{{M}}}$, end-to-end memory training (one-batch) memory usage at 100 iterations, and the test performance of T5-small~\cite{t5} and MarianMT fine-tuned on WMT16 En-to-Ro Task~\cite{wmt16} during the 190740 steps using the learning-rate 0.00005 and batch size 64. MarianMT is a variant of BART~\cite{bart} where the layer normalization at embedding layer is removed.}
    \label{tab:additional_exp_T5_small_translation_en_ro}
\end{table}

\begin{table}[H]
    \centering
    \begin{tabular}{c|c|c|c|c|c|c}
        \toprule
        \multicolumn{7}{c}{Training T5-small on CNN/Daily Mail (Summarization)}  \\

        \midrule
        \multirow{2}{*}{Optimizer}  &   Optimizer       &   End-to-End      &   \multirow{2}{*}{ROUGE1}  &  \multirow{2}{*}{ROUGE2}  & \multirow{2}{*}{ROUGE-L}    & \multirow{2}{*}{ROUGE-L-Sum}   \\
                                    &   Memory [MiB]    &   Memory [MiB]    &&& \\
        \bottomrule
        Adam    &   462 &   750 &   41.4    &   19.0    &   29.3    &   38.6    \\
        \optim  &   \textbf{8.3}   &   \textbf{294}   &   41.5    &   19.1    &   29.4    &   38.7    \\
        \hline
        
        \toprule
        \multicolumn{7}{c}{Training T5-small on XSUM (Summarization)}  \\
        \bottomrule
        Adam    &   462 &   740 &   34.5    &   11.9    &   27.1    &   27.1    \\
        \optim  &   \textbf{8.3}   &   \textbf{260}   &   34.5    &   12.0    &   27.2    &   27.1    \\
    \end{tabular}
    \caption{The optimizer memory (MiB) including $\MATRIX{S}_{\MATRIX{{M}}}$, end-to-end memory training (one-batch) memory usage at 100 iterations, and the test performance of T5-small~\cite{t5} fine-tuned on CNN/Daily Mail~\cite{cnn_dailymail} during 107670 steps and XSUM~\cite{xsum} during 255060 steps using the learning-rate 0.00005 and batch size 8.}
    \label{tab:additional_exp_T5_small_summarization1}
\end{table}

\begin{table}[H]
    \centering
    \begin{tabular}{c|c|c|c|c|c|c}
        \toprule
        \multicolumn{7}{c}{Training BART-base on CNN/Daily Mail (Summarization)}  \\

        \midrule
        \multirow{2}{*}{Optimizer}  &   Optimizer       &   End-to-End      &   \multirow{2}{*}{ROUGE1}  &  \multirow{2}{*}{ROUGE2}  & \multirow{2}{*}{ROUGE-L}    & \multirow{2}{*}{ROUGE-L-Sum}   \\
                                    &   Memory [MiB]    &   Memory [MiB]    &&& \\
        \bottomrule
        Adam    &   1068    &   1639    &   43.5    &   20.1    &   30.0    &   40.8    \\
        \optim  &   \textbf{18.5}   &   \textbf{582}   &   43.4    &   19.9    &   30.1    &   40.8    \\
        \hline
        
        \toprule
        \multicolumn{7}{c}{Training BART-base on XSUM (Summarization)}  \\
        \bottomrule
        Adam    &   1071 &   1620 &   40.6    &   17.8    &   32.9    &   32.9    \\
        \optim  &   \textbf{18.5}   &   \textbf{575}   &   40.8    &   17.8    &   32.8    &   32.8    \\
    \end{tabular}
    \caption{The optimizer memory (MiB) including $\MATRIX{S}_{\MATRIX{{M}}}$, end-to-end memory training (one-batch) memory usage at 100 iterations, and the test performance of BART-base~\cite{bart} fine-tuned on CNN/Daily Mail~\cite{cnn_dailymail} during 717800 steps and XSUM~\cite{xsum} during 510120 steps using the learning-rate 0.00005 and batch size 8.}
    \label{tab:additional_exp_T5_small_summarization2}
\end{table}

\begin{table}[H]
    \centering
    \begin{tabular}{c|c|c|c|c|c|c}
        \toprule
        \multicolumn{7}{c}{Training mBART on Il-Post (Summarization)}  \\

        \midrule
        \multirow{2}{*}{Optimizer}  &   Optimizer       &   End-to-End      &   \multirow{2}{*}{ROUGE1}  &  \multirow{2}{*}{ROUGE2}  & \multirow{2}{*}{ROUGE-L}    & \multirow{2}{*}{ROUGE-L-Sum}   \\
                                    &   Memory [MiB]    &   Memory [MiB]    &&& \\
        \bottomrule
        Adam    &   4661    &   7046    &   41.0    &   24.4    &   34.6    &   37.4    \\
        \optim  &   \textbf{77.8}   &   \textbf{2477}   &   41.3    &   24.4    &   34.8    &   37.6    \\
        \hline
        
        \toprule
        \multicolumn{7}{c}{Training mBART on Fanpage (Summarization)}  \\
        \bottomrule
        Adam    &   4663 &   7093 &   37.7    &   19.0    &   27.8    &   31.7    \\
        \optim  &   \textbf{77.5}   &   \textbf{2499}   &   37.6    &   18.9    &   27.7    &   31.6    \\
        
    \end{tabular}
    \caption{The optimizer memory (MiB) including $\MATRIX{S}_{\MATRIX{{M}}}$, end-to-end memory training (one-batch) memory usage at 100 iterations, and the test performance of mBART~\cite{mbart} fine-tuned on Il-Post~\cite{ilpost} during 88020 steps and Fanpage~\cite{fanpage} using the learning-rate 0.00005 and batch size 8.}
    \label{tab:additional_exp_T5_small_summarization3}
\end{table}

\begin{figure}[H]
    \centering
    \includegraphics[width=0.45\linewidth]{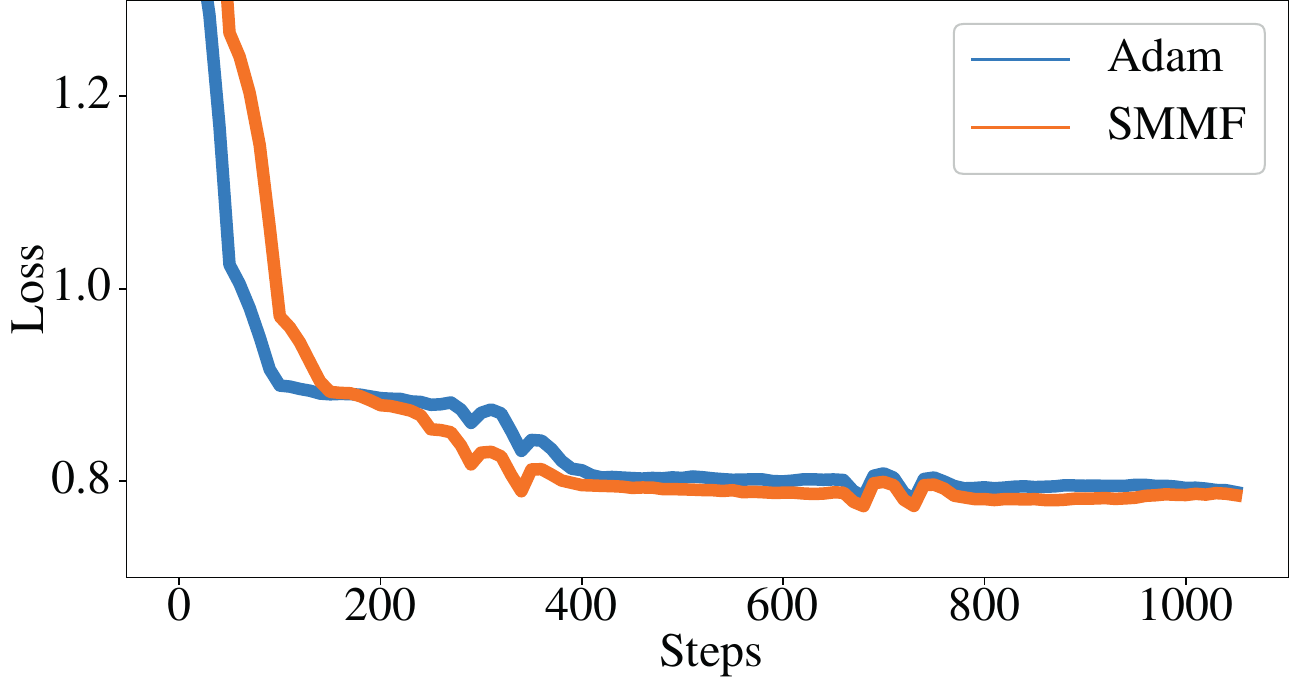}
    \caption{The loss graph of LLaMA-2-7B with LoRA fine-tuned on Alpaca dataset during the 1000 steps. The blue line indicates the Adam's performance and the oriange line indicates the \optim's performance.}
    \label{fig:llama2-Alpaca}
\end{figure}
\Cref{fig:llama2-Alpaca} shows the loss graph of LLaMA-2-7B~\cite{llama2} fine-tuned on Alpaca~\cite{alpaca} with LoRA~\cite{lora} during the 1160 steps. Adam~\cite{adam} shows better performance (i.e., lower loss) than \optim at the initial few steps, but \optim shows comparable/better performance after the initial few steps with much less memory consumption. 

\section{Training Configurations}
\label{config:training_configurations}
Since there are two types of weight decay: 1) Adam~\cite{adam} and 2) AdamW~\cite{adamw}, we implement the two weight decay method (\Cref{alg:weight_decay_adam,alg:weight_decay_adamw}). In this training configurations, we follow the weight decay method of Adam (\Cref{alg:weight_decay_adam}). As default schedulers for $\beta_{1, t}$ and $\beta_{2, t}$, we use the \Cref{alg:scheduling}. The scheduler for $\beta_{1,t}$ is from AdamNC~\cite{adam} and the scheduler for $\beta_{2,t}$ is from Adafactor~\cite{adafactor} for stable training. The rest of tables are the training configurations using the default $beta$ schedulers. To train LLaMA-7b~\cite{llama1}, we train the model using LoRA~\cite{lora} with two learning-rate 5E-5 and 1E-4 which are default LoRA fine-tuning learning-rate, and choose the best result. We conduct the experiments on Ubuntu OS using PyTorch~\cite{pytorch} framework and NumPy~\cite{numpy}.

\begin{algorithm}[H]
    \caption{Adam's weight decay method.}
    \label{alg:weight_decay_adam}
    \begin{algorithmic}
        \STATE {\bfseries Input:} Weight decay coefficient ($c$), weight parameter $\WEIGHT_t$, and gradient $\MATRIX{G}_t$ at step t.

        \STATE Take gradient $\MATRIX{G}_t$
        \STATE $\MATRIX{G}_t \leftarrow \MATRIX{G}_t + c \times \WEIGHT_t$
        \STATE Update momentums and weight
    \end{algorithmic}
\end{algorithm}

\begin{algorithm}[H]
    \caption{AdamW's weight decay method.}
    \label{alg:weight_decay_adamw}
    \begin{algorithmic}
        \STATE {\bfseries Input:} Weight decay coefficient ($c$), weight parameter $\WEIGHT_t$ at step t.
        
        \STATE Take gradient
        \STATE $\WEIGHT_{t} \leftarrow \WEIGHT_t - c \times \WEIGHT_t$
        \STATE Update momentums and weight
    \end{algorithmic}
\end{algorithm}

\begin{algorithm}[H]
    \caption{The default schedulers for $\beta_{1, t}$ and $\beta_{2, t}$}
    \label{alg:scheduling}
    \begin{algorithmic}
        \STATE {\bfseries Input:}  Current step $t$, $1^{st}$ momentum coefficient $\beta_1$, $1^{st}$ momentum growth rate $\lambda$, $2^{nd}$ momentum decay- rate $\gamma$. The recommended growth rate is 0.999 and the decay-rate is -0.5 for CNN models (stable learning) and -0.8 for Transformer based models (better performance)
        
        \STATE $\beta_{1t} = \beta_1\lambda^{t-1}$
        \STATE $\beta_{2t} = 1 - t^{\gamma}$
    \end{algorithmic}
\end{algorithm}

\input{tables/configs_cnn_cifar100}
\input{tables/configs_cnn_imagenet}
\input{tables/config_yolov5}
\input{tables/configs_transformer_wmt32k}
\input{tables/configs_transformer_bookcorpus}
\input{tables/configs_transformer_gleu}

\clearpage
\section{Code}
\label{sec:code}
We implement \optim using PyTorch framework at github\footnote{https://anonymous.4open.science/r/SMMF-D2E8}. Here is the code:
\begin{lstlisting}
import torch
from typing import Optional

class SMMF(torch.optim.Optimizer):
    def __init__(
        self, params, lr:float=1e-3, beta:Optional[float]=0.9, eps:float=1e-8, weight_decay:float=0.0,
        decay_rate:float=-0.5, growth_rate:Optional[float]=0.999, vector_reshape:bool=True, weight_decay_mode='adamw'
    ):
        if not (0.0 <= lr):
            raise ValueError(f'Learning-rate should be greater than or equal to 0.0. Current: {lr}')
        if beta != None:
            if not (0.0 <= beta and beta <= 1.0):
                raise ValueError(f'Beta should be in [0.0, 1.0]. Current: {beta}')
        if not (0.0 <= eps):
            raise ValueError(f'Epilson should be greater than or equal to 0.0. Current: {eps}')
        if not (0.0 <= weight_decay):
            raise ValueError(f'Weight-decay should be greater than or equal to 0.0. Current: {weight_decay}')
        if not (-1.0 <= decay_rate and decay_rate <= 0.0):
            raise ValueError(f'Decay-rate should be in [-1.0, 0.0]. Current: {decay_rate}')
        if not (0.0 <= growth_rate and growth_rate <= 1.0):
            raise ValueError(f'Growth-rate should be in [0.0, 1.0]. Current: {growth_rate}')
        
        if (beta == None) and (growth_rate != None):
            Warning("Beta is not used, but growth_rate is defined.")
        defaults = dict(
            lr=lr, beta=beta, eps=eps, weight_decay=weight_decay,
            decay_rate=decay_rate, growth_rate=growth_rate, vector_reshape=vector_reshape
        )
        super(SMMF, self).__init__(params, defaults)
        self.weight_decay_mode = weight_decay_mode
    
    @torch.no_grad()
    def step(self, closure=None):
        '''Implements SMMF Algorithm.
        Args:
            params (iterable): Iterable of parameters to optimize or dicts defining parameter groups
            lr: Learning-rate (default: 1e-3)
            beta: Coefficient used for computing running average of gradient (default: 0.9)
            eps: Regularization constant for square gradient (default: 1e-8)
            weight_decay: Weight-decay (default: 0.0)
            decay_rate: Decay-rate coefficient used for computing running average of gradient (default: -0.8)
            growth_rate: Growth-rate coefficient used for computing running average of square gradient (default: 0.999)
            vector_reshape: Vector Square-Matricization (default: True)
        '''
        loss = None
        if closure is not None:
            with torch.enable_grad():
                loss = closure()
            
        for group in self.param_groups:
            for params in group['params']:
                grad = params.grad
                if grad is None:
                    continue
                
                if group['weight_decay'] != 0.0 and self.weight_decay_mode == 'adam':
                    grad = grad.add(params, alpha=group['weight_decay'])
                elif group['weight_decay'] != 0.0 and self.weight_decay_mode == 'adamw':
                    params.mul_(1 - group['lr'] * group['weight_decay'])
                
                dimension = len(grad.squeeze().shape)
                factorization = not (dimension == 1 and (not group['vector_reshape']))
                
                if factorization:
                    update = self._factorized_adam(params, group)
                else:
                    update = self._adam(params, group)
                
                params.add_(update, alpha=-group['lr'])
        return loss
    
    @torch.no_grad()
    def _factorized_adam(self, params, group):
        beta = group['beta']
        eps = group['eps']
        decay_rate = group['decay_rate']
        growth_rate = group['growth_rate']
        
        grad = params.grad
        state = self.state[params]
        original_shape = grad.shape
        device = grad.device
        
        if len(state) == 0:
            state['step'] = 1
            state['effective_shape'] = self._get_effective_shape(params.numel())
            if beta != None:
                state['momentum_m'] = (
                    torch.zeros(state['effective_shape'][0], device=device),
                    torch.zeros(state['effective_shape'][1], device=device),
                )
                state['sign'] = torch.zeros(state['effective_shape'], dtype=torch.bool, device=device)              
            state['momentum_v'] = (
                torch.zeros(state['effective_shape'][0], device=device),
                torch.zeros(state['effective_shape'][1], device=device),
            )
        
        if not grad.is_contiguous():
            grad = grad.contiguous()
        grad = grad.view(state['effective_shape'])
        
        # Decompressing
        if beta != None:
            update_m = self._decompression(state, 'momentum_m')
        update_v = self._decompression(state, 'momentum_v')
        
        # Update
        if beta != None:
            beta_m = beta * growth_rate ** (state['step'] - 1.0)
            update_m.mul_(beta_m).add_(grad, alpha=(1.0 - beta_m))
        beta_v = 1.0 - state['step'] ** decay_rate
        update_v.mul_(beta_v).add_(grad ** 2, alpha=(1.0 - beta_v))
        
        # Compressing
        if beta != None:
            self._compression(update_m, state, 'momentum_m')
        self._compression(update_v, state, 'momentum_v')
        
        # Compute and Reshape
        if beta != None:
            update = update_m / (update_v.sqrt() + eps)
        else:
            update = grad / (update_v.sqrt() + eps)
        update = update.contiguous().view(original_shape)
        
        state['step'] += 1.0
        return update
    
    @torch.no_grad()
    def _adam(self, params, group):
        beta = group['beta']
        eps = group['eps']
        decay_rate = group['decay_rate']
        growth_rate = group['growth_rate']
        
        grad = params.grad
        state = self.state[params]
        
        if len(state) == 0:
            state['step'] = 1
            if beta != None:
                state['momentum_m'] = torch.zeros_like(params)
            state['momentum_v'] = torch.zeros_like(params)
            
        if beta != None:
            update_m = state['momentum_m']
        update_v = state['momentum_v']
        
        if beta != None:
            beta_m = beta * growth_rate ** (state['step'] - 1.0)
            update_m.mul_(beta_m).add_(grad, alpha=(1.0 - beta_m))
        beta_v = 1.0 - state['step'] ** decay_rate
        update_v.mul_(beta_v).add_(grad ** 2, alpha=(1.0 - beta_v))
        
        if beta != None:
            state['momentum_m'] = update_m
        state['momentum_v'] = update_v
        
        if beta != None:
            update = update_m / (update_v.sqrt() + eps)
        else:
            update = grad / (update_v.sqrt() + eps)
        
        state['step'] += 1.0
        return update
        
    @torch.no_grad()
    def _get_effective_shape(self, numel:int)->tuple:
        sqrt_num = int(numel ** 0.5) ** 2
        
        if numel == sqrt_num:
            sqrt_num = int(numel ** 0.5)
            return (sqrt_num, sqrt_num)
        
        reversed_range = reversed(range(1, int(numel **0.5) + 1))
        for i in reversed_range:
            if numel % i == 0:
                return (numel // i, i)
            
        return (numel, 1)
    
    @torch.no_grad()
    def _decompression(self, state, momentum:str)->torch.Tensor:
        update = self._unnmf(state[momentum])
        if momentum == 'momentum_m':
            sign = state['sign']
            if sign.dtype != torch.bool:
                sign = sign.type(torch.bool)
            torch.where(sign, update, -update, out=update)
        return update
        
    @torch.no_grad()
    def _compression(self, matrix:torch.Tensor, state, momentum:str)->tuple:
        if momentum == 'momentum_m':
            state['sign'] = matrix > 0
            self._nnmf(torch.abs(matrix), out=state[momentum] )
        else:
            self._nnmf(matrix, out=state[momentum])
        return state[momentum]
            
    @torch.no_grad()
    def _unnmf(self, row_col:tuple)->torch.Tensor:
        return torch.outer(row_col[0], row_col[1])

    @torch.no_grad()
    def _nnmf(self, matrix:torch.Tensor, out)->tuple:
        shape = matrix.shape
        torch.sum(matrix, dim=1, out=out[0])
        torch.sum(matrix, dim=0, out=out[1])

        if shape[0] < shape[1]:
            scale = out[0].sum()
            if scale != 0:
                torch.div(out[0], scale, out=out[0])
        else:
            scale = out[1].sum()
            if scale != 0:
                torch.div(out[1], scale, out=out[1])

        return out
\end{lstlisting}

%% file: algorithms/nnmf.tex
\begin{algorithm}[H]
    \caption{Non-Negative Matrix Factorization (NNMF)~\cite{adafactor}}
    \label{alg:nnmf}
    \begin{algorithmic}
        \STATE {\bfseries Input:} Non-negative matrix $\MATRIX{M} \in \RSET^{n \times m}$
        \STATE {\bfseries Output:} Factorized two vectors, $\VECTOR{r} \in \RSET^{n \times 1}$ and $\VECTOR{c} \in \RSET^{1 \times m}$

        \STATE $\VECTOR{r} = \MATRIX{M}\VECTOR{1}_{m}$
        \STATE $\VECTOR{c} = \VECTOR{1}_{n}^{\top} \MATRIX{M}$
        \STATE $\VECTOR{c} = \dfrac{\VECTOR{c}}{\VECTOR{c}\VECTOR{1}_{m}}$
    \end{algorithmic}
\end{algorithm}

%% file: tables/configs_cnn_cifar100.tex
\begin{table*}[H]
\begin{center}
        \begin{tabular}{l||l|l|l|l|l|l}
        \toprule
        \multicolumn{7}{c}{Training Configurations of MobileNetV2 and ResNet50 on CIFAR100} \\
        \multicolumn{7}{c}{Using Adam, Adafactor, SM3, CAME, and \optim}\\
        
        \midrule
        \multicolumn{1}{l||}{Model}      & Configurations & Adam                            & Adafactor & SM3      & CAME     & \optim           \\
        \bottomrule
                                       & epochs         & 200                             & 200       & 200      & 200      & 200                             \\
                                       & batch size     & 128                             & 128       & 128      & 128      & 128                             \\
                                       & warmup-steps   & 100                             & x         & 100      & 100      & 100                             \\
                                       & learning-rate  & 0.001                           & x         & 0.001    & 0.001    & 0.001                           \\
                                       & weight-decay   & 0.0005                          & 0.0005    & 0.0005   & 0.0005   & 0.0005                          \\
                                       & $\beta_1$      & 0.9                             & 0.9       & 0.9      & 0.9      & 0.9                             \\
                                       & $\beta_2$      & 0.999                           & x         & 0.999    & 0.999    & x                               \\
                                       & $\beta_3$      & x                               & x         & x        & 0.9999   & x                               \\
                                       & $\gamma$       & x                               & -0.8      & x        & x        & -0.5                            \\
                                       & $d$            & x                               & 1         & x        & 1        & x                               \\
                                       & $\lambda$      & x                               & x         & x        & x        & 0.999                           \\
                                       & $\epsilon_1$   & {\color[HTML]{1A1C1F} 1.00E-08} & 1.00E-30  & 1.00E-30 & 1.00E-30 & {\color[HTML]{1A1C1F} 1.00E-08} \\
        \multirow{-13}{*}{MobileNetV2} & $\epsilon_2$   & x                               & 1.00E-03  & x        & 1.00E-16 & x                               \\
        \hline
                                       & epochs         & 200                             & 200       & 200      & 200      & 200                             \\
                                       & batch size     & 128                             & 128       & 128      & 128      & 128                             \\
                                       & warmup-steps   & 100                             & x         & 100      & 100      & 100                             \\
                                       & learning-rate  & 0.001                           & x         & 0.001    & 0.001    & 0.001                           \\
                                       & weight-decay   & 0.0005                          & 0.0005    & 0.0005   & 0.0005   & 0.0005                          \\
                                       & $\beta_1$      & 0.9                             & 0.9       & 0.9      & 0.9      & 0.9                             \\
                                       & $\beta_2$      & 0.999                           & x         & 0.999    & 0.999    & x                               \\
                                       & $\beta_3$      & x                               & x         & x        & 0.9999   & x                               \\
                                       & $\gamma$       & x                               & -0.8      & x        & x        & -0.5                            \\
                                       & $d$            & x                               & 1         & x        & 1        & x                               \\
                                       & $\lambda$      & x                               & x         & x        & x        & 0.999                           \\
                                       & $\epsilon_1$   & {\color[HTML]{1A1C1F} 1.00E-08} & 1.00E-30  & 1.00E-30 & 1.00E-30 & {\color[HTML]{1A1C1F} 1.00E-08} \\
        \multirow{-13}{*}{ResNet50}    & $\epsilon_2$   & x                               & 1.00E-03  & x        & 1.00E-16 & x                              \\
        \hline
    \end{tabular}
\end{center}
\caption{The training configurations of MobileNetV2 and ResNet50 on CIFAR100 using Adam, Adafactor, SM3, CAME, and \optim. $\lambda$ means the growth-rate, $\gamma$ means the decay-rate, $d$ means the clip-threshold. We use cosine learning-rate scheduler for 200 epochs and use an A6000 GPU.}
\label{tab:configs_cnn_cifar100}
\end{table*}

%% file: tables/configs_cnn_imagenet.tex
\begin{table*}[p!]
\begin{center}
        \begin{tabular}{l||l|l|l|l|l|l}

        \toprule
        \multicolumn{7}{c}{Training Configurations of MobileNetV2 and ResNet50 on ImageNet} \\
        \multicolumn{7}{c}{Using Adam, Adafactor, SM3, CAME, and \optim}\\
        \midrule
        \multicolumn{1}{l||}{Model}      & Configurations & Adam                            & Adafactor & SM3      & CAME     & \optim           \\
        \bottomrule
                                       & epochs         & 500                             & 500       & 500      & 500      & 500                             \\
                                       & batch size     & 128                             & 128       & 128      & 128      & 128                             \\
                                       & warmup-epochs   & 1                             & x         & 1      & 1      & 1                             \\
                                       & learning-rate  & 0.001                           & x         & 0.001    & 0.001    & 0.001                           \\
                                       & weight-decay   & 0.0005                          & 0.0005    & 0.0005   & 0.0005   & 0.0005                          \\
                                       & $\beta_1$      & 0.9                             & 0.9       & 0.9      & 0.9      & 0.9                             \\
                                       & $\beta_2$      & 0.999                           & x         & 0.999    & 0.999    & x                               \\
                                       & $\beta_3$      & x                               & x         & x        & 0.9999   & x                               \\
                                       & $\gamma$       & x                               & -0.8      & x        & x        & -0.5                            \\
                                       & $d$            & x                               & 1         & x        & 1        & x                               \\
                                       & $\lambda$      & x                               & x         & x        & x        & 0.999                           \\
                                       & $\epsilon_1$   & {\color[HTML]{1A1C1F} 1.00E-08} & 1.00E-30  & 1.00E-30 & 1.00E-30 & {\color[HTML]{1A1C1F} 1.00E-08} \\
        \multirow{-13}{*}{MobileNetV2} & $\epsilon_2$   & x                               & 1.00E-03  & x        & 1.00E-16 & x                               \\
        \hline
                                       & epochs         & 500                             & 500       & 500      & 500      & 500                             \\
                                       & batch size     & 128                             & 128       & 128      & 128      & 128                             \\
                                       & warmup-epochs   & 1                             & x         & 1      & 1      & 1                             \\
                                       & learning-rate  & 0.001                           & x         & 0.001    & 0.001    & 0.001                           \\
                                       & weight-decay   & 0.0005                          & 0.0005    & 0.0005   & 0.0005   & 0.0005                          \\
                                       & $\beta_1$      & 0.9                             & 0.9       & 0.9      & 0.9      & 0.9                             \\
                                       & $\beta_2$      & 0.999                           & x         & 0.999    & 0.999    & x                               \\
                                       & $\beta_3$      & x                               & x         & x        & 0.9999   & x                               \\
                                       & $\gamma$       & x                               & -0.8      & x        & x        & -0.5                            \\
                                       & $d$            & x                               & 1         & x        & 1        & x                               \\
                                       & $\lambda$      & x                               & x         & x        & x        & 0.999                           \\
                                       & $\epsilon_1$   & {\color[HTML]{1A1C1F} 1.00E-08} & 1.00E-30  & 1.00E-30 & 1.00E-30 & {\color[HTML]{1A1C1F} 1.00E-08} \\
        \multirow{-13}{*}{ResNet50}    & $\epsilon_2$   & x                               & 1.00E-03  & x        & 1.00E-16 & x                              \\
        \hline
    \end{tabular}
\end{center}
\caption{The training configurations of MobileNetV2 and ResNet50 on ImageNet using Adam, Adafactor, SM3, CAME, and \optim. $\lambda$ means the growth-rate, $\gamma$ means the decay-rate, $d$ means the clip-threshold. We use \textit{ReduceLROnPlateau} scheduler for 500 epochs and use four 3090 GPUs.}
\label{tab:configs_cnn_imagenet}
\end{table*}

%% file: tables/config_yolov5.tex
\begin{table*}[p!]
\begin{center}
        \begin{tabular}{l||l|l|l|l|l|l}
        \toprule
        \multicolumn{7}{c}{Training Configurations of YOLOv5s and YOLOv5m on COCO} \\
        \multicolumn{7}{c}{Using Adam, Adafactor, SM3, CAME, and \optim}\\
        \midrule
        \multicolumn{1}{l||}{Model}  & Configurations & Adam                            & Adafactor & SM3      & CAME     & \optim                           \\
        \bottomrule
                                   & epochs         & 300                             & 300       & 300      & 300      & 300                             \\
                                   & batch size     & 64                              & 64        & 64       & 64       & 64                              \\
                                   & warmup-epochs  & 3                               & x         & 3        & 3        & 3                               \\
                                   & learning-rate0 & 0.001                           & x         & 0.001    & 0.001    & 0.001                           \\
                                   & weight-decay   & 0.0005                          & 0.0005    & 0.0005   & 0.0005   & 0.0005                          \\
                                   & $\beta_1$      & 0.937                           & 0.937     & 0.937    & 0.937    & 0.937                           \\
                                   & $\beta_2$      & 0.999                           & x         & 0.999    & 0.999    & x                               \\
                                   & $\beta_3$      & x                               & x         & x        & 0.9999   & x                               \\
                                   & $\gamma$       & x                               & -0.8      & x        & x        & -0.5                            \\
                                   & $d$            & x                               & 1         & x        & 1        & x                               \\
                                   & $\lambda$      & x                               & x         & x        & x        & 0.999                           \\
                                   & $\epsilon_1$   & {\color[HTML]{1A1C1F} 1.00E-08} & 1.00E-30  & 1.00E-30 & 1.00E-30 & {\color[HTML]{1A1C1F} 1.00E-08} \\
        \multirow{-13}{*}{YOLOv5s} & $\epsilon_2$   & x                               & 1.00E-03  & x        & 1.00E-16 & x                               \\
        \hline
                                   & epochs         & 200                             & 200       & 200      & 200      & 200                             \\
                                   & batch size     & 40                              & 40        & 40       & 40       & 40                              \\
                                   & warmup-epochs  & 3                               & x         & 3        & 3        & 3                               \\
                                   & learning-rate0 & 0.001                           & x         & 0.001    & 0.001    & 0.001                           \\
                                   & weight-decay   & 0.0005                          & 0.0005    & 0.0005   & 0.0005   & 0.0005                          \\
                                   & $\beta_1$      & 0.937                           & 0.937     & 0.937    & 0.937    & 0.937                           \\
                                   & $\beta_2$      & 0.999                           & x         & 0.999    & 0.999    & x                               \\
                                   & $\beta_3$      & x                               & x         & x        & 0.9999   & x                               \\
                                   & $\gamma$       & x                               & -0.8      & x        & x        & -0.8                            \\
                                   & $d$            & x                               & 1         & x        & 1        & x                               \\
                                   & $\lambda$      & x                               & x         & x        & x        & 0.99                            \\
                                   & $\epsilon_1$   & {\color[HTML]{1A1C1F} 1.00E-08} & 1.00E-30  & 1.00E-30 & 1.00E-30 & {\color[HTML]{1A1C1F} 1.00E-08} \\
        \multirow{-13}{*}{YOLOv5m} & $\epsilon_2$   & x                               & 1.00E-03  & x        & 1.00E-16 & x                              \\
        \hline
        \end{tabular}
\end{center}
\caption{The training configurations of YOLOv5s and YOLOv5m on COCO using Adam, Adafactor, SM3, CAME, and \optim. $\lambda$ means the growth-rate, $\gamma$ means the decay-rate, $d$ means the clip-threshold. We use recommended configurations in YOLOv5 official code and an A6000 GPU.}
\label{tab:configs_yolov5}
\end{table*}

%% file: tables/configs_transformer_wmt32k.tex
\begin{table*}[p!]
\begin{center}
\begin{tabular}{l||l|l|l|l|l|l}
        \toprule
        \multicolumn{7}{c}{Full-Training Configurations of Transformer-base and big on WMT32k} \\
        \multicolumn{7}{c}{Using Adam, Adafactor, SM3, CAME, and \optim}\\
        \midrule
        Model                               & Configurations  & Adam                            & Adafactor & SM3      & CAME     & \optim                           \\
        \bottomrule
                                            & epochs          & 400                             & 400       & 400      & 400      & 400                             \\
                                            & batch size      & 4096                            & 4096      & 4096     & 4096     & 4096                            \\
                                            & warmup-steps    & 16000                           & x         & 16000    & 16000    & 16000                           \\
                                            & weight-decay    & 0                               & 0         & 0        & 0        & 0                               \\
                                            & $\beta_1$       & 0.9                             & 0.9       & 0.9      & 0.9      & 0.9                             \\
                                            & $\beta_2$       & 0.997                           & x         & 0.997    & 0.997    & x                               \\
                                            & $\beta_3$       & x                               & x         & x        & 0.9999   & x                               \\
                                            & $\gamma$        & x                               & -0.8      & x        & x        & -0.5                            \\
                                            & $d$             & x                               & 1         & x        & 1        & x                               \\
                                            & $\lambda$       & x                               & x         & x        & x        & 0.999                           \\
                                            & $\epsilon_1$    & {\color[HTML]{1A1C1F} 1.00E-08} & 1.00E-30  & 1.00E-30 & 1.00E-30 & {\color[HTML]{1A1C1F} 1.00E-08} \\
                                            & $\epsilon_2$    & x                               & 1.00E-03  & x        & 1.00E-16 & x                               \\
        \multirow{-13}{*}{Transformer-base} & label-smoothing & 0.1                             & 0.1       & 0.1      & 0.1      & 0.1                             \\
        \hline
                                            & epochs          & 400                             & 400       & 400      & 400      & 400                             \\
                                            & batch size      & 4096                            & 4096      & 4096     & 4096     & 4096                            \\
                                            & warmup-steps    & 16000                           & x         & 16000    & 16000    & 16000                           \\
                                            & weight-decay    & 0                               & 0         & 0        & 0        & 0                               \\
                                            & $\beta_1$       & 0.9                             & 0.9       & 0.9      & 0.9      & 0.9                             \\
                                            & $\beta_2$       & 0.997                           & x         & 0.997    & 0.997    & x                               \\
                                            & $\beta_3$       & x                               & x         & x        & 0.9999   & x                               \\
                                            & $\gamma$        & x                               & -0.8      & x        & x        & -0.5                            \\
                                            & $d$             & x                               & 1         & x        & 1        & x                               \\
                                            & $\lambda$       & x                               & x         & x        & x        & 0.999                           \\
                                            & $\epsilon_1$    & {\color[HTML]{1A1C1F} 1.00E-08} & 1.00E-30  & 1.00E-30 & 1.00E-30 & {\color[HTML]{1A1C1F} 1.00E-08} \\
                                            & $\epsilon_2$    & x                               & 1.00E-03  & x        & 1.00E-16 & x                               \\
        \multirow{-13}{*}{Transformer-big}  & label-smoothing & 0.1                             & 0.1       & 0.1      & 0.1      & 0.1                            \\
        \hline
\end{tabular}
\end{center}
\caption{The full-training configurations of Transformer-base and big on WMT32k using Adam, Adafactor, SM3, CAME, and \optim. $\lambda$ means the growth-rate, $\gamma$ means the decay-rate, $d$ means the clip-threshold. We use an A6000 GPU.}
\label{tab:configs_transformer_wmt32k}
\end{table*}

%% file: tables/configs_transformer_bookcorpus.tex
\begin{table*}[p!]
\begin{center}
    \begin{tabular}{l||l|l|l|l|l|l}
    \toprule
    \multicolumn{7}{c}{Pre-Training Configurations of BERT, GPT-2, and T5 on BookCorpus \& Wikipedia-EN} \\
    \multicolumn{7}{c}{Using Adam, Adafactor, SM3, CAME and \optim}\\
    \midrule
    \multicolumn{1}{l||}{Model} & Configurations    & Adam                            & Adafactor & SM3      & CAME     & \optim                           \\
    \bottomrule
                              & scheduler         & \multicolumn{5}{c}{linear}                                                                          \\
                              & iterations        & 1,000,000                         & 1,000,000   & 1,000,000  & 1,000,000  & 1,000,000                         \\
                              & micro-batch size  & 4                               & 4         & 4        & 4        & 4                               \\
                              & global-batch size & 32                              & 32        & 32       & 32       & 32                              \\
                              & warmup-steps      & 10,000                           & 10,000     & 10,000    & 10,000    & 10,000                           \\
                              & learning-rate     & 0.0001                          & 0.0001    & 0.0001   & 0.0001   & 0.0001                          \\
                              & weight-decay      & 0.01                            & 0.01      & 0.01     & 0.01     & 0.01                            \\
                              & $\beta_1$         & 0.9                             & 0.9       & 0.9      & 0.9      & 0.9                             \\
                              & $\beta_2$         & 0.999                           & x         & 0.999    & 0.999    & x                               \\
                              & $\beta_3$         & x                               & x         & x        & 0.9999   & x                               \\
                              & $\gamma$          & x                               & -0.8      & x        & x        & -0.5                            \\
                              & $d$               & x                               & 1         & x        & 1        & x                               \\
                              & $\lambda$         & x                               & x         & x        & x        & 0.999                           \\
                              & $\epsilon_1$      & {\color[HTML]{1A1C1F} 1.00E-08} & 1.00E-30  & 1.00E-30 & 1.00E-30 & {\color[HTML]{1A1C1F} 1.00E-08} \\
    \multirow{-15}{*}{BERT}   & $\epsilon_2$      & x                               & 1.00E-03  & x        & 1.00E-16 & x                               \\
    \hline
                              & scheduler         & \multicolumn{5}{c}{cosine}                                                                          \\
                              & iterations        & 500,000                          & 500,000    & 500,000   & 500,000   & 500,000                          \\
                              & micro-batch size  & 4                               & 4         & 4        & 4        & 4                               \\
                              & global-batch size & 64                              & 64        & 64       & 64       & 64                              \\
                              & warmup-steps      & 5000                            & 5000      & 5000     & 5000     & 5000                            \\
                              & learning-rate     & 0.00015                         & 0.00015   & 0.00015  & 0.00015  & 0.00015                         \\
                              & weight-decay      & 0.01                            & 0.01      & 0.01     & 0.01     & 0.01                            \\
                              & $\beta_1$         & 0.9                             & 0.9       & 0.9      & 0.9      & 0.9                             \\
                              & $\beta_2$         & 0.999                           & x         & 0.999    & 0.999    & x                               \\
                              & $\beta_3$         & x                               & x         & x        & 0.9999   & x                               \\
                              & $\gamma$          & x                               & -0.8      & x        & x        & -0.5                            \\
                              & $d$               & x                               & 1         & x        & 1        & x                               \\
                              & $\lambda$         & x                               & x         & x        & x        & 0.999                           \\
                              & $\epsilon_1$      & {\color[HTML]{1A1C1F} 1.00E-08} & 1.00E-30  & 1.00E-30 & 1.00E-30 & {\color[HTML]{1A1C1F} 1.00E-08} \\
    \multirow{-15}{*}{GPT-2}   & $\epsilon_2$      & x                               & 1.00E-03  & x        & 1.00E-16 & x                               \\
    \hline
                              & scheduler         & \multicolumn{5}{c}{linear}                                                                          \\
                              & iterations        & 1,000,000                         & 1,000,000   & 1,000,000  & 1,000,000  & 1,000,000                         \\
                              & micro-batch size  & 16                              & 16        & 16       & 16       & 16                              \\
                              & global-batch size & 128                             & 128       & 128      & 128      & 128                             \\
                              & warmup-steps      & 10,000                           & 10,000     & 10,000    & 10,000    & 10,000                           \\
                              & learning-rate     & 0.0001                          & 0.0001    & 0.0001   & 0.0001   & 0.0001                          \\
                              & weight-decay      & 0.01                            & 0.01      & 0.01     & 0.01     & 0.01                            \\
                              & $\beta_1$         & 0.9                             & 0.9       & 0.9      & 0.9      & 0.9                             \\
                              & $\beta_2$         & 0.999                           & x         & 0.999    & 0.999    & x                               \\
                              & $\beta_3$         & x                               & x         & x        & 0.9999   & x                               \\
                              & $\gamma$          & x                               & -0.8      & x        & x        & -0.5                            \\
                              & $d$               & x                               & 1         & x        & 1        & x                               \\
                              & $\lambda$         & x                               & x         & x        & x        & 0.999                           \\
                              & $\epsilon_1$      & {\color[HTML]{1A1C1F} 1.00E-08} & 1.00E-30  & 1.00E-30 & 1.00E-30 & {\color[HTML]{1A1C1F} 1.00E-08} \\
    \multirow{-15}{*}{T5}     & $\epsilon_2$      & x                               & 1.00E-03  & x        & 1.00E-16 & x                              \\
    \hline
    \end{tabular}
\end{center}
\caption{The pre-training configurations of BERT, GPT2 and T5 on BookCorpus \& Wikipedia-EN using Adam, Adafactor, SM3, CAME, and \optim. $\lambda$ means the growth-rate, $\gamma$ means the decay-rate, $d$ means the clip-threshold. We use four A6000 GPUs to train BERT, four 3090 GPUs to train T5, and four A100 to train GPT-2. We also use NVIDIA Megatron-LM code. In this task, we use low-precision weight parameter used in the Megatron-LM code.}
\label{tab:configs_transformer_bookcorpus}
\end{table*}

%% file: tables/configs_transformer_gleu.tex
\begin{table*}[p!]

\begin{center}
\begin{tabular}{l||l|l|l|l|l|l}
\toprule
\multicolumn{7}{c}{Fine-tuning Configurations of BERT, GPT-2, and T5-small on QNLI}\\
\multicolumn{7}{c}{Using Adam, Adafactor, SM3, CAME, and SCMF} \\

\midrule
\multicolumn{1}{l}{Model}    & Configurations & Adam                            & Adafactor & SM3      & CAME     & \optim                           \\

\bottomrule
                             & scheduler      & \multicolumn{5}{c}{linear}                                                                          \\
                             & epochs         & 50                              & 50        & 50       & 50       & 50                              \\
                             & batch size     & 128                             & 128       & 128      & 128      & 128                             \\
                             & warmup-steps   & 10                              & 10        & 10       & 10       & 10                              \\
                             & learning-rate  & 2.00E-05                        & 2.00E-05  & 2.00E-05 & 2.00E-05 & 2.00E-05                        \\
                             & weight-decay   & 0.0005                          & 0.0005    & 0.0005   & 0.0005   & 0.0005                          \\
                             & $\beta_1$      & 0.9                             & 0.9       & 0.9      & 0.9      & 0.9                             \\
                             & $\beta_2$      & 0.999                           & x         & 0.999    & 0.999    & x                               \\
                             & $\beta_3$      & x                               & x         & x        & 0.9999   & x                               \\
                             & $\gamma$       & x                               & -0.8      & x        & x        & -0.8                            \\
                             & $d$            & x                               & 1         & x        & 1        & x                               \\
                             & $\lambda$      & x                               & x         & x        & x        & 0.999                           \\
                             & $\epsilon_1$   & {\color[HTML]{1A1C1F} 1.00E-08} & 1.00E-30  & 1.00E-30 & 1.00E-30 & {\color[HTML]{1A1C1F} 1.00E-08} \\
\multirow{-14}{*}{BERT}      & $\epsilon_2$   & x                               & 1.00E-03  & x        & 1.00E-16 & x                               \\
\hline
                             & scheduler      & \multicolumn{5}{c}{linear}                                                                          \\
                             & epochs         & 50                              & 50        & 50       & 50       & 50                              \\
                             & batch size     & 128                             & 128       & 128      & 128      & 128                             \\
                             & warmup-steps   & 100                             & 100       & 100      & 100      & 100                             \\
                             & learning-rate  & 2.00E-05                        & 2.00E-05  & 2.00E-05 & 2.00E-05 & 2.00E-05                        \\
                             & weight-decay   & 0.0005                          & 0.0005    & 0.0005   & 0.0005   & 0.0005                          \\
                             & $\beta_1$      & 0.9                             & 0.9       & 0.9      & 0.9      & 0.9                             \\
                             & $\beta_2$      & 0.999                           & x         & 0.999    & 0.999    & x                               \\
                             & $\beta_3$      & x                               & x         & x        & 0.9999   & x                               \\
                             & $\gamma$       & x                               & -0.8      & x        & x        & -0.8                            \\
                             & $d$            & x                               & 1         & x        & 1        & x                               \\
                             & $\lambda$      & x                               & x         & x        & x        & 0.999                           \\
                             & $\epsilon_1$   & {\color[HTML]{1A1C1F} 1.00E-08} & 1.00E-30  & 1.00E-30 & 1.00E-30 & {\color[HTML]{1A1C1F} 1.00E-08} \\
\multirow{-14}{*}{GPT2}      & $\epsilon_2$   & x                               & 1.00E-03  & x        & 1.00E-16 & x                               \\
\hline
                             & scheduler      & \multicolumn{5}{c}{linear}                                                                          \\
                             & epochs         & 50                              & 50        & 50       & 50       & 50                              \\
                             & batch size     & 128                             & 128       & 128      & 128      & 128                             \\
                             & warmup-steps   & 100                             & 100       & 100      & 100      & 100                             \\
                             & learning-rate  & 2.00E-05                        & 2.00E-05  & 2.00E-05 & 2.00E-05 & 2.00E-05                        \\
                             & weight-decay   & 0.0005                          & 0.0005    & 0.0005   & 0.0005   & 0.0005                          \\
                             & $\beta_1$      & 0.9                             & 0.9       & 0.9      & 0.9      & 0.9                             \\
                             & $\beta_2$      & 0.999                           & x         & 0.999    & 0.999    & x                               \\
                             & $\beta_3$      & x                               & x         & x        & 0.9999   & x                               \\
                             & $\gamma$       & x                               & -0.8      & x        & x        & -0.8                            \\
                             & $d$            & x                               & 1         & x        & 1        & x                               \\
                             & $\lambda$      & x                               & x         & x        & x        & 0.999                           \\
                             & $\epsilon_1$   & {\color[HTML]{1A1C1F} 1.00E-08} & 1.00E-30  & 1.00E-30 & 1.00E-30 & {\color[HTML]{1A1C1F} 1.00E-08} \\
\multirow{-14}{*}{T5 (small)} & $\epsilon_2$   & x                               & 1.00E-03  & x        & 1.00E-16 & x                              \\
\hline
\end{tabular}
\end{center}
\caption{The fine-tuning configurations of BERT, GPT-2, and T5-small on QNLI using Adam, Adafactor, SM3, CAME, and \optim. $\lambda$ means the growth-rate, $\gamma$ means the decay-rate, $d$ means the clip-threshold. We use a 3090 GPU. We use Hugging Face's BERT (\url{https://huggingface.co/bert-base-uncased}), GPT-2 (\url{https://huggingface.co/gpt2}), and T5-small (\url{https://huggingface.co/t5-small}).}
\label{tab:configs_transformer_gleu_qnli}

\end{table*}

\begin{table*}[p!]
\begin{center}
\begin{tabular}{l||l|l|l|l|l|l}
\toprule
\multicolumn{7}{c}{Fine-tuning Configurations of BERT, GPT-2, and T5-small on MNLI}\\
\multicolumn{7}{c}{Using Adam, Adafactor, SM3, CAME, and SCMF} \\

\midrule
\multicolumn{1}{l}{Model}    & Configurations & Adam                            & Adafactor & SM3      & CAME     & \optim                           \\

\bottomrule
                             & scheduler      & \multicolumn{5}{c}{linear}                                                                          \\
                             & epochs         & 50                              & 50        & 50       & 50       & 50                              \\
                             & batch size     & 128                             & 128       & 128      & 128      & 128                             \\
                             & warmup-steps   & 100                              & 100        & 100       & 100       & 100                              \\
                             & learning-rate  & 2.00E-05                        & 2.00E-05  & 2.00E-05 & 2.00E-05 & 2.00E-05                        \\
                             & weight-decay   & 0.0005                          & 0.0005    & 0.0005   & 0.0005   & 0.0005                          \\
                             & $\beta_1$      & 0.9                             & 0.9       & 0.9      & 0.9      & 0.9                             \\
                             & $\beta_2$      & 0.999                           & x         & 0.999    & 0.999    & x                               \\
                             & $\beta_3$      & x                               & x         & x        & 0.9999   & x                               \\
                             & $\gamma$       & x                               & -0.8      & x        & x        & -0.8                            \\
                             & $d$            & x                               & 1         & x        & 1        & x                               \\
                             & $\lambda$      & x                               & x         & x        & x        & 0.999                           \\
                             & $\epsilon_1$   & {\color[HTML]{1A1C1F} 1.00E-08} & 1.00E-30  & 1.00E-30 & 1.00E-30 & {\color[HTML]{1A1C1F} 1.00E-08} \\
\multirow{-14}{*}{BERT}      & $\epsilon_2$   & x                               & 1.00E-03  & x        & 1.00E-16 & x                               \\
\hline
                             & scheduler      & \multicolumn{5}{c}{linear}                                                                          \\
                             & epochs         & 50                              & 50        & 50       & 50       & 50                              \\
                             & batch size     & 128                             & 128       & 128      & 128      & 128                             \\
                             & warmup-steps   & 100                             & 100       & 100      & 100      & 100                             \\
                             & learning-rate  & 2.00E-05                        & 2.00E-05  & 2.00E-05 & 2.00E-05 & 2.00E-05                        \\
                             & weight-decay   & 0.0005                          & 0.0005    & 0.0005   & 0.0005   & 0.0005                          \\
                             & $\beta_1$      & 0.9                             & 0.9       & 0.9      & 0.9      & 0.9                             \\
                             & $\beta_2$      & 0.999                           & x         & 0.999    & 0.999    & x                               \\
                             & $\beta_3$      & x                               & x         & x        & 0.9999   & x                               \\
                             & $\gamma$       & x                               & -0.8      & x        & x        & -0.8                            \\
                             & $d$            & x                               & 1         & x        & 1        & x                               \\
                             & $\lambda$      & x                               & x         & x        & x        & 0.999                           \\
                             & $\epsilon_1$   & {\color[HTML]{1A1C1F} 1.00E-08} & 1.00E-30  & 1.00E-30 & 1.00E-30 & {\color[HTML]{1A1C1F} 1.00E-08} \\
\multirow{-14}{*}{GPT2}      & $\epsilon_2$   & x                               & 1.00E-03  & x        & 1.00E-16 & x                               \\
\hline
                             & scheduler      & \multicolumn{5}{c}{linear}                                                                          \\
                             & epochs         & 50                              & 50        & 50       & 50       & 50                              \\
                             & batch size     & 128                             & 128       & 128      & 128      & 128                             \\
                             & warmup-steps   & 100                             & 100       & 100      & 100      & 100                             \\
                             & learning-rate  & 2.00E-05                        & 2.00E-05  & 2.00E-05 & 2.00E-05 & 2.00E-05                        \\
                             & weight-decay   & 0.0005                          & 0.0005    & 0.0005   & 0.0005   & 0.0005                          \\
                             & $\beta_1$      & 0.9                             & 0.9       & 0.9      & 0.9      & 0.9                             \\
                             & $\beta_2$      & 0.999                           & x         & 0.999    & 0.999    & x                               \\
                             & $\beta_3$      & x                               & x         & x        & 0.9999   & x                               \\
                             & $\gamma$       & x                               & -0.8      & x        & x        & -0.8                            \\
                             & $d$            & x                               & 1         & x        & 1        & x                               \\
                             & $\lambda$      & x                               & x         & x        & x        & 0.999                           \\
                             & $\epsilon_1$   & {\color[HTML]{1A1C1F} 1.00E-08} & 1.00E-30  & 1.00E-30 & 1.00E-30 & {\color[HTML]{1A1C1F} 1.00E-08} \\
\multirow{-14}{*}{T5 (small)} & $\epsilon_2$   & x                               & 1.00E-03  & x        & 1.00E-16 & x                              \\
\hline
\end{tabular}
\end{center}
\caption{The fine-tuning configurations of BERT, GPT-2, and T5-small on MNLI using Adam, Adafactor, SM3, CAME, and \optim. $\lambda$ means the growth-rate, $\gamma$ means the decay-rate, $d$ means the clip-threshold. We use a 3090 GPU. We use Hugging Face's BERT (\url{https://huggingface.co/bert-base-uncased}), GPT-2 (\url{https://huggingface.co/gpt2}), and T5-small (\url{https://huggingface.co/t5-small}).}
\label{tab:configs_transformer_gleu_mnli}

\end{table*}

\begin{table*}[p!]

\begin{center}
\begin{tabular}{l||l|l|l|l|l|l}
\toprule
\multicolumn{7}{c}{Fine-tuning Configurations of BERT, GPT-2, and T5-small on QQP}\\
\multicolumn{7}{c}{Using Adam, Adafactor, SM3, CAME, and SCMF} \\

\midrule
\multicolumn{1}{l}{Model}    & Configurations & Adam                            & Adafactor & SM3      & CAME     & \optim                           \\

\bottomrule
                             & scheduler      & \multicolumn{5}{c}{linear}                                                                          \\
                             & epochs         & 50                              & 50        & 50       & 50       & 50                              \\
                             & batch size     & 128                             & 128       & 128      & 128      & 128                             \\
                             & warmup-steps   & 100                              & 100        & 100       & 100       & 100                              \\
                             & learning-rate  & 2.00E-05                        & 2.00E-05  & 2.00E-05 & 2.00E-05 & 2.00E-05                        \\
                             & weight-decay   & 0.0005                          & 0.0005    & 0.0005   & 0.0005   & 0.0005                          \\
                             & $\beta_1$      & 0.9                             & 0.9       & 0.9      & 0.9      & 0.9                             \\
                             & $\beta_2$      & 0.999                           & x         & 0.999    & 0.999    & x                               \\
                             & $\beta_3$      & x                               & x         & x        & 0.9999   & x                               \\
                             & $\gamma$       & x                               & -0.8      & x        & x        & -0.8                            \\
                             & $d$            & x                               & 1         & x        & 1        & x                               \\
                             & $\lambda$      & x                               & x         & x        & x        & 0.999                           \\
                             & $\epsilon_1$   & {\color[HTML]{1A1C1F} 1.00E-08} & 1.00E-30  & 1.00E-30 & 1.00E-30 & {\color[HTML]{1A1C1F} 1.00E-08} \\
\multirow{-14}{*}{BERT}      & $\epsilon_2$   & x                               & 1.00E-03  & x        & 1.00E-16 & x                               \\
\hline
                             & scheduler      & \multicolumn{5}{c}{linear}                                                                          \\
                             & epochs         & 50                              & 50        & 50       & 50       & 50                              \\
                             & batch size     & 128                             & 128       & 128      & 128      & 128                             \\
                             & warmup-steps   & 100                             & 100       & 100      & 100      & 100                             \\
                             & learning-rate  & 2.00E-05                        & 2.00E-05  & 2.00E-05 & 2.00E-05 & 2.00E-05                        \\
                             & weight-decay   & 0.0005                          & 0.0005    & 0.0005   & 0.0005   & 0.0005                          \\
                             & $\beta_1$      & 0.9                             & 0.9       & 0.9      & 0.9      & 0.9                             \\
                             & $\beta_2$      & 0.999                           & x         & 0.999    & 0.999    & x                               \\
                             & $\beta_3$      & x                               & x         & x        & 0.9999   & x                               \\
                             & $\gamma$       & x                               & -0.8      & x        & x        & -0.8                            \\
                             & $d$            & x                               & 1         & x        & 1        & x                               \\
                             & $\lambda$      & x                               & x         & x        & x        & 0.999                           \\
                             & $\epsilon_1$   & {\color[HTML]{1A1C1F} 1.00E-08} & 1.00E-30  & 1.00E-30 & 1.00E-30 & {\color[HTML]{1A1C1F} 1.00E-08} \\
\multirow{-14}{*}{GPT2}      & $\epsilon_2$   & x                               & 1.00E-03  & x        & 1.00E-16 & x                               \\
\hline
                             & scheduler      & \multicolumn{5}{c}{linear}                                                                          \\
                             & epochs         & 50                              & 50        & 50       & 50       & 50                              \\
                             & batch size     & 128                             & 128       & 128      & 128      & 128                             \\
                             & warmup-steps   & 100                             & 100       & 100      & 100      & 100                             \\
                             & learning-rate  & 2.00E-05                        & 2.00E-05  & 2.00E-05 & 2.00E-05 & 2.00E-05                        \\
                             & weight-decay   & 0.0005                          & 0.0005    & 0.0005   & 0.0005   & 0.0005                          \\
                             & $\beta_1$      & 0.9                             & 0.9       & 0.9      & 0.9      & 0.9                             \\
                             & $\beta_2$      & 0.999                           & x         & 0.999    & 0.999    & x                               \\
                             & $\beta_3$      & x                               & x         & x        & 0.9999   & x                               \\
                             & $\gamma$       & x                               & -0.8      & x        & x        & -0.8                            \\
                             & $d$            & x                               & 1         & x        & 1        & x                               \\
                             & $\lambda$      & x                               & x         & x        & x        & 0.999                           \\
                             & $\epsilon_1$   & {\color[HTML]{1A1C1F} 1.00E-08} & 1.00E-30  & 1.00E-30 & 1.00E-30 & {\color[HTML]{1A1C1F} 1.00E-08} \\
\multirow{-14}{*}{T5 (small)} & $\epsilon_2$   & x                               & 1.00E-03  & x        & 1.00E-16 & x                              \\
\hline
\end{tabular}
\end{center}
\caption{The fine-tuning configurations of BERT, GPT-2, and T5-small on QQP using Adam, Adafactor, SM3, CAME, and \optim. $\lambda$ means the growth-rate, $\gamma$ means the decay-rate, $d$ means the clip-threshold. We use a 3090 GPU. We use Hugging Face's BERT (\url{https://huggingface.co/bert-base-uncased}), GPT-2 (\url{https://huggingface.co/gpt2}), and T5-small (\url{https://huggingface.co/t5-small}).}
\label{tab:configs_transformer_gleu_qqp}

\end{table*}

\begin{table*}[p!]

\begin{center}
\begin{tabular}{l||l|l|l|l|l|l}
\toprule
\multicolumn{7}{c}{Fine-tuning Configurations of BERT, GPT-2, and T5-small on STSB}\\
\multicolumn{7}{c}{Using Adam, Adafactor, SM3, CAME, and SCMF} \\

\midrule
\multicolumn{1}{l}{Model}    & Configurations & Adam                            & Adafactor & SM3      & CAME     & \optim                           \\

\bottomrule
                             & scheduler      & \multicolumn{5}{c}{linear}                                                                          \\
                             & epochs         & 50                              & 50        & 50       & 50       & 50                              \\
                             & batch size     & 128                             & 128       & 128      & 128      & 128                             \\
                             & warmup-steps   & 100                              & 100        & 100       & 100       & 100                              \\
                             & learning-rate  & 2.00E-05                        & 2.00E-05  & 2.00E-05 & 2.00E-05 & 2.00E-05                        \\
                             & weight-decay   & 0.0005                          & 0.0005    & 0.0005   & 0.0005   & 0.0005                          \\
                             & $\beta_1$      & 0.9                             & 0.9       & 0.9      & 0.9      & 0.9                             \\
                             & $\beta_2$      & 0.999                           & x         & 0.999    & 0.999    & x                               \\
                             & $\beta_3$      & x                               & x         & x        & 0.9999   & x                               \\
                             & $\gamma$       & x                               & -0.8      & x        & x        & -0.8                            \\
                             & $d$            & x                               & 1         & x        & 1        & x                               \\
                             & $\lambda$      & x                               & x         & x        & x        & 0.999                           \\
                             & $\epsilon_1$   & {\color[HTML]{1A1C1F} 1.00E-08} & 1.00E-30  & 1.00E-30 & 1.00E-30 & {\color[HTML]{1A1C1F} 1.00E-08} \\
\multirow{-14}{*}{BERT}      & $\epsilon_2$   & x                               & 1.00E-03  & x        & 1.00E-16 & x                               \\
\hline
                             & scheduler      & \multicolumn{5}{c}{linear}                                                                          \\
                             & epochs         & 50                              & 50        & 50       & 50       & 50                              \\
                             & batch size     & 128                             & 128       & 128      & 128      & 128                             \\
                             & warmup-steps   & 100                             & 100       & 100      & 100      & 100                             \\
                             & learning-rate  & 2.00E-05                        & 2.00E-05  & 2.00E-05 & 2.00E-05 & 2.00E-05                        \\
                             & weight-decay   & 0.0005                          & 0.0005    & 0.0005   & 0.0005   & 0.0005                          \\
                             & $\beta_1$      & 0.9                             & 0.9       & 0.9      & 0.9      & 0.9                             \\
                             & $\beta_2$      & 0.999                           & x         & 0.999    & 0.999    & x                               \\
                             & $\beta_3$      & x                               & x         & x        & 0.9999   & x                               \\
                             & $\gamma$       & x                               & -0.8      & x        & x        & -0.8                            \\
                             & $d$            & x                               & 1         & x        & 1        & x                               \\
                             & $\lambda$      & x                               & x         & x        & x        & 0.999                           \\
                             & $\epsilon_1$   & {\color[HTML]{1A1C1F} 1.00E-08} & 1.00E-30  & 1.00E-30 & 1.00E-30 & {\color[HTML]{1A1C1F} 1.00E-08} \\
\multirow{-14}{*}{GPT2}      & $\epsilon_2$   & x                               & 1.00E-03  & x        & 1.00E-16 & x                               \\
\hline
                             & scheduler      & \multicolumn{5}{c}{linear}                                                                          \\
                             & epochs         & 50                              & 50        & 50       & 50       & 50                              \\
                             & batch size     & 128                             & 128       & 128      & 128      & 128                             \\
                             & warmup-steps   & 100                             & 100       & 100      & 100      & 100                             \\
                             & learning-rate  & 2.00E-05                        & 2.00E-05  & 2.00E-05 & 2.00E-05 & 2.00E-05                        \\
                             & weight-decay   & 0.0005                          & 0.0005    & 0.0005   & 0.0005   & 0.0005                          \\
                             & $\beta_1$      & 0.9                             & 0.9       & 0.9      & 0.9      & 0.9                             \\
                             & $\beta_2$      & 0.999                           & x         & 0.999    & 0.999    & x                               \\
                             & $\beta_3$      & x                               & x         & x        & 0.9999   & x                               \\
                             & $\gamma$       & x                               & -0.8      & x        & x        & -0.8                            \\
                             & $d$            & x                               & 1         & x        & 1        & x                               \\
                             & $\lambda$      & x                               & x         & x        & x        & 0.999                           \\
                             & $\epsilon_1$   & {\color[HTML]{1A1C1F} 1.00E-08} & 1.00E-30  & 1.00E-30 & 1.00E-30 & {\color[HTML]{1A1C1F} 1.00E-08} \\
\multirow{-14}{*}{T5 (small)} & $\epsilon_2$   & x                               & 1.00E-03  & x        & 1.00E-16 & x                              \\
\hline
\end{tabular}
\end{center}
\caption{The fine-tuning configurations of BERT, GPT-2, and T5-small on STSB using Adam, Adafactor, SM3, CAME, and \optim. $\lambda$ means the growth-rate, $\gamma$ means the decay-rate, $d$ means the clip-threshold. We use a 3090 GPU. We use Hugging Face's BERT (\url{https://huggingface.co/bert-base-uncased}), GPT-2 (\url{https://huggingface.co/gpt2}), and T5-small (\url{https://huggingface.co/t5-small}).}
\label{tab:configs_transformer_gleu_stsb}

\end{table*}

\begin{table*}[p!]

\begin{center}
\begin{tabular}{l||l|l|l|l|l|l}
\toprule
\multicolumn{7}{c}{Fine-tuning Configurations of BERT, GPT-2, and T5-small on MRPC}\\
\multicolumn{7}{c}{Using Adam, Adafactor, SM3, CAME, and SCMF} \\

\midrule
\multicolumn{1}{l}{Model}    & Configurations & Adam                            & Adafactor & SM3      & CAME     & \optim                           \\

\bottomrule
                             & scheduler      & \multicolumn{5}{c}{linear}                                                                          \\
                             & epochs         & 50                              & 50        & 50       & 50       & 50                              \\
                             & batch size     & 128                             & 128       & 128      & 128      & 128                             \\
                             & warmup-steps   & 100                              & 100        & 100       & 100       & 100                              \\
                             & learning-rate  & 2.00E-05                        & 2.00E-05  & 2.00E-05 & 2.00E-05 & 2.00E-05                        \\
                             & weight-decay   & 0.0005                          & 0.0005    & 0.0005   & 0.0005   & 0.0005                          \\
                             & $\beta_1$      & 0.9                             & 0.9       & 0.9      & 0.9      & 0.9                             \\
                             & $\beta_2$      & 0.999                           & x         & 0.999    & 0.999    & x                               \\
                             & $\beta_3$      & x                               & x         & x        & 0.9999   & x                               \\
                             & $\gamma$       & x                               & -0.8      & x        & x        & -0.8                            \\
                             & $d$            & x                               & 1         & x        & 1        & x                               \\
                             & $\lambda$      & x                               & x         & x        & x        & 0.999                           \\
                             & $\epsilon_1$   & {\color[HTML]{1A1C1F} 1.00E-08} & 1.00E-30  & 1.00E-30 & 1.00E-30 & {\color[HTML]{1A1C1F} 1.00E-08} \\
\multirow{-14}{*}{BERT}      & $\epsilon_2$   & x                               & 1.00E-03  & x        & 1.00E-16 & x                               \\
\hline
                             & scheduler      & \multicolumn{5}{c}{linear}                                                                          \\
                             & epochs         & 50                              & 50        & 50       & 50       & 50                              \\
                             & batch size     & 128                             & 128       & 128      & 128      & 128                             \\
                             & warmup-steps   & 100                             & 100       & 100      & 100      & 100                             \\
                             & learning-rate  & 2.00E-05                        & 2.00E-05  & 2.00E-05 & 2.00E-05 & 2.00E-05                        \\
                             & weight-decay   & 0.0005                          & 0.0005    & 0.0005   & 0.0005   & 0.0005                          \\
                             & $\beta_1$      & 0.9                             & 0.9       & 0.9      & 0.9      & 0.9                             \\
                             & $\beta_2$      & 0.999                           & x         & 0.999    & 0.999    & x                               \\
                             & $\beta_3$      & x                               & x         & x        & 0.9999   & x                               \\
                             & $\gamma$       & x                               & -0.8      & x        & x        & -0.8                            \\
                             & $d$            & x                               & 1         & x        & 1        & x                               \\
                             & $\lambda$      & x                               & x         & x        & x        & 0.999                           \\
                             & $\epsilon_1$   & {\color[HTML]{1A1C1F} 1.00E-08} & 1.00E-30  & 1.00E-30 & 1.00E-30 & {\color[HTML]{1A1C1F} 1.00E-08} \\
\multirow{-14}{*}{GPT2}      & $\epsilon_2$   & x                               & 1.00E-03  & x        & 1.00E-16 & x                               \\
\hline
                             & scheduler      & \multicolumn{5}{c}{linear}                                                                          \\
                             & epochs         & 50                              & 50        & 50       & 50       & 50                              \\
                             & batch size     & 128                             & 128       & 128      & 128      & 128                             \\
                             & warmup-steps   & 100                             & 100       & 100      & 100      & 100                             \\
                             & learning-rate  & 2.00E-05                        & 2.00E-05  & 2.00E-05 & 2.00E-05 & 2.00E-05                        \\
                             & weight-decay   & 0.0005                          & 0.0005    & 0.0005   & 0.0005   & 0.0005                          \\
                             & $\beta_1$      & 0.9                             & 0.9       & 0.9      & 0.9      & 0.9                             \\
                             & $\beta_2$      & 0.999                           & x         & 0.999    & 0.999    & x                               \\
                             & $\beta_3$      & x                               & x         & x        & 0.9999   & x                               \\
                             & $\gamma$       & x                               & -0.8      & x        & x        & -0.8                            \\
                             & $d$            & x                               & 1         & x        & 1        & x                               \\
                             & $\lambda$      & x                               & x         & x        & x        & 0.999                           \\
                             & $\epsilon_1$   & {\color[HTML]{1A1C1F} 1.00E-08} & 1.00E-30  & 1.00E-30 & 1.00E-30 & {\color[HTML]{1A1C1F} 1.00E-08} \\
\multirow{-14}{*}{T5 (small)} & $\epsilon_2$   & x                               & 1.00E-03  & x        & 1.00E-16 & x                              \\
\hline
\end{tabular}
\end{center}
\caption{The fine-tuning configurations of BERT, GPT-2, and T5-small on MRPC using Adam, Adafactor, SM3, CAME, and \optim. $\lambda$ means the growth-rate, $\gamma$ means the decay-rate, $d$ means the clip-threshold. We use a 3090 GPU. We use Hugging Face's BERT (\url{https://huggingface.co/bert-base-uncased}), GPT-2 (\url{https://huggingface.co/gpt2}), and T5-small (\url{https://huggingface.co/t5-small}).}
\label{tab:configs_transformer_gleu_mrpc}

\end{table*}

\begin{table*}[p!]

\begin{center}
\begin{tabular}{l||l|l|l|l|l|l}
\toprule
\multicolumn{7}{c}{Fine-tuning Configurations of BERT and GPT-2 on SQuAD}\\
\multicolumn{7}{c}{Using Adam, Adafactor, SM3, CAME, and SCMF} \\

\midrule
\multicolumn{1}{l}{Model}    & Configurations & Adam                            & Adafactor & SM3      & CAME     & \optim                           \\

\bottomrule
                             & scheduler      & \multicolumn{5}{c}{linear}                                                                          \\
                             & epochs         & 2                              & 2        & 2       & 2       & 2                              \\
                             & batch size     & 12                             & 12       & 12      & 12      & 12                             \\
                             & warmup-steps   & 0                              & 0        & 0       & 0       & 0                              \\
                             & learning-rate  & 1.00E-05                        & 1.00E-05  & 2.5E-03 & 1.00E-05 & 1.00E-05                        \\
                             & weight-decay   & 0.0                          & 0.0    & 0.0   & 0.0   & 0.0                          \\
                             & $\beta_1$      & 0.9                             & 0.9       & 0.9      & 0.9      & 0.9                             \\
                             & $\beta_2$      & 0.999                           & x         & 0.999    & 0.999    & x                               \\
                             & $\beta_3$      & x                               & x         & x        & 0.9999   & x                               \\
                             & $\gamma$       & x                               & -0.8      & x        & x        & -0.5                            \\
                             & $d$            & x                               & 1         & x        & 1        & x                               \\
                             & $\lambda$      & x                               & x         & x        & x        & 0.999                           \\
                             & $\epsilon_1$   & {\color[HTML]{1A1C1F} 1.00E-08} & 1.00E-30  & 1.00E-30 & 1.00E-30 & {\color[HTML]{1A1C1F} 1.00E-08} \\
\multirow{-14}{*}{BERT}      & $\epsilon_2$   & x                               & 1.00E-03  & x        & 1.00E-16 & x                               \\
\hline
                             & scheduler      & \multicolumn{5}{c}{linear}                                                                          \\
                             & epochs         & 2                              & 2        & 2       & 2       & 2                              \\
                             & batch size     & 12                             &12       & 12      & 12      & 12                             \\
                             & warmup-steps   & 0                             & 0       & 0      & 0      & 0                             \\
                             & learning-rate  & 3.00E-04                        & 3.00E-04  & 2.50E-03 & 1.00E-05 & 3.00E-04                        \\
                             & weight-decay   & 0.0                          & 0.0    & 0.0   & 0.0   & 0.0                          \\
                             & $\beta_1$      & 0.9                             & 0.9       & 0.9      & 0.9      & 0.9                             \\
                             & $\beta_2$      & 0.999                           & x         & 0.999    & 0.999    & x                               \\
                             & $\beta_3$      & x                               & x         & x        & 0.9999   & x                               \\
                             & $\gamma$       & x                               & -0.8      & x        & x        & -0.5                            \\
                             & $d$            & x                               & 1         & x        & 1        & x                               \\
                             & $\lambda$      & x                               & x         & x        & x        & 0.999                           \\
                             & $\epsilon_1$   & {\color[HTML]{1A1C1F} 1.00E-08} & 1.00E-30  & 1.00E-30 & 1.00E-30 & {\color[HTML]{1A1C1F} 1.00E-08} \\
\multirow{-14}{*}{GPT2}      & $\epsilon_2$   & x                               & 1.00E-03  & x        & 1.00E-16 & x                               \\
\hline
\end{tabular}
\end{center}
\caption{The fine-tuning configurations of BERT and GPT-2 on SQuAD using Adam, Adafactor, SM3, CAME, and \optim. $\lambda$ means the growth-rate, $\gamma$ means the decay-rate, $d$ means the clip-threshold. We use a A6000 GPU. We use Hugging Face's BERT (\url{https://huggingface.co/bert-base-uncased}) and GPT-2 (\url{https://huggingface.co/gpt2}).}
\label{tab:configs_transformer_squad1}

\end{table*}
\begin{table*}[tp!]
\begin{center}
\begin{tabular}{l||l|l|l|l|l|l}
\toprule
\multicolumn{7}{c}{Fine-tuning Configurations of T5-small on SQuADv2}\\
\multicolumn{7}{c}{Using Adam, Adafactor, SM3, CAME, and SCMF} \\

\midrule
\multicolumn{1}{l}{Model}    & Configurations & Adam                            & Adafactor & SM3      & CAME     & \optim                           \\

\bottomrule
                             & scheduler      & \multicolumn{5}{c}{linear}                                                                          \\
                             & epochs         & 2                              & 2        & 2       & 2       & 2                              \\
                             & batch size     & 12                             & 12       & 12      & 12      & 12                             \\
                             & warmup-steps   & 0                              & 0        & 0       & 0       & 0                              \\
                             & learning-rate  & 3.00E-04                        & 3.00E-04  & 2.50E-03 & 1.00E-05 & 3.00E-04                        \\
                             & weight-decay   & 0.0                          & 0.0    & 0.0   & 0.0   & 0.0                          \\
                             & $\beta_1$      & 0.9                             & 0.9       & 0.9      & 0.9      & 0.9                             \\
                             & $\beta_2$      & 0.999                           & x         & 0.999    & 0.999    & x                               \\
                             & $\beta_3$      & x                               & x         & x        & 0.9999   & x                               \\
                             & $\gamma$       & x                               & -0.8      & x        & x        & -0.5                            \\
                             & $d$            & x                               & 1         & x        & 1        & x                               \\
                             & $\lambda$      & x                               & x         & x        & x        & 0.999                           \\
                             & $\epsilon_1$   & {\color[HTML]{1A1C1F} 1.00E-08} & 1.00E-30  & 1.00E-30 & 1.00E-30 & {\color[HTML]{1A1C1F} 1.00E-08} \\
\multirow{-14}{*}{T5 (small)}      & $\epsilon_2$   & x                               & 1.00E-03  & x        & 1.00E-16 & x                               \\

\hline
\end{tabular}
\end{center}
\caption{The fine-tuning configurations of T5-small on SQuADv2 using Adam, Adafactor, SM3, CAME, and \optim. $\lambda$ means the growth-rate, $\gamma$ means the decay-rate, $d$ means the clip-threshold. We use a A6000 GPU. We use Hugging Face's T5-small (\url{https://huggingface.co/t5-small}).}
\label{tab:configs_transformer_squad2}

\end{table*}

%% file: aaai25.bbl
\begin{thebibliography}{54}
\providecommand{\natexlab}[1]{#1}

\bibitem[{Abdulkadirov, Lyakhov, and Nagornov(2023)}]{abdulkadirov2023survey}
Abdulkadirov, R.; Lyakhov, P.; and Nagornov, N. 2023.
\newblock Survey of Optimization Algorithms in Modern Neural Networks.
\newblock \emph{Mathematics}, 11(11): 2466.

\bibitem[{Aghajanyan, Zettlemoyer, and Gupta(2020)}]{low_dim02}
Aghajanyan, A.; Zettlemoyer, L.; and Gupta, S. 2020.
\newblock Intrinsic Dimensionality Explains the Effectiveness of Language Model Fine-Tuning.
\newblock arXiv:2012.13255.

\bibitem[{Amari(2010)}]{amari2010information}
Amari, S.-i. 2010.
\newblock Information geometry in optimization, machine learning and statistical inference.
\newblock \emph{Frontiers of Electrical and Electronic Engineering in China}, 5: 241--260.

\bibitem[{Anil et~al.(2019)Anil, Gupta, Koren, and Singer}]{sm3}
Anil, R.; Gupta, V.; Koren, T.; and Singer, Y. 2019.
\newblock Memory Efficient Adaptive Optimization.
\newblock In Wallach, H.; Larochelle, H.; Beygelzimer, A.; d\textquotesingle Alch\'{e}-Buc, F.; Fox, E.; and Garnett, R., eds., \emph{Advances in Neural Information Processing Systems}, volume~32. Curran Associates, Inc.

\bibitem[{Bojar et~al.(2014)Bojar, Buck, Federmann, Haddow, Koehn, Leveling, Monz, Pecina, Post, Saint-Amand et~al.}]{bojar2014findings_wmt32k}
Bojar, O.; Buck, C.; Federmann, C.; Haddow, B.; Koehn, P.; Leveling, J.; Monz, C.; Pecina, P.; Post, M.; Saint-Amand, H.; et~al. 2014.
\newblock Findings of the 2014 workshop on statistical machine translation.
\newblock In \emph{Proceedings of the ninth workshop on statistical machine translation}, 12--58.

\bibitem[{Bojar et~al.(2016)Bojar, Chatterjee, Federmann, Graham, Haddow, Huck, Jimeno~Yepes, Koehn, Logacheva, Monz, Negri, Neveol, Neves, Popel, Post, Rubino, Scarton, Specia, Turchi, Verspoor, and Zampieri}]{wmt16}
Bojar, O.~r.; Chatterjee, R.; Federmann, C.; Graham, Y.; Haddow, B.; Huck, M.; Jimeno~Yepes, A.; Koehn, P.; Logacheva, V.; Monz, C.; Negri, M.; Neveol, A.; Neves, M.; Popel, M.; Post, M.; Rubino, R.; Scarton, C.; Specia, L.; Turchi, M.; Verspoor, K.; and Zampieri, M. 2016.
\newblock Findings of the 2016 Conference on Machine Translation.
\newblock In \emph{Proceedings of the First Conference on Machine Translation}, 131--198. Berlin, Germany: Association for Computational Linguistics.

\bibitem[{Chen et~al.(2016)Chen, Xu, Zhang, and Guestrin}]{gradient_checkpointing}
Chen, T.; Xu, B.; Zhang, C.; and Guestrin, C. 2016.
\newblock Training deep nets with sublinear memory cost.
\newblock \emph{arXiv preprint arXiv:1604.06174}.

\bibitem[{Devlin et~al.(2018)Devlin, Chang, Lee, and Toutanova}]{bert}
Devlin, J.; Chang, M.-W.; Lee, K.; and Toutanova, K. 2018.
\newblock Bert: Pre-training of deep bidirectional transformers for language understanding.
\newblock \emph{arXiv preprint arXiv:1810.04805}.

\bibitem[{Dong et~al.(2020)Dong, Zhou, Ruan, and Li}]{mobilenet}
Dong, K.; Zhou, C.; Ruan, Y.; and Li, Y. 2020.
\newblock MobileNetV2 Model for Image Classification.
\newblock In \emph{2020 2nd International Conference on Information Technology and Computer Application (ITCA)}, 476--480.

\bibitem[{Finesso and Spreij(2006)}]{NNMF}
Finesso, L.; and Spreij, P. 2006.
\newblock Nonnegative matrix factorization and I-divergence alternating minimization.
\newblock \emph{Linear Algebra and its Applications}, 416(2): 270--287.

\bibitem[{Google(2022)}]{coral_devboard_micro}
Google. 2022.
\newblock Coral Dev Board Micro.
\newblock Https://coral.ai/products/dev-board-micro/.

\bibitem[{Harris et~al.(2020)Harris, Millman, van~der Walt, Gommers, Virtanen, Cournapeau, Wieser, Taylor, Berg, Smith, Kern, Picus, Hoyer, van Kerkwijk, Brett, Haldane, del R{\'{i}}o, Wiebe, Peterson, G{\'{e}}rard-Marchant, Sheppard, Reddy, Weckesser, Abbasi, Gohlke, and Oliphant}]{numpy}
Harris, C.~R.; Millman, K.~J.; van~der Walt, S.~J.; Gommers, R.; Virtanen, P.; Cournapeau, D.; Wieser, E.; Taylor, J.; Berg, S.; Smith, N.~J.; Kern, R.; Picus, M.; Hoyer, S.; van Kerkwijk, M.~H.; Brett, M.; Haldane, A.; del R{\'{i}}o, J.~F.; Wiebe, M.; Peterson, P.; G{\'{e}}rard-Marchant, P.; Sheppard, K.; Reddy, T.; Weckesser, W.; Abbasi, H.; Gohlke, C.; and Oliphant, T.~E. 2020.
\newblock Array programming with {NumPy}.
\newblock \emph{Nature}, 585(7825): 357--362.

\bibitem[{He et~al.(2016)He, Zhang, Ren, and Sun}]{resnet}
He, K.; Zhang, X.; Ren, S.; and Sun, J. 2016.
\newblock Deep Residual Learning for Image Recognition.
\newblock In \emph{2016 IEEE Conference on Computer Vision and Pattern Recognition (CVPR)}, 770--778.

\bibitem[{Hinton, Srivastava, and Swersky(2012)}]{rmsprop}
Hinton, G.; Srivastava, N.; and Swersky, K. 2012.
\newblock Neural networks for machine learning lecture 6a overview of mini-batch gradient descent.
\newblock \emph{Cited on}, 14(8): 2.

\bibitem[{Hu et~al.(2021)Hu, Shen, Wallis, Allen-Zhu, Li, Wang, Wang, and Chen}]{lora}
Hu, E.~J.; Shen, Y.; Wallis, P.; Allen-Zhu, Z.; Li, Y.; Wang, S.; Wang, L.; and Chen, W. 2021.
\newblock LoRA: Low-Rank Adaptation of Large Language Models.
\newblock arXiv:2106.09685.

\bibitem[{Kingma and Ba(2014)}]{adam}
Kingma, D.~P.; and Ba, J. 2014.
\newblock Adam: A method for stochastic optimization.
\newblock \emph{arXiv preprint arXiv:1412.6980}.

\bibitem[{Krizhevsky, Hinton et~al.(2009)}]{cifar}
Krizhevsky, A.; Hinton, G.; et~al. 2009.
\newblock Learning multiple layers of features from tiny images.

\bibitem[{Kumar et~al.(2019)Kumar, Panigrahy, Rahimi, and Woodruff}]{bmf2}
Kumar, R.; Panigrahy, R.; Rahimi, A.; and Woodruff, D. 2019.
\newblock Faster Algorithms for Binary Matrix Factorization.
\newblock In Chaudhuri, K.; and Salakhutdinov, R., eds., \emph{Proceedings of the 36th International Conference on Machine Learning}, volume~97 of \emph{Proceedings of Machine Learning Research}, 3551--3559. PMLR.

\bibitem[{Lan et~al.(2020)Lan, Chen, Goodman, Gimpel, Sharma, and Soricut}]{albert}
Lan, Z.; Chen, M.; Goodman, S.; Gimpel, K.; Sharma, P.; and Soricut, R. 2020.
\newblock ALBERT: A Lite BERT for Self-supervised Learning of Language Representations.
\newblock arXiv:1909.11942.

\bibitem[{Landro et~al.(2022{\natexlab{a}})Landro, Gallo, La~Grassa, and Federici}]{ilpost}
Landro, N.; Gallo, I.; La~Grassa, R.; and Federici, E. 2022{\natexlab{a}}.
\newblock Two New Datasets for Italian-Language Abstractive Text Summarization.
\newblock \emph{Information}, 13(5).

\bibitem[{Landro et~al.(2022{\natexlab{b}})Landro, Gallo, La~Grassa, and Federici}]{fanpage}
Landro, N.; Gallo, I.; La~Grassa, R.; and Federici, E. 2022{\natexlab{b}}.
\newblock Two New Datasets for Italian-Language Abstractive Text Summarization.
\newblock \emph{Information}, 13(5).

\bibitem[{Lee and Seung(1999)}]{Idivergence}
Lee, D.~D.; and Seung, H.~S. 1999.
\newblock Learning the parts of objects by non-negative matrix factorization.
\newblock \emph{Nature}, 401(6755): 788--791.

\bibitem[{Lewis et~al.(2019)Lewis, Liu, Goyal, Ghazvininejad, Mohamed, Levy, Stoyanov, and Zettlemoyer}]{bart}
Lewis, M.; Liu, Y.; Goyal, N.; Ghazvininejad, M.; Mohamed, A.; Levy, O.; Stoyanov, V.; and Zettlemoyer, L. 2019.
\newblock BART: Denoising Sequence-to-Sequence Pre-training for Natural Language Generation, Translation, and Comprehension.
\newblock arXiv:1910.13461.

\bibitem[{Li et~al.(2018)Li, Farkhoor, Liu, and Yosinski}]{low_dim01}
Li, C.; Farkhoor, H.; Liu, R.; and Yosinski, J. 2018.
\newblock Measuring the Intrinsic Dimension of Objective Landscapes.
\newblock arXiv:1804.08838.

\bibitem[{Lin et~al.(2015)Lin, Maire, Belongie, Bourdev, Girshick, Hays, Perona, Ramanan, Zitnick, and Dollár}]{lin2015microsoft}
Lin, T.-Y.; Maire, M.; Belongie, S.; Bourdev, L.; Girshick, R.; Hays, J.; Perona, P.; Ramanan, D.; Zitnick, C.~L.; and Dollár, P. 2015.
\newblock Microsoft COCO: Common Objects in Context.
\newblock arXiv:1405.0312.

\bibitem[{Liu and Nocedal(1989)}]{liu1989limited}
Liu, D.~C.; and Nocedal, J. 1989.
\newblock On the limited memory BFGS method for large scale optimization.
\newblock \emph{Mathematical programming}, 45(1-3): 503--528.

\bibitem[{Liu et~al.(2020)Liu, Gu, Goyal, Li, Edunov, Ghazvininejad, Lewis, and Zettlemoyer}]{mbart}
Liu, Y.; Gu, J.; Goyal, N.; Li, X.; Edunov, S.; Ghazvininejad, M.; Lewis, M.; and Zettlemoyer, L. 2020.
\newblock Multilingual Denoising Pre-training for Neural Machine Translation.
\newblock arXiv:2001.08210.

\bibitem[{Liu et~al.(2019)Liu, Ott, Goyal, Du, Joshi, Chen, Levy, Lewis, Zettlemoyer, and Stoyanov}]{roberta}
Liu, Y.; Ott, M.; Goyal, N.; Du, J.; Joshi, M.; Chen, D.; Levy, O.; Lewis, M.; Zettlemoyer, L.; and Stoyanov, V. 2019.
\newblock RoBERTa: A Robustly Optimized BERT Pretraining Approach.
\newblock arXiv:1907.11692.

\bibitem[{Loshchilov and Hutter(2019)}]{adamw}
Loshchilov, I.; and Hutter, F. 2019.
\newblock Decoupled Weight Decay Regularization.
\newblock arXiv:1711.05101.

\bibitem[{Luo et~al.(2023)Luo, Ren, Zheng, Jiang, Jiang, and You}]{came}
Luo, Y.; Ren, X.; Zheng, Z.; Jiang, Z.; Jiang, X.; and You, Y. 2023.
\newblock {CAME}: Confidence-guided Adaptive Memory Efficient Optimization.
\newblock In Rogers, A.; Boyd-Graber, J.; and Okazaki, N., eds., \emph{Proceedings of the 61st Annual Meeting of the Association for Computational Linguistics (Volume 1: Long Papers)}, 4442--4453. Toronto, Canada: Association for Computational Linguistics.

\bibitem[{Martens(2016)}]{martens2016second}
Martens, J. 2016.
\newblock \emph{Second-order optimization for neural networks}.
\newblock University of Toronto (Canada).

\bibitem[{Nallapati et~al.(2016)Nallapati, Zhou, dos santos, Gulcehre, and Xiang}]{cnn_dailymail}
Nallapati, R.; Zhou, B.; dos santos, C.~N.; Gulcehre, C.; and Xiang, B. 2016.
\newblock Abstractive Text Summarization Using Sequence-to-Sequence RNNs and Beyond.
\newblock arXiv:1602.06023.

\bibitem[{Narayan, Cohen, and Lapata(2018)}]{xsum}
Narayan, S.; Cohen, S.~B.; and Lapata, M. 2018.
\newblock Don't Give Me the Details, Just the Summary! Topic-Aware Convolutional Neural Networks for Extreme Summarization.
\newblock \emph{ArXiv}, abs/1808.08745.

\bibitem[{OpenAI et~al.(2024)OpenAI, Achiam, Adler, Agarwal, Ahmad, Akkaya, Aleman, Almeida, Altenschmidt, Altman, Anadkat, Avila, Babuschkin, Balaji, Balcom, Baltescu, Bao, Bavarian, Belgum, Bello, Berdine, Bernadett-Shapiro, Berner, Bogdonoff, Boiko, Boyd, Brakman, Brockman, Brooks, Brundage, Button, Cai, Campbell, Cann, Carey, Carlson, Carmichael, Chan, Chang, Chantzis, Chen, Chen, Chen, Chen, Chen, Chess, Cho, Chu, Chung, Cummings, Currier, Dai, Decareaux, Degry, Deutsch, Deville, Dhar, Dohan, Dowling, Dunning, Ecoffet, Eleti, Eloundou, Farhi, Fedus, Felix, Fishman, Forte, Fulford, Gao, Georges, Gibson, Goel, Gogineni, Goh, Gontijo-Lopes, Gordon, Grafstein, Gray, Greene, Gross, Gu, Guo, Hallacy, Han, Harris, He, Heaton, Heidecke, Hesse, Hickey, Hickey, Hoeschele, Houghton, Hsu, Hu, Hu, Huizinga, Jain, Jain, Jang, Jiang, Jiang, Jin, Jin, Jomoto, Jonn, Jun, Kaftan, Łukasz Kaiser, Kamali, Kanitscheider, Keskar, Khan, Kilpatrick, Kim, Kim, Kim, Kirchner, Kiros, Knight, Kokotajlo, Łukasz Kondraciuk,
  Kondrich, Konstantinidis, Kosic, Krueger, Kuo, Lampe, Lan, Lee, Leike, Leung, Levy, Li, Lim, Lin, Lin, Litwin, Lopez, Lowe, Lue, Makanju, Malfacini, Manning, Markov, Markovski, Martin, Mayer, Mayne, McGrew, McKinney, McLeavey, McMillan, McNeil, Medina, Mehta, Menick, Metz, Mishchenko, Mishkin, Monaco, Morikawa, Mossing, Mu, Murati, Murk, Mély, Nair, Nakano, Nayak, Neelakantan, Ngo, Noh, Ouyang, O'Keefe, Pachocki, Paino, Palermo, Pantuliano, Parascandolo, Parish, Parparita, Passos, Pavlov, Peng, Perelman, de~Avila Belbute~Peres, Petrov, de~Oliveira~Pinto, Michael, Pokorny, Pokrass, Pong, Powell, Power, Power, Proehl, Puri, Radford, Rae, Ramesh, Raymond, Real, Rimbach, Ross, Rotsted, Roussez, Ryder, Saltarelli, Sanders, Santurkar, Sastry, Schmidt, Schnurr, Schulman, Selsam, Sheppard, Sherbakov, Shieh, Shoker, Shyam, Sidor, Sigler, Simens, Sitkin, Slama, Sohl, Sokolowsky, Song, Staudacher, Such, Summers, Sutskever, Tang, Tezak, Thompson, Tillet, Tootoonchian, Tseng, Tuggle, Turley, Tworek, Uribe, Vallone,
  Vijayvergiya, Voss, Wainwright, Wang, Wang, Wang, Ward, Wei, Weinmann, Welihinda, Welinder, Weng, Weng, Wiethoff, Willner, Winter, Wolrich, Wong, Workman, Wu, Wu, Wu, Xiao, Xu, Yoo, Yu, Yuan, Zaremba, Zellers, Zhang, Zhang, Zhao, Zheng, Zhuang, Zhuk, and Zoph}]{gpt4}
OpenAI; Achiam, J.; Adler, S.; Agarwal, S.; Ahmad, L.; Akkaya, I.; Aleman, F.~L.; Almeida, D.; Altenschmidt, J.; Altman, S.; Anadkat, S.; Avila, R.; Babuschkin, I.; Balaji, S.; Balcom, V.; Baltescu, P.; Bao, H.; Bavarian, M.; Belgum, J.; Bello, I.; Berdine, J.; Bernadett-Shapiro, G.; Berner, C.; Bogdonoff, L.; Boiko, O.; Boyd, M.; Brakman, A.-L.; Brockman, G.; Brooks, T.; Brundage, M.; Button, K.; Cai, T.; Campbell, R.; Cann, A.; Carey, B.; Carlson, C.; Carmichael, R.; Chan, B.; Chang, C.; Chantzis, F.; Chen, D.; Chen, S.; Chen, R.; Chen, J.; Chen, M.; Chess, B.; Cho, C.; Chu, C.; Chung, H.~W.; Cummings, D.; Currier, J.; Dai, Y.; Decareaux, C.; Degry, T.; Deutsch, N.; Deville, D.; Dhar, A.; Dohan, D.; Dowling, S.; Dunning, S.; Ecoffet, A.; Eleti, A.; Eloundou, T.; Farhi, D.; Fedus, L.; Felix, N.; Fishman, S.~P.; Forte, J.; Fulford, I.; Gao, L.; Georges, E.; Gibson, C.; Goel, V.; Gogineni, T.; Goh, G.; Gontijo-Lopes, R.; Gordon, J.; Grafstein, M.; Gray, S.; Greene, R.; Gross, J.; Gu, S.~S.; Guo, Y.; Hallacy,
  C.; Han, J.; Harris, J.; He, Y.; Heaton, M.; Heidecke, J.; Hesse, C.; Hickey, A.; Hickey, W.; Hoeschele, P.; Houghton, B.; Hsu, K.; Hu, S.; Hu, X.; Huizinga, J.; Jain, S.; Jain, S.; Jang, J.; Jiang, A.; Jiang, R.; Jin, H.; Jin, D.; Jomoto, S.; Jonn, B.; Jun, H.; Kaftan, T.; Łukasz Kaiser; Kamali, A.; Kanitscheider, I.; Keskar, N.~S.; Khan, T.; Kilpatrick, L.; Kim, J.~W.; Kim, C.; Kim, Y.; Kirchner, J.~H.; Kiros, J.; Knight, M.; Kokotajlo, D.; Łukasz Kondraciuk; Kondrich, A.; Konstantinidis, A.; Kosic, K.; Krueger, G.; Kuo, V.; Lampe, M.; Lan, I.; Lee, T.; Leike, J.; Leung, J.; Levy, D.; Li, C.~M.; Lim, R.; Lin, M.; Lin, S.; Litwin, M.; Lopez, T.; Lowe, R.; Lue, P.; Makanju, A.; Malfacini, K.; Manning, S.; Markov, T.; Markovski, Y.; Martin, B.; Mayer, K.; Mayne, A.; McGrew, B.; McKinney, S.~M.; McLeavey, C.; McMillan, P.; McNeil, J.; Medina, D.; Mehta, A.; Menick, J.; Metz, L.; Mishchenko, A.; Mishkin, P.; Monaco, V.; Morikawa, E.; Mossing, D.; Mu, T.; Murati, M.; Murk, O.; Mély, D.; Nair, A.; Nakano, R.;
  Nayak, R.; Neelakantan, A.; Ngo, R.; Noh, H.; Ouyang, L.; O'Keefe, C.; Pachocki, J.; Paino, A.; Palermo, J.; Pantuliano, A.; Parascandolo, G.; Parish, J.; Parparita, E.; Passos, A.; Pavlov, M.; Peng, A.; Perelman, A.; de~Avila Belbute~Peres, F.; Petrov, M.; de~Oliveira~Pinto, H.~P.; Michael; Pokorny; Pokrass, M.; Pong, V.~H.; Powell, T.; Power, A.; Power, B.; Proehl, E.; Puri, R.; Radford, A.; Rae, J.; Ramesh, A.; Raymond, C.; Real, F.; Rimbach, K.; Ross, C.; Rotsted, B.; Roussez, H.; Ryder, N.; Saltarelli, M.; Sanders, T.; Santurkar, S.; Sastry, G.; Schmidt, H.; Schnurr, D.; Schulman, J.; Selsam, D.; Sheppard, K.; Sherbakov, T.; Shieh, J.; Shoker, S.; Shyam, P.; Sidor, S.; Sigler, E.; Simens, M.; Sitkin, J.; Slama, K.; Sohl, I.; Sokolowsky, B.; Song, Y.; Staudacher, N.; Such, F.~P.; Summers, N.; Sutskever, I.; Tang, J.; Tezak, N.; Thompson, M.~B.; Tillet, P.; Tootoonchian, A.; Tseng, E.; Tuggle, P.; Turley, N.; Tworek, J.; Uribe, J. F.~C.; Vallone, A.; Vijayvergiya, A.; Voss, C.; Wainwright, C.; Wang,
  J.~J.; Wang, A.; Wang, B.; Ward, J.; Wei, J.; Weinmann, C.; Welihinda, A.; Welinder, P.; Weng, J.; Weng, L.; Wiethoff, M.; Willner, D.; Winter, C.; Wolrich, S.; Wong, H.; Workman, L.; Wu, S.; Wu, J.; Wu, M.; Xiao, K.; Xu, T.; Yoo, S.; Yu, K.; Yuan, Q.; Zaremba, W.; Zellers, R.; Zhang, C.; Zhang, M.; Zhao, S.; Zheng, T.; Zhuang, J.; Zhuk, W.; and Zoph, B. 2024.
\newblock GPT-4 Technical Report.
\newblock arXiv:2303.08774.

\bibitem[{Paszke et~al.(2017)Paszke, Gross, Chintala, Chanan, Yang, DeVito, Lin, Desmaison, Antiga, and Lerer}]{pytorch}
Paszke, A.; Gross, S.; Chintala, S.; Chanan, G.; Yang, E.; DeVito, Z.; Lin, Z.; Desmaison, A.; Antiga, L.; and Lerer, A. 2017.
\newblock Automatic Differentiation in PyTorch.
\newblock In \emph{NIPS 2017 Workshop on Autodiff}.

\bibitem[{Radford et~al.(2019)Radford, Wu, Child, Luan, Amodei, Sutskever et~al.}]{gpt2}
Radford, A.; Wu, J.; Child, R.; Luan, D.; Amodei, D.; Sutskever, I.; et~al. 2019.
\newblock Language models are unsupervised multitask learners.
\newblock \emph{OpenAI blog}, 1(8): 9.

\bibitem[{Raffel et~al.(2020)Raffel, Shazeer, Roberts, Lee, Narang, Matena, Zhou, Li, and Liu}]{t5}
Raffel, C.; Shazeer, N.; Roberts, A.; Lee, K.; Narang, S.; Matena, M.; Zhou, Y.; Li, W.; and Liu, P.~J. 2020.
\newblock Exploring the limits of transfer learning with a unified text-to-text transformer.
\newblock \emph{The Journal of Machine Learning Research}, 21(1): 5485--5551.

\bibitem[{Rajpurkar, Jia, and Liang(2018)}]{squadv2}
Rajpurkar, P.; Jia, R.; and Liang, P. 2018.
\newblock Know What You Don't Know: Unanswerable Questions for SQuAD.
\newblock arXiv:1806.03822.

\bibitem[{Rajpurkar et~al.(2016)Rajpurkar, Zhang, Lopyrev, and Liang}]{squad}
Rajpurkar, P.; Zhang, J.; Lopyrev, K.; and Liang, P. 2016.
\newblock SQuAD: 100,000+ Questions for Machine Comprehension of Text.
\newblock arXiv:1606.05250.

\bibitem[{Reddi, Kale, and Kumar(2019)}]{amsgrad}
Reddi, S.~J.; Kale, S.; and Kumar, S. 2019.
\newblock On the convergence of adam and beyond.
\newblock \emph{arXiv preprint arXiv:1904.09237}.

\bibitem[{Rombach et~al.(2022)Rombach, Blattmann, Lorenz, Esser, and Ommer}]{ldm}
Rombach, R.; Blattmann, A.; Lorenz, D.; Esser, P.; and Ommer, B. 2022.
\newblock High-Resolution Image Synthesis with Latent Diffusion Models.
\newblock arXiv:2112.10752.

\bibitem[{Ruder(2016)}]{ruder2016overview}
Ruder, S. 2016.
\newblock An overview of gradient descent optimization algorithms.
\newblock \emph{arXiv preprint arXiv:1609.04747}.

\bibitem[{Russakovsky et~al.(2015)Russakovsky, Deng, Su, Krause, Satheesh, Ma, Huang, Karpathy, Khosla, Bernstein, Berg, and Fei-Fei}]{imagenet}
Russakovsky, O.; Deng, J.; Su, H.; Krause, J.; Satheesh, S.; Ma, S.; Huang, Z.; Karpathy, A.; Khosla, A.; Bernstein, M.; Berg, A.~C.; and Fei-Fei, L. 2015.
\newblock {ImageNet Large Scale Visual Recognition Challenge}.
\newblock \emph{International Journal of Computer Vision (IJCV)}, 115(3): 211--252.

\bibitem[{Shazeer and Stern(2018)}]{adafactor}
Shazeer, N.; and Stern, M. 2018.
\newblock Adafactor: Adaptive Learning Rates with Sublinear Memory Cost.
\newblock In Dy, J.; and Krause, A., eds., \emph{Proceedings of the 35th International Conference on Machine Learning}, volume~80 of \emph{Proceedings of Machine Learning Research}, 4596--4604. PMLR.

\bibitem[{Shi, Wang, and Shi(2014)}]{bmf3}
Shi, Z.; Wang, L.; and Shi, L. 2014.
\newblock Approximation method to rank-one binary matrix factorization.
\newblock In \emph{2014 IEEE International Conference on Automation Science and Engineering (CASE)}, 800--805.

\bibitem[{Takase et~al.(2024)Takase, Kiyono, Kobayashi, and Suzuki}]{loss-spike-01}
Takase, S.; Kiyono, S.; Kobayashi, S.; and Suzuki, J. 2024.
\newblock Spike No More: Stabilizing the Pre-training of Large Language Models.
\newblock arXiv:2312.16903.

\bibitem[{Taori et~al.(2023)Taori, Gulrajani, Zhang, Dubois, Li, Guestrin, Liang, and Hashimoto}]{alpaca}
Taori, R.; Gulrajani, I.; Zhang, T.; Dubois, Y.; Li, X.; Guestrin, C.; Liang, P.; and Hashimoto, T.~B. 2023.
\newblock Stanford Alpaca: An Instruction-following LLaMA model.
\newblock \url{https://github.com/tatsu-lab/stanford_alpaca}.

\bibitem[{Touvron et~al.(2023{\natexlab{a}})Touvron, Lavril, Izacard, Martinet, Lachaux, Lacroix, Rozière, Goyal, Hambro, Azhar, Rodriguez, Joulin, Grave, and Lample}]{llama1}
Touvron, H.; Lavril, T.; Izacard, G.; Martinet, X.; Lachaux, M.-A.; Lacroix, T.; Rozière, B.; Goyal, N.; Hambro, E.; Azhar, F.; Rodriguez, A.; Joulin, A.; Grave, E.; and Lample, G. 2023{\natexlab{a}}.
\newblock LLaMA: Open and Efficient Foundation Language Models.
\newblock arXiv:2302.13971.

\bibitem[{Touvron et~al.(2023{\natexlab{b}})Touvron, Martin, Stone, Albert, Almahairi, Babaei, Bashlykov, Batra, Bhargava, Bhosale, Bikel, Blecher, Ferrer, Chen, Cucurull, Esiobu, Fernandes, Fu, Fu, Fuller, Gao, Goswami, Goyal, Hartshorn, Hosseini, Hou, Inan, Kardas, Kerkez, Khabsa, Kloumann, Korenev, Koura, Lachaux, Lavril, Lee, Liskovich, Lu, Mao, Martinet, Mihaylov, Mishra, Molybog, Nie, Poulton, Reizenstein, Rungta, Saladi, Schelten, Silva, Smith, Subramanian, Tan, Tang, Taylor, Williams, Kuan, Xu, Yan, Zarov, Zhang, Fan, Kambadur, Narang, Rodriguez, Stojnic, Edunov, and Scialom}]{llama2}
Touvron, H.; Martin, L.; Stone, K.; Albert, P.; Almahairi, A.; Babaei, Y.; Bashlykov, N.; Batra, S.; Bhargava, P.; Bhosale, S.; Bikel, D.; Blecher, L.; Ferrer, C.~C.; Chen, M.; Cucurull, G.; Esiobu, D.; Fernandes, J.; Fu, J.; Fu, W.; Fuller, B.; Gao, C.; Goswami, V.; Goyal, N.; Hartshorn, A.; Hosseini, S.; Hou, R.; Inan, H.; Kardas, M.; Kerkez, V.; Khabsa, M.; Kloumann, I.; Korenev, A.; Koura, P.~S.; Lachaux, M.-A.; Lavril, T.; Lee, J.; Liskovich, D.; Lu, Y.; Mao, Y.; Martinet, X.; Mihaylov, T.; Mishra, P.; Molybog, I.; Nie, Y.; Poulton, A.; Reizenstein, J.; Rungta, R.; Saladi, K.; Schelten, A.; Silva, R.; Smith, E.~M.; Subramanian, R.; Tan, X.~E.; Tang, B.; Taylor, R.; Williams, A.; Kuan, J.~X.; Xu, P.; Yan, Z.; Zarov, I.; Zhang, Y.; Fan, A.; Kambadur, M.; Narang, S.; Rodriguez, A.; Stojnic, R.; Edunov, S.; and Scialom, T. 2023{\natexlab{b}}.
\newblock Llama 2: Open Foundation and Fine-Tuned Chat Models.
\newblock arXiv:2307.09288.

\bibitem[{Ultralytics(2021)}]{ultralytics2021yolov5}
Ultralytics. 2021.
\newblock {YOLOv5}: {A} state-of-the-art real-time object detection system.
\newblock \url{https://docs.ultralytics.com}.

\bibitem[{Vaswani et~al.(2017)Vaswani, Shazeer, Parmar, Uszkoreit, Jones, Gomez, Kaiser, and Polosukhin}]{transformer}
Vaswani, A.; Shazeer, N.; Parmar, N.; Uszkoreit, J.; Jones, L.; Gomez, A.~N.; Kaiser, {\L}.; and Polosukhin, I. 2017.
\newblock Attention is all you need.
\newblock \emph{Advances in neural information processing systems}, 30.

\bibitem[{Wang et~al.(2018)Wang, Singh, Michael, Hill, Levy, and Bowman}]{glue}
Wang, A.; Singh, A.; Michael, J.; Hill, F.; Levy, O.; and Bowman, S.~R. 2018.
\newblock GLUE: A multi-task benchmark and analysis platform for natural language understanding.
\newblock \emph{arXiv preprint arXiv:1804.07461}.

\bibitem[{Zhang and Xu(2023)}]{loss-spike-02}
Zhang, Z.; and Xu, Z.-Q.~J. 2023.
\newblock Loss Spike in Training Neural Networks.
\newblock arXiv:2305.12133.

\bibitem[{Zhu et~al.(2015)Zhu, Kiros, Zemel, Salakhutdinov, Urtasun, Torralba, and Fidler}]{bookcorpus}
Zhu, Y.; Kiros, R.; Zemel, R.; Salakhutdinov, R.; Urtasun, R.; Torralba, A.; and Fidler, S. 2015.
\newblock Aligning Books and Movies: Towards Story-Like Visual Explanations by Watching Movies and Reading Books.
\newblock In \emph{The IEEE International Conference on Computer Vision (ICCV)}.

\end{thebibliography}
